\documentclass{article} %
\usepackage{iclr2027_conference,times}
\iclrpreprintcopy

\usepackage[utf8]{inputenc} %
\usepackage[T1]{fontenc}    %
\usepackage{hyperref}       %
\usepackage{url}            %
\usepackage{booktabs}       %
\usepackage{amsfonts}       %
\usepackage{nicefrac}       %
\usepackage{microtype}      %
\usepackage[table]{xcolor}
\usepackage{multirow}    
\usepackage{makecell}  
\usepackage{graphicx}
\usepackage{tikz}
\usetikzlibrary{shapes.geometric}
\usepackage{subcaption}
\usepackage{caption}
\usepackage{adjustbox}

\newlength{\panelheight}

\AddToHook{env/table/begin}{\small}
\AddToHook{env/table*/begin}{\small}
\AddToHook{env/figure/begin}{\small}
\AddToHook{env/figure*/begin}{\small}
\AddToHook{env/wrapfigure/begin}{\small}

\usepackage{amsmath}
\usepackage{wrapfig}
\usepackage{placeins}
\usepackage{needspace}
\usepackage{colortbl}
\usepackage{amssymb}
\usepackage{anyfontsize}
\usepackage[most]{tcolorbox}
\usepackage{algorithm}      %
\usepackage{algpseudocode}  %
\usepackage{soul}

\definecolor{degred}{RGB}{163, 45, 45}
\definecolor{retgreen}{RGB}{29, 158, 117}
\definecolor{nagray}{RGB}{160, 160, 160}
\definecolor{selfsolvecolor}{RGB}{155, 70, 70}
\definecolor{accuracycolor}{RGB}{90, 130, 90}

\newcommand{\csalong}{Capability Self-Assessment}
\newcommand{\csa}{CSA}

\newcommand{\sftlabel}{SFT\texorpdfstring{\textsubscript{Label}}{Label}}
\newcommand{\sftself}{SFT\texorpdfstring{\textsubscript{Self}}{Self}}
\newcommand{\sftteacher}{SFT\texorpdfstring{\textsubscript{Teacher}}{Teacher}}
\newcommand{\CDS}{CDS}
\newcommand{\CDSfull}{Capability Discrimination Score}
\newcommand{\MF}{M-F1}

\newcommand{\MFfull}{Macro F1}
\newcommand{\selfsolve}{\textsc{Self-SOLVE}}
\newcommand{\delegate}{\textsc{DELEGATE}}
\newcommand{\CR}{CR}
\newcommand{\CRfull}{Capability Ratio}
\newcommand{\rlvr}{RLVR}
\usepackage{bbm}

\title{Capability Self-Assessment in Large Language Models}

\author{%
  \textbf{Haoyan Yang\textsuperscript{1}\thanks{Equal Contributions.}},\,\,\,
  \textbf{Reza Shirkavand\textsuperscript{2}\footnotemark[1]},\,\,\,
  \textbf{Yukai Jin\textsuperscript{3}},\\[2pt]
  \textbf{Jiawei Zhou\textsuperscript{1}\thanks{Correspondence to: Jiawei Zhou (\texttt{jiawei.zhou.1@stonybrook.edu}), Shangqian Gao (\texttt{sgao@cs.fsu.edu}), Heng Huang (\texttt{heng@cs.umd.edu}).}},\,\,\,
  \textbf{Shangqian Gao\textsuperscript{3}\footnotemark[2]},\,\,\,
  \textbf{Heng Huang\textsuperscript{2}\footnotemark[2]}\\[4pt]
  \textsuperscript{1}Stony Brook University \quad
  \textsuperscript{2}University of Maryland \quad
  \textsuperscript{3}Florida State University
}

\begin{document}

\maketitle
\begin{abstract}\label{sec:abstract}
The ability to recognize one's own limitations and decide whether to solve a problem or seek help is fundamental to reliable intelligent systems. Yet modern large language models tend to overestimate their competence and attempt queries they cannot solve.
We study how models can learn \csalong\ (\csa) while preserving their problem-solving ability. Because the same model both assesses and solves queries, self-assessment training can change the capability it must recognize. We systematically compare supervised fine-tuning (SFT) and reinforcement learning (RL), evaluating each trained model’s self-assessment against its updated capability and measuring how well it retains its solving ability. Across model families, scales, and domains, RL provides the strongest assessment–retention trade-off. We also find that \csa\ transfers to the out-of-distribution settings we evaluate, and demonstrate its practical value for local--cloud routing at inference time and targeted data selection during training.
Code is available at \url{https://github.com/Joyyang158/llm-csa}.
\end{abstract}

\section{Introduction}\label{sec:intro}

An ideal intelligent system not only solves problems well, but also recognizes its limits. It asks for help when a problem exceeds its ability or uses failure to identify what it should learn next, as humans do \citep{madras2018predictresponsiblyimprovingfairness}. However, large language models~(LLMs) often elect to answer queries they cannot solve reliably, leading to hallucinated outputs~\citep{farquhar2024detecting-hallucinations-entropy} and wasted compute~\citep{shirkavand2025costaware}.
Prior work has studied whether LLMs can assess answer correctness and recognize unanswerable questions~\citep{kadavath2022llms-mostly-know,yin2023do-llms-know-what-they-dont-know}, and trained models to estimate confidence~\citep{kapoor2024llms-must-be-taugh} or abstain when uncertain~\citep{zhang2024r}.

We ask whether a generative model can be post-trained to predict its own query-level success, \emph{while preserving its problem-solving ability}. We use \csalong\ (\csa) to denote a pre-answer choice between \selfsolve, where the model attempts the query itself, and \delegate, where the query is handed to a stronger system.
Because the same model both assesses and solves queries, self-assessment training can shift the capability boundary it must recognize. We therefore study how to learn and evaluate model-specific self-assessment policies while accounting for changes in problem-solving ability.

To investigate this question, we compare supervised fine-tuning (SFT) with reinforcement learning with verifiable rewards (RLVR)~\citep{shao2024deepseekmath,guo2025deepseek-r1}. We evaluate both approaches within a common \csa\ framework, measuring self-assessment quality alongside changes in the model's original capabilities. Across model families, scales, and domains, we find that SFT can improve self-assessment at the expense of task performance, whereas RLVR can improve \csa\ while preserving the model's  problem-solving ability. We observe transfer across domains and datasets, suggesting that the learned assessment policy remains useful beyond its training distribution.

We then explore two practical uses of this learned decision signal. At inference time, \csa\ improves local--cloud routing by identifying queries that require a stronger model. During training, \csa\ can be used to identify examples that may lie near or beyond the model's current competence, providing a targeted signal for data selection. 
In this way, learned \csa\ can play a dual role analogous to competence assessment in effective problem solving, guiding when to act or delegate at inference time, and guiding what to learn next during training.

In summary, our contributions are:
\begin{enumerate}
\item We evaluate pre-answer capability assessment jointly with the effects of assessment training on the model. We use both aggregate task performance and query-level capability stability to distinguish assessment gains from changes in the capability being assessed.

\item We systematically compare training-free methods, SFT, and RLVR for acquiring \csa. Our experiments show that RLVR can improve self-assessment while preserving problem-solving ability, with evidence of transfer across domains and datasets.

\item We demonstrate the practical value of \csa\ for local--cloud routing and targeted data selection, guiding both when to seek help and what to learn next.
\end{enumerate}

\section{Related Work}\label{sec:rel-work}
\begin{table}[t]
\centering
\scriptsize
\setlength{\tabcolsep}{3pt}
\renewcommand{\arraystretch}{1.2}

\caption{
Comparison with representative capability-assessment and
uncertainty-estimation methods.
\emph{Trained for assessment} means model parameters are updated
specifically for assessment. \emph{Assessment before answering} means
the assessment is made without generating any candidate answer.
\emph{Same model assesses and solves} allows an auxiliary assessment
head on the solving model. \emph{Solving ability evaluated after
training} denotes an explicit comparison of solving performance before
and after assessment training.
$\checkmark$ indicates an explicitly studied feature. --- indicates
that it is not part of the listed setting.
}
\label{tab:related_work}

\begin{tabular}{
    @{}
    >{\raggedright\arraybackslash}p{0.32\linewidth}
    >{\centering\arraybackslash}
        p{\dimexpr0.14\linewidth-2\tabcolsep\relax}
    >{\centering\arraybackslash}
        p{\dimexpr0.14\linewidth-2\tabcolsep\relax}
    >{\centering\arraybackslash}
        p{\dimexpr0.21\linewidth-2\tabcolsep\relax}
    >{\centering\arraybackslash}
        p{\dimexpr0.19\linewidth-2\tabcolsep\relax}
    @{}
}
\toprule
Method
& \shortstack{Trained for\\assessment}
& \shortstack{Assessment\\before answering}
& \shortstack{Same model\\assesses and solves}
& \shortstack{Solving ability\\evaluated after training}
\\
\midrule

Self-consistency~\citep{wang2023selfconsistency}
& --- & --- & --- & ---
\\

Semantic entropy~\citep{farquhar2024detecting-hallucinations-entropy}
& --- & --- & --- & ---
\\

$P(\mathrm{IK})$~\citep{kadavath2022llms-mostly-know}
& $\checkmark$ & $\checkmark$ & $\checkmark$ & ---
\\

Internal Confidence~\citep{chen2026querylevel}
& --- & $\checkmark$ & $\checkmark$ & ---
\\

Calibration tuning~\citep{kapoor2024llms-must-be-taugh}
& $\checkmark$ & --- & --- & ---
\\

Initial self-assessment~\citep{barkan2026do}
& --- & $\checkmark$ & $\checkmark$ & ---
\\

RL-trained confidence~\citep{damani2026beyond}
& $\checkmark$ & --- & $\checkmark$ & Partial
\\

\midrule
\textbf{Ours (CSA training)}
& $\checkmark$ & $\checkmark$ & $\checkmark$ & $\checkmark$
\\

\bottomrule
\end{tabular}
\vspace{-8pt}
\end{table}

\textbf{Pre-answer capability prediction.} 
\citet{kadavath2022llms-mostly-know} distinguish assessment of a generated answer through P(True) from prediction of whether a model knows an answer through P(IK). Their learned P(IK) predictor and cross-task evaluations are direct precedents for our setting. \citet{chen2026querylevel} introduce a training-free query-level uncertainty signal and demonstrate adaptive inference, including model cascading. \citet{barkan2026do} study predictions of task success and their relationship to overconfidence and decision making. We build on this literature by comparing assessment-training methods while measuring their effects on the model and evaluating predictions against its post-training capability.

\textbf{Confidence, abstention, and deferral.}
Confidence estimation, abstention, and learning to defer provide closely related objectives and decision rules~\citep{madras2018predictresponsiblyimprovingfairness,kapoor2024llms-must-be-taugh,zhang2024r,wen2025abstention-survey}. Sampling-based methods estimate uncertainty from multiple generated outputs~\citep{wang2023selfconsistency,manakul2023selfcheckgpt,farquhar2024detecting-hallucinations-entropy,kossen2024semantic-entropy}. \citet{gul2025mash} and \citet{damani2026beyond} use RL to train help-seeking and calibrated confidence, but neither extensively evaluates per-query capability retention.

Our experiments compare these signals and training approaches within the assessment–retention setting. Moreover, unlike much prior work that treats self-knowledge as a safeguard against hallucination, using confidence or abstention to avoid unreliable answers, we explore the model's own capability boundary as a signal for adaptation. The same learned decision determines when the model acts or seeks help at inference time and which examples it learns from during training.
Table~\ref{tab:related_work} provides a comparison of our work and existing studies. Appendix~\ref{sec:rel-work-full} provides a broader discussion. 

\section{\csa{} through Training}\label{sec:method}
\vspace{-3pt}
Our prompt-only study shows that prompting alone does not elicit reliable \csa\ in the evaluated settings (Appendix~\ref{app:prompting-study}). We therefore compare training strategies for improving this assessment. We formulate \csa\ learning as policy optimization with a capability-preservation goal and evaluate how well different training methods meet it.

\subsection{\csa{} Formulation}
\label{sec:csa-formulation}

Given a query $q_i$, a model makes a binary decision between \selfsolve\ and \delegate\ before attempting an answer. Let $y_i$
denote its capability label under a specified task and evaluation
protocol. The policy $\pi_\theta^{\mathrm{CSA}}$ takes
$x_i = [p; q_i]$ as input, where $p$ is a \csa\ instruction, and
predicts a decision $\hat{y}_i$ without observing a candidate solution.

\begin{figure}
    \centering
    \includegraphics[width=\textwidth]{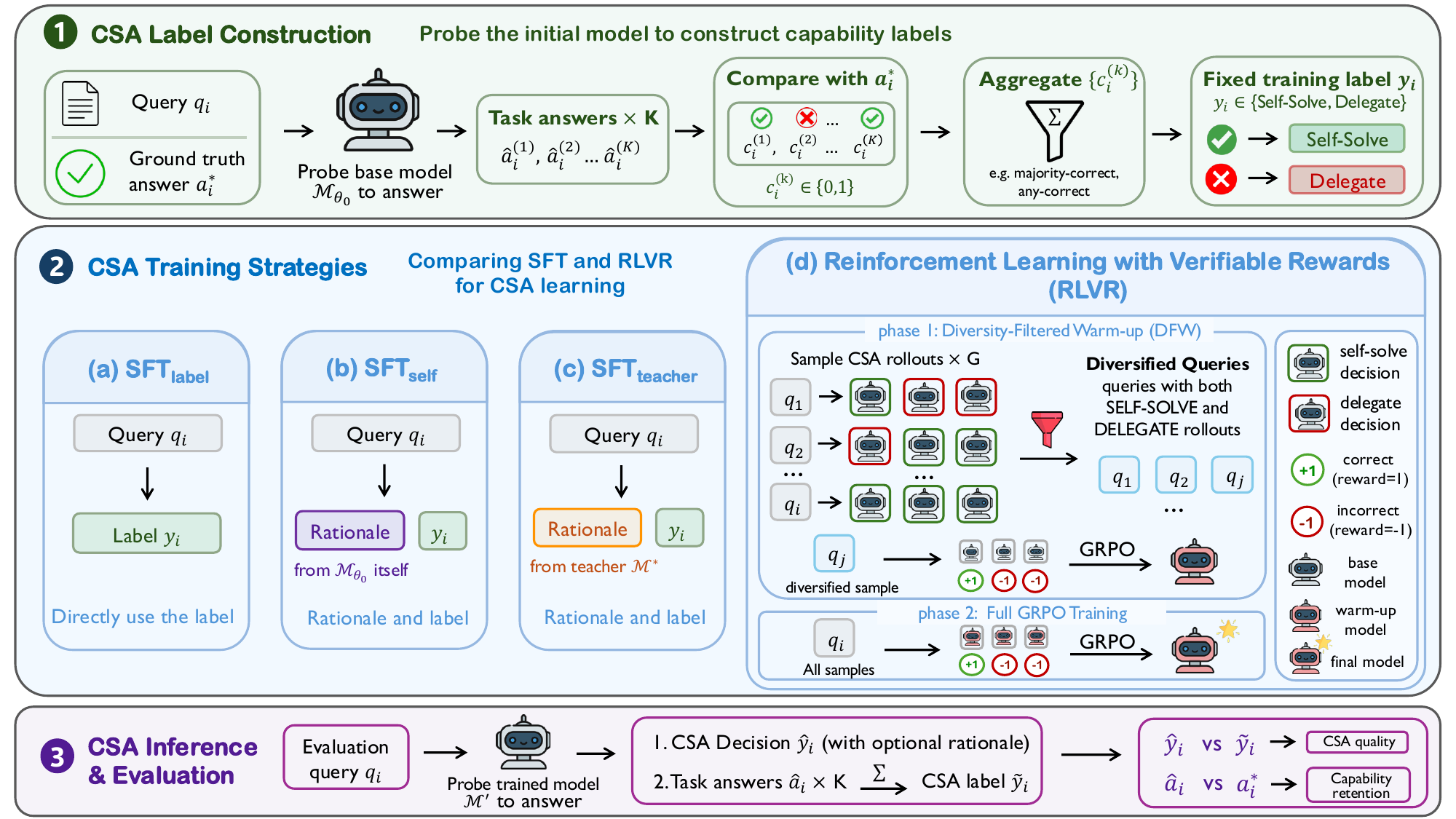}
    \caption{Overview of the \csa\ training and evaluation framework.
    \textbf{(1) Label construction}: aggregate $K$ solution attempts from the initial model into fixed capability labels $y_i$.
    \textbf{(2) Training}: learn self-assessment using these labels via three SFT variants (label-only, self-rationale, and teacher-rationale) or RLVR.
    \textbf{(3) Inference and Evaluation}: predict \selfsolve\ or \delegate\ before answering and evaluate decisions against updated capability labels $\tilde{y}_i$.}
    \label{fig:pipeline}
    \vspace{-8pt}
\end{figure}

We formulate \csa\ learning as improving this decision policy while preserving the model's underlying problem-solving ability. Let $\theta_0$ denote the initial model parameters, $\Delta_{\mathrm{cap}}(\theta,\theta_0)$ measure capability change under the same evaluation protocol, and $u(\hat{y}_i,y_i)$ measure agreement between the predicted decision $\hat{y}_i$ and the
capability label $y_i$. The learning objective is
\begin{equation}
    \theta^\star
    =
    \underset{\theta}{\operatorname{arg\,max}}\;
    \mathbb{E}_{\substack{
        q_i\sim\mathcal{D},\,
        \hat{y}_i\sim\pi_\theta^{\mathrm{CSA}}(\cdot\mid x_i)
    }}
    \left[u\!\left(\hat{y}_i,y_i\right)\right]
    \quad\text{s.t.}\quad
    \Delta_{\mathrm{cap}}(\theta,\theta_0)\leq\varepsilon,
    \label{eq:csa-constrained-objective}
\end{equation}
where $\varepsilon$ specifies the tolerated capability change. The choice of $\Delta_{\mathrm{cap}}$ determines whether preservation is assessed through aggregate task performance or finer-grained capability profiles. We use fixed capability labels from the initial model as training targets. Because training may change the capability being assessed, we re-probe each trained model for evaluation and separately compare its solving performance with the initial checkpoint. The preservation constraint expresses the desired learning goal, and our training methods do not explicitly enforce it.
Figure~\ref{fig:pipeline} summarizes the training and evaluation framework.

\subsection{\csa{} Label Construction}
\label{sec:csa-definition}

We construct training labels by probing the initial model $\mathcal{M}_{\theta_0}$ under a fixed task and evaluation protocol. Given a dataset $\mathcal{D} = \{(q_i, a_i^*)\}_{i=1}^{N}$, we sample $K$ solution attempts $\{\hat{a}_i^{(k)}\}_{k=1}^{K}$ for each query. Let $c_i^{(k)} = \mathrm{Correct}(\hat{a}_i^{(k)}, a_i^*) \in \{0,1\}$ indicate whether an attempt passes the task-specific correctness check. We aggregate these outcomes into a binary capability label:
\vspace{-10pt}
\begin{equation}
    y_i = \begin{cases} \textsc{Self-Solve}, & \text{if } g\big(\{c_i^{(k)}\}_{k=1}^{K}\big) = 1, \\ \textsc{Delegate}, & \text{otherwise,} \end{cases}
    \label{eq:label}
\end{equation}
where $g$ is an aggregation rule, such as any-correct or majority-correct. Thus, $y_i$ denotes the initial model's empirical capability label for query $q_i$ under the chosen sampling and aggregation protocol.

We construct these labels once from the initial model and use them across all training strategies. They provide a common training target, but may no longer describe the model after training. We therefore evaluate each trained policy against freshly probed capability labels and separately measure problem-solving retention.

\subsection{\csa\ Training and Evaluation}
\label{sec:training}
\label{sec:evaluation}

We compare two post-training paradigms for learning the \csa\ policy: supervised fine-tuning (SFT) and reinforcement learning with verifiable rewards (RLVR). Both use the same fixed initial capability labels and query inputs. Each input combines a \csa\ prompt $p$ with the query, $x_i = [p; q_i]$. See Appendix~\ref{app:prompts} for all details.

Within SFT, we compare three forms of supervision.

\textbf{(a) \sftlabel}~\citep{zhang2024r,kapoor2024llms-must-be-taugh}. The target is the bare capability label, $o_i^{\text{label}} = y_i$. The model learns to emit the decision directly.

\textbf{(b) \sftself}~\citep{kadavath2022llms-mostly-know,chaudhry2024finetuninglanguagemodelsemit}. The initial model generates a rationale conditioned on the query input and capability label, $r_i^{\text{self}} \sim \mathcal{M}_{\theta_0}(\cdot \mid x_i, y_i)$. The training target is the rationale followed by the label, $o_i^{\text{self}} = [r_i^{\text{self}}; y_i]$. 

\textbf{(c) \sftteacher}~\citep{hager2025uncertaintydistillationteachinglanguage}. A stronger teacher model generates the rationale under the same conditioning, $r_i^{\text{teacher}} \sim \mathcal{M}^{*}(\cdot \mid x_i, y_i)$. The training target is $o_i^{\text{teacher}} = [r_i^{\text{teacher}}; y_i]$. 

All three SFT variants minimize the same negative log-likelihood objective:
\vspace{-5pt}
\begin{equation}
    \mathcal{L}_{\text{SFT}}(\theta) = -\sum_{i=1}^{N} \log p_\theta\!\left(o_i \,\middle|\, x_i\right), \qquad o_i \in \{o_i^{\text{label}},\, o_i^{\text{self}},\, o_i^{\text{teacher}}\},
    \label{eq:loss-sft}
\end{equation}

\vspace{-5pt}
\textbf{(d) RLVR.}
RLVR generates a rationale and a decision, $o_i^{(g)} = [r_i^{(g)}; \hat{y}_i^{(g)}]$, and rewards agreement between the decision and the fixed capability label: $R_i^{(g)}=+1$ if $\hat{y}_i^{(g)}=y_i$, and $-1$ otherwise. The reward evaluates the \csa\ decision, not a generated task answer. We optimize this policy using Group Relative Policy Optimization (GRPO)~\citep{shao2024deepseekmath}.

Training has two stages. \emph{Diversity-Filtered Warm-up (DFW)} runs GRPO on queries whose initial $G$ rollouts contain both \selfsolve\ and \delegate\ decisions. \emph{Full GRPO training} then continues from the warm-up checkpoint on the full dataset. Appendices~\ref{app:algorithm} and~\ref{app:rollout-warmup} provide more details. Both stages use the following objective:

\vspace{-6pt}
{\small
\begin{equation}
    \mathcal{L}_{\text{GRPO}}(\theta;\mathcal{B})
    = -\,\mathbb{E}_{\substack{x_i \sim \mathcal{B} \\ \{o_i^{(g)}\} \sim \pi_{\theta_{\text{old}}}}}
    \!\left[ \frac{1}{G}\sum_{g=1}^{G} \min\!\Big( \rho_i^{(g)}(\theta) A_i^{(g)},\!
    \mathrm{clip}\!\big(\rho_i^{(g)}(\theta), 1{-}\epsilon, 1{+}\epsilon\big) A_i^{(g)} \Big) \right]
    + \beta\,\mathrm{KL}\!\left[\pi_\theta \,\|\, \pi_{\text{ref}}\right],
\label{eq:loss-grpo}
\end{equation}
}
where $\rho_i^{(g)}(\theta) = \pi_\theta(o_i^{(g)} \mid x_i) / \pi_{\theta_{\text{old}}}(o_i^{(g)} \mid x_i)$, $\epsilon$ is the clipping threshold, $\beta$ is the KL strength, and $A_i^{(g)} = (R_i^{(g)} - \mathrm{mean}_{g'} R_i^{(g')}) / \mathrm{std}_{g'} R_i^{(g')}$, which is the group-normalized advantage within each rollout group. $\mathcal{B}$ is a mini-batch from $\mathcal{D}_{\text{div}}$ (Stage 1) or $\mathcal{D}$ (Stage 2).

\paragraph{Connection to Eq.~\ref{eq:csa-constrained-objective}.}
A Lagrangian relaxation of Eq.~\ref{eq:csa-constrained-objective}
produces a regularized objective with a capability penalty
$\Delta_{\mathrm{cap}}(\theta,\theta_0)$.
RLVR (Eq.~\ref{eq:loss-grpo}) uses KL regularization as a heuristic
approximation to this penalty. SFT (Eq.~\ref{eq:loss-sft}) maximizes target-output
likelihood and can optionally include the same KL penalty;
we use vanilla SFT by default, following common practice.

\paragraph{Inference and Evaluation.}
At inference time, the trained model receives $[p;q]$ and outputs a \csa\ decision, preceded by a rationale for all methods except \sftlabel. We re-probe each trained model on the evaluation queries using the sampling protocol and aggregation rule in Eq.~\eqref{eq:label}, obtaining updated capability labels $\{\tilde{y}_i\}$. We measure \csa\ quality by comparing the predicted decisions $\hat{y}_i$ with these labels. Separately, we assess capability retention by comparing task performance before and after training.

\begingroup
\setlength{\textfloatsep}{10pt plus 2pt minus 2pt}
\setlength{\floatsep}{6pt plus 2pt minus 2pt}
\setlength{\intextsep}{8pt plus 2pt minus 2pt}
\setlength{\parskip}{2pt}
\section{Experiments and Analysis}\label{sec:experiments}

\subsection{Setup}
\label{sec:experiment_setup}

We conduct experiments on the \textbf{Qwen3} \citep{yang2025qwen3technicalreport} family (1.7B, 4B, 8B, 14B), as well as \textbf{OLMo2} \citep{olmo20242} (7B, 13B). Evaluation spans two domains: \textbf{Math} (GSM8K~\citep{cobbe2021trainingverifierssolvemath},
MATH500~\citep{lightman2023lets}, and AIME~\citep{aime_1983_2024}) and \textbf{Science} (MMLU-Pro \citep{wang2024mmluprorobustchallengingmultitask} across the biology, chemistry, health, and physics categories). We use \textbf{\MFfull\ (\MF)} as the primary metric of \csa\ quality and report \textbf{\CDSfull\ (\CDS)} as an auxiliary metric. We use \textbf{\CRfull\ (\CR)} to measure average retention of the model's problem-solving ability. Full experimental details appear in Appendix~\ref{app:setup}.

\subsection{Main Results}\label{sec:main-results}

For each query, we instantiate Eq.~\eqref{eq:label} with $K=5$ independently sampled task answers. Repeated attempts reduce reliance on a single stochastic outcome. We use \emph{any-correct} for open-ended Math: at least one answer must be correct. For multiple-choice Science, we use \emph{majority-correct}: at least three answers must be correct, since guessing an option is more likely to succeed. These labels define capability under the sampling protocol, rather than single-attempt accuracy. Appendix~\ref{app:label} explains the construction, and Appendix~\ref{app:label-ablations} varies $K$ and the aggregation threshold at $K=5$.

\begin{table}[t]
\centering
\small
\caption{CSA performance across training strategies and domains (higher is better). \textbf{Bold} and \underline{underline} mark the best and second-best trained models. ---: \CDS\ is unavailable due to highly imbalanced decisions.}
\label{tab:main}
\setlength{\tabcolsep}{3.5pt}
\renewcommand{\arraystretch}{1.05}
\resizebox{\textwidth}{!}{%
\begin{tabular}{@{}ll *{5}{c} *{5}{c}@{}}
\toprule
& & \multicolumn{5}{c}{\textbf{Math}} & \multicolumn{5}{c}{\textbf{Science}} \\
\cmidrule(lr){3-7} \cmidrule(lr){8-12}
\textbf{Model} & \textbf{Metric}
& Vanilla & \sftlabel & \sftself & \sftteacher & \rlvr
& Vanilla & \sftlabel & \sftself & \sftteacher & \rlvr \\
\midrule
\multirow{3}{*}{Qwen3-1.7B}
  & \MF   & 0.53 & 0.40 & 0.68 & \underline{0.69} & \textbf{0.80} & 0.46 & 0.49 & \underline{0.61} & 0.55 & \textbf{0.67} \\
  & \CDS  & 5.15  & 3.41  & 7.67  & \underline{7.86}  & \textbf{17.76} & 3.02  & 5.30  & \underline{6.39}  & 2.75  & \textbf{10.12} \\
  & \CR  & 100    & 10.5  & \underline{91.9} & 73.1 & \textbf{98.8}  & 100    & 37.3  & \underline{89.5}  & 81.1  & \textbf{102.1} \\
\arrayrulecolor{black!25}\midrule\arrayrulecolor{black}
\multirow{3}{*}{Qwen3-4B}
  & \MF   & 0.51       & 0.41 & \underline{0.65} & 0.64 & \textbf{0.80} & 0.42       & 0.31 & 0.60 & \underline{0.64} & \textbf{0.64} \\
  & \CDS  & --- & 5.51  & \underline{6.14}  & 5.15  & \textbf{17.17} & --- & 4.47  & 5.92  & \underline{7.38}  & \textbf{8.07} \\
  & \CR  & 100          & 13.3  & 95.0              & \underline{96.4} & \textbf{100.7} & 100 & 20.7 & 86.9 & \underline{87.6} & \textbf{101.9} \\
\arrayrulecolor{black!25}\midrule\arrayrulecolor{black}
\multirow{3}{*}{Qwen3-8B}
  & \MF   & 0.49       & \underline{0.76} & 0.66 & 0.68 & \textbf{0.77} & 0.45       & 0.50          & 0.58 & \underline{0.62} & \textbf{0.62} \\
  & \CDS  & --- & \underline{11.10} & 6.08  & 6.35  & \textbf{14.13} & --- & \textbf{10.73} & 4.19  & 6.34  & \underline{6.72} \\
  & \CR  & 100          & 89.2              & 98.8  & \textbf{102.3} & \underline{101.4} & 100 & 49.1 & 91.5 & \underline{94.0} & \textbf{101.5} \\
\arrayrulecolor{black!25}\midrule\arrayrulecolor{black}
\multirow{3}{*}{Qwen3-14B}
  & \MF   & 0.51       & \underline{0.73} & 0.67 & 0.60 & \textbf{0.75} & 0.45       & 0.20 & \textbf{0.61} & \underline{0.60} & 0.57 \\
  & \CDS  & --- & \underline{9.31}  & 7.05  & 3.43  & \textbf{10.94} & --- & 2.25  & \textbf{5.57}  & 5.12  & \underline{5.25} \\
  & \CR  & 100          & 84.3              & 98.2  & \textbf{106.2} & \underline{100.9} & 100 & 6.7 & 92.5 & \underline{95.1} & \textbf{100.3} \\
\arrayrulecolor{black!25}\midrule\arrayrulecolor{black}
\multirow{3}{*}{OLMo2-7B}
  & \MF   & 0.74 & 0.30 & \underline{0.84} & 0.83 & \textbf{0.89} & 0.55 & 0.44 & \underline{0.66} & 0.65 & \textbf{0.69} \\
  & \CDS  & 10.87 & 3.63  & \underline{19.07} & 18.26 & \textbf{26.60} & 3.44  & 0.00  & \underline{7.83}  & 7.08  & \textbf{10.94} \\
  & \CR  & 100    & 2.1   & \underline{96.6}  & 92.8  & \textbf{101.9} & 100    & 0.6   & \underline{88.5}  & 76.8  & \textbf{106.7} \\
\arrayrulecolor{black!25}\midrule\arrayrulecolor{black}
\multirow{3}{*}{OLMo2-13B}
  & \MF   & 0.67 & 0.28 & 0.78             & \underline{0.79} & \textbf{0.89} & 0.54 & 0.42 & 0.61 & \underline{0.64} & \textbf{0.66} \\
  & \CDS  & 7.24  & 0.90  & \underline{16.06} & 15.02 & \textbf{22.73} & 6.13 & 0.00  & 5.70  & \underline{6.99}  & \textbf{8.40} \\
  & \CR  & 100    & 3.0   & 49.7              & \underline{72.2}  & \textbf{100.8} & 100    & 5.0   & \underline{76.4}     & 61.2 & \textbf{99.6} \\
\bottomrule
\end{tabular}%
}
\end{table}

\paragraph{RLVR outperforms SFT in balancing assessment quality and capability retention.}
Table~\ref{tab:main} shows that, across the evaluated models and tasks, RLVR outperforms the SFT baselines in terms of \MF\ in most cases. It also retains approximately 100\% of the original task performance on average. Mean \CR\ is 100.8\% on Math and 102.0\% on Science, compared with only 33.7\% and 19.9\% for \sftlabel. The rationale-based \sftself\ and \sftteacher\ retain 82.6--90.5\% on average across methods and domains. Appendix~\ref{app:training-dynamics} illustrates this with training dynamics on Qwen3-4B Science, where RLVR's \csa\ predictions remain aligned with its current problem-solving ability, whereas \sftlabel\ increasingly fits the original labels despite its declining ability. 

Furthermore, Table~\ref{tab:rvs} shows that this advantage persists when supervised training is modified to preserve capability, using LoRA, KL regularization, and task mixing. RLVR achieves the highest \MF\ for both model sizes on Math. The results suggest that RLVR's advantage is not solely attributable to the capability loss induced by standard SFT. See Appendix~\ref{app:training-baseline-results} for more details. 

A closer look at individual decisions helps explain this remaining advantage. SFT can closely match the overall self-solve rate without accurately identifying which queries the model can solve. In Figure~\ref{fig:csa_scaling}, \sftteacher\ predicts a \selfsolve\ rate of 82.4\%, close to its true rate of 81.8\%, yet decision accuracy is only 79.0\% for \sftteacher\ and 76.3\% for \sftself, compared with 87.8\% for RLVR. A policy can match the overall rate yet choose the wrong queries to solve or delegate. RLVR's higher decision accuracy indicates better discrimination between these queries, beyond matching how often to self-solve.

\setlength{\panelheight}{0.28\textwidth}
\begin{figure*}[t]
\centering
\begin{minipage}[t][\panelheight][s]{0.48\textwidth}
\vspace*{0pt}
\captionof{table}{RLVR's advantage persists over capability-preserving training baselines on Math.
Columns report \MF\ and \CR\ for Qwen3-1.7B and 4B, averaged over
three seeds. Full Math and Science results appear in
Appendix~\ref{app:training-baseline-results}.}
\vspace{-10pt}
\label{tab:rvs}
\small
\setlength{\tabcolsep}{2.5pt}
\renewcommand{\arraystretch}{1.0}
\vfill
\begin{tabular*}{\linewidth}{@{\extracolsep{\fill}}lrrrr@{}}
\toprule
& \multicolumn{2}{c}{1.7B} & \multicolumn{2}{c}{4B} \\
\cmidrule(lr){2-3}\cmidrule(lr){4-5}
Method & \MF & \CR & \MF & \CR \\
\midrule
Base & 0.53 & 100.0 & 0.51 & 100.0 \\
SFT with smaller lr & 0.64 & 72.4 & 0.67 & 97.3 \\
LoRA SFT & 0.73 & 99.0 & 0.62 & 100.1 \\
KL SFT & 0.63 & 90.6 & 0.66 & 97.5 \\
Task-mixed SFT & 0.63 & 94.6 & 0.75 & 100.2 \\
DFW-matched SFT & 0.69 & 72.2 & 0.71 & 97.4\\
Query Probe & 0.70 & 100.0 & 0.71 & 100.0 \\
RLVR & \textbf{0.84} & 100.3 & \textbf{0.85} & 102.8 \\
\bottomrule
\end{tabular*}
\end{minipage}\hfill
\begin{minipage}[t][\panelheight][s]{0.48\textwidth}
\vspace*{0pt}
\centering
\includegraphics[width=\linewidth]{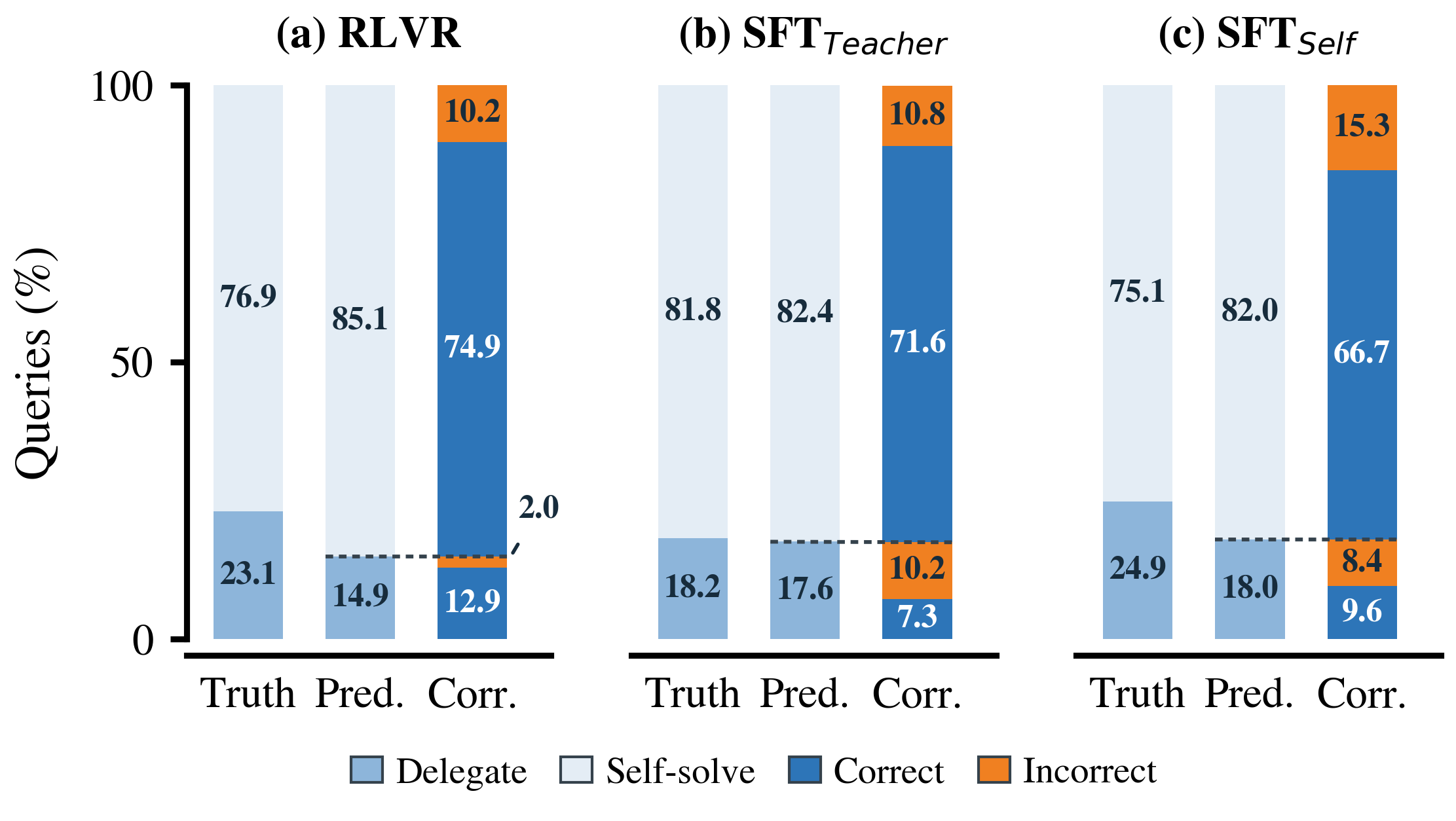}
\vfill
\vspace{0.5pt}
\captionof{figure}{Qwen3-4B on Math: (a) RLVR,
(b) \sftteacher, (c) \sftself. Pred.: predicted;
Corr.: correctness. The two regions of Corr. correspond to predicted
\delegate\ (bottom) and \selfsolve\ (top), each split into correct
and incorrect decisions. 
}
\label{fig:csa_scaling}
\end{minipage}
\end{figure*}

\paragraph{RLVR vs. training-free baselines.}
In Figure~\ref{fig:training-free-efficiency}, \csa-based RLVR outperforms both training-free query-based methods and methods that require generated answers on Qwen3-4B Math. It achieves \MF\ of 0.80, compared with at most 0.57 for query-based baselines, 0.76 for minimum token probability from one answer, and 0.78 and 0.75 for self-consistency and semantic entropy using five answers. Appendix~\ref{app:training-free-baselines} provides the full baseline setup and more experimental results. Compared with competitive answer-based methods, \csa\ avoids generating full task solutions before deciding whether to solve or delegate, using about 151 rationale and decision tokens per query in this setting (Appendix~\ref{app:csa-efficiency}). A decision-first format can further reduce token use by stopping as soon as the \selfsolve\ or \delegate\ decision is emitted. Appendix~\ref{app:csa-output-order-ablation} compares this format with analysis-first output across three settings and finds similar \csa\ performance.

\subsection{Capability Retention Analysis}

\textbf{Why retention matters.} We could assess the solver with a separate adapter or probe, leaving its solving ability untouched. But these methods are less accurate: they score 0.62--0.73 \MF\ on Math and 0.51–0.56 on Science, versus 0.84–0.85 and 0.66–0.67 for RLVR (Tables~\ref{tab:rvs} and ~\ref{tab:rvs-full}). LoRA SFT also performs worse than full-parameter SFT in three of four settings, suggesting insufficient capacity for this task. The best assessment in our experiments comes from changing the solver’s own weights, so preserving its solving ability matters. This also fits practical efficiency, where one updated checkpoint handles both solving and assessment. See Appendix~\ref{app:training-baseline-results} for more.

The \CR\ results in Tables~\ref{tab:main} and~\ref{tab:rvs} show that RLVR largely preserves aggregate task performance, but this can hide changes in which individual queries the model can solve. We therefore compare query-level capability before and after \csa\ training across eight mathematics and coding settings. As shown in Figure~\ref{fig:capability-stability}, on average 4.5\% of capability labels change, compared with 5.4\% disagreement expected from repeated finite-sample measurements of the same capability. Agreement between the pre- and post-training labels remains high ($\kappa=0.895$), while the mean Capability Ratio is 101.0\%. These results suggest that \csa\ training preserves both aggregate performance and the model’s query-level capability boundary within the resolution of our evaluation. Full details are provided in Appendix~\ref{app:capability-stability}.

\begin{figure}[!t]
\centering\small
\begin{minipage}[b]{0.49\textwidth}
\vspace{0pt}\centering\small
\begin{minipage}[c][0.95\linewidth][c]{\linewidth}
\centering
\makebox[\linewidth][c]{\hspace*{-32pt}\includegraphics[width=\linewidth]{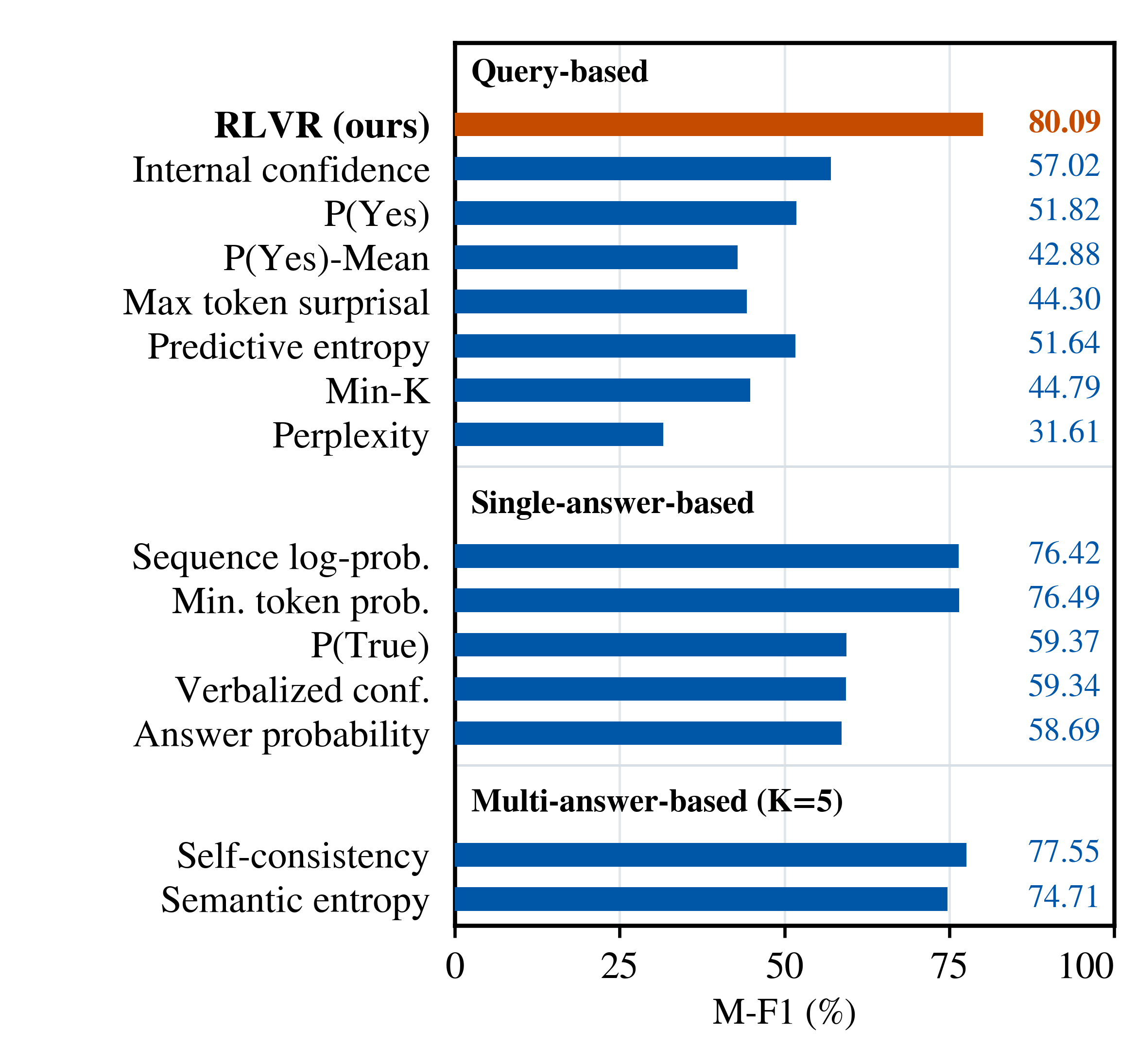}}
\end{minipage}
\captionsetup{skip=4pt}
\captionof{figure}{\MF{} on Qwen3-4B Math. \textbf{Query-based} methods assess capability from the query without generating a task answer. \textbf{Single-answer-based} methods use a single answer and its likelihood or confidence. \textbf{Multi-answer-based} methods use agreement or entropy across five sampled answers.}
\label{fig:training-free-efficiency}
\end{minipage}\hfill
\begin{minipage}[b]{0.49\textwidth}
\vspace{0pt}\centering\small
\begin{minipage}[c][0.95\linewidth][c]{\linewidth}
\centering
\includegraphics[width=\linewidth]{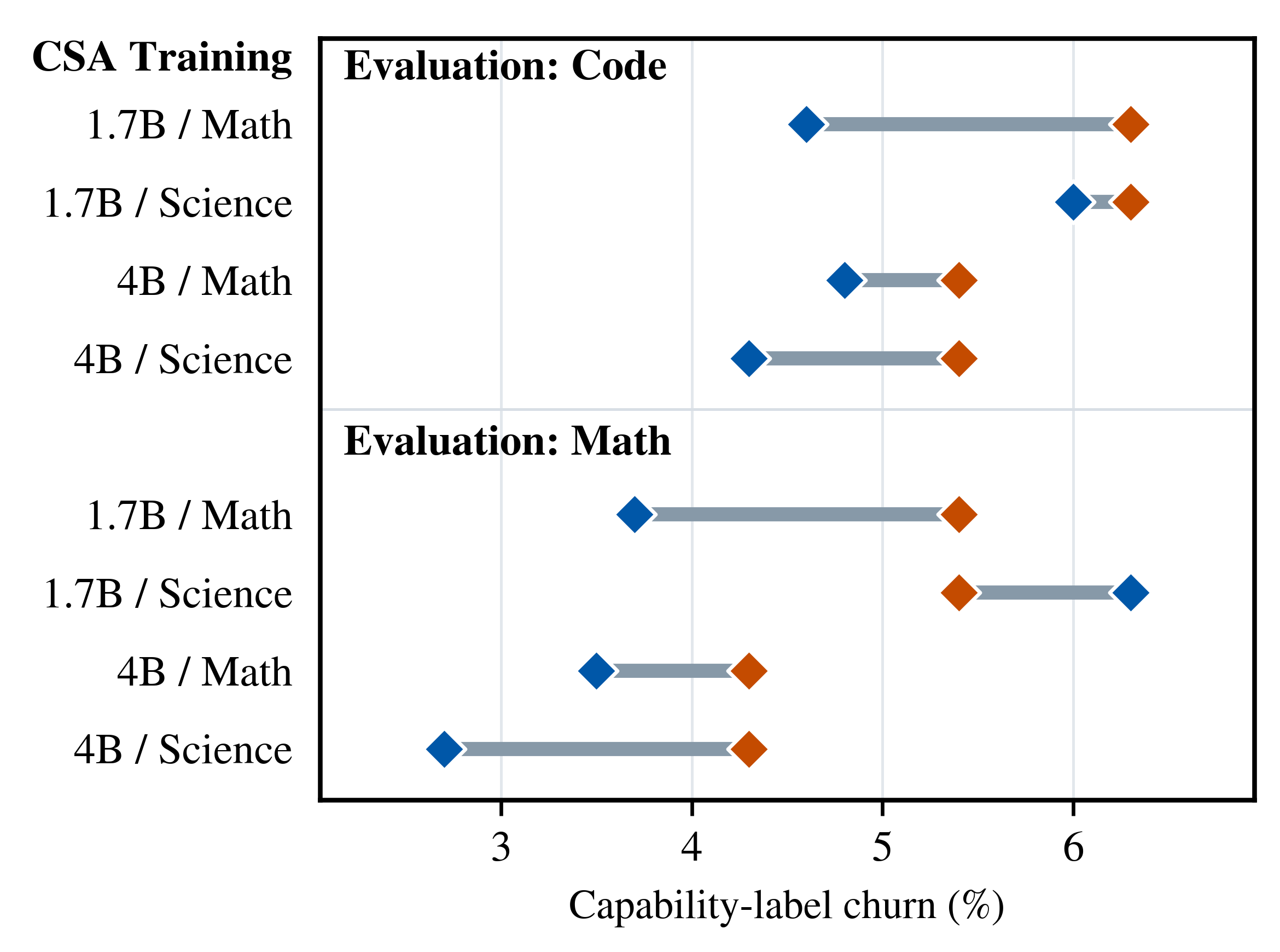}
\end{minipage}
\captionsetup{skip=4pt}
\captionof{figure}{\rlvr\ demonstrates query-level capability stability. Connected diamonds compare observed label churn (blue) with sampling noise (orange). Groups identify the evaluation domain; rows give model size / \csa\ training domain. Mean churn is $4.5\%$, versus $5.4\%$ sampling noise.}
\label{fig:capability-stability}
\end{minipage}
\end{figure}

\setlength{\panelheight}{0.35\textwidth}
\begin{figure}[!b]
\centering\small
\begin{minipage}[t]{0.49\textwidth}
  \vspace{0pt}\centering
  \begin{minipage}[c][\panelheight][c]{\linewidth}
    \centering
    \makebox[\linewidth][c]{\hspace*{-14pt}\includegraphics[width=\linewidth]{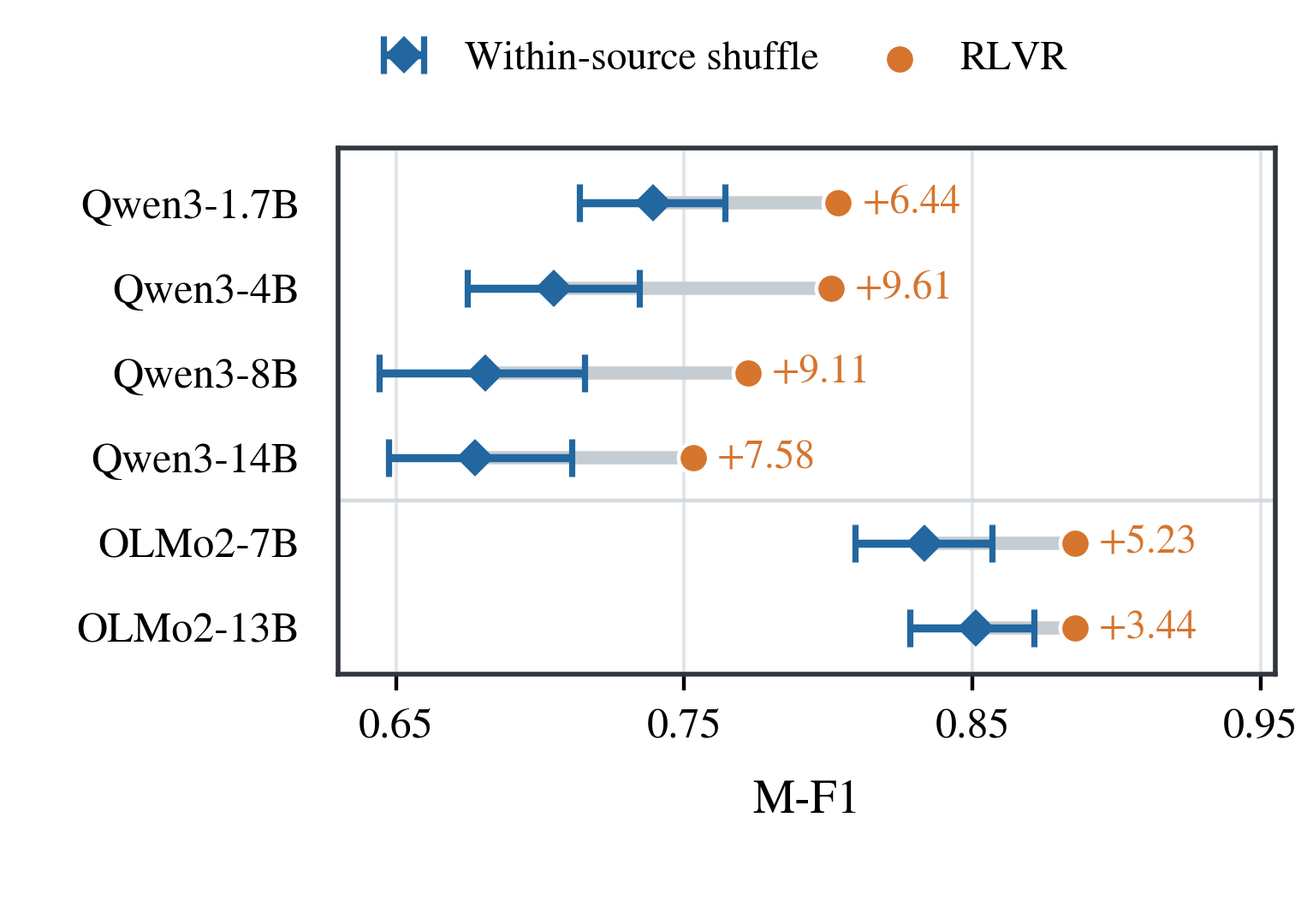}}
  \end{minipage}\par
  \begin{minipage}[t][30pt][t]{\linewidth}
    \caption{Math \MF\ versus within-source shuffling: 95\% null intervals and gains in percentage points.}
    \label{fig:difficulty-routing}
  \end{minipage}
\end{minipage}\hfill
\begin{minipage}[t]{0.49\textwidth}
  \vspace{0pt}\centering
  \begin{minipage}[c][\panelheight][c]{\linewidth}
    \centering
    \includegraphics[width=\linewidth]{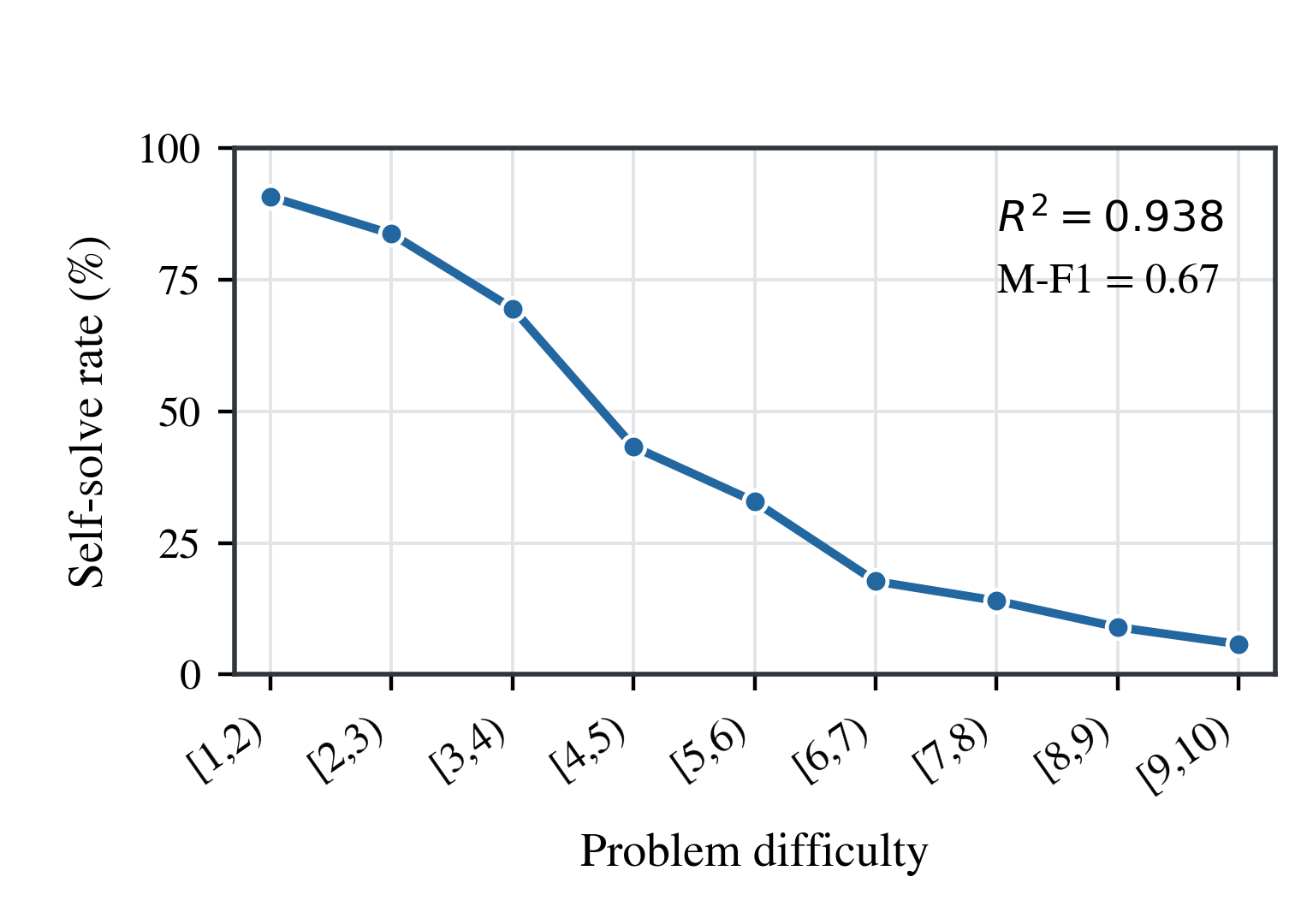}
  \end{minipage}\par
  \begin{minipage}[t][30pt][t]{\linewidth}
    \caption{Qwen3-4B-Math on Omni-MATH: self-solve rate decreases with problem difficulty.
    }
    \label{fig:omnimath-difficulty}
  \end{minipage}
\end{minipage}
\vspace{-14pt}
\end{figure}

\subsection{Generalization of \csa}
\label{sec:generalization}
\label{sec:benchmark-identity}

\paragraph{Within-Source Discrimination}
A Math policy could exploit benchmark source identity as a shortcut instead of assessing its ability to solve each query. To test this explanation, we shuffle decisions within GSM8K, MATH500, and AIME while preserving each source's routing counts and the model's post-training capability labels. This preserves any advantage from source-level routing but removes the association between individual decisions and query solvability. Figure~\ref{fig:difficulty-routing} shows that all six models exceed their null means by 3.44--9.61 \MF\ percentage points.
Thus, learned \csa\ captures query-level differences in solvability within benchmark sources, beyond exploiting benchmark identity. Appendix~\ref{app:within-source-math} gives the experimental protocol and permutation details.

\paragraph{Math--Science Transfer}

We assess out-of-distribution generalization by evaluating math-trained policies on science problems and science-trained policies on math problems. Table~\ref{tab:cross-domain-compact} shows that RLVR achieves the highest \MF\ in three of four cross-domain settings, including both transfer directions for Qwen3-1.7B and Science-to-Math transfer for Qwen3-4B. These results show that its assessment advantage extends beyond the training domain.

\paragraph{Generalization to Omni-MATH}

We further evaluate on Omni-MATH~\citep{gao2025omni}, a held-out mathematics dataset with annotated difficulty levels. Evaluating within a single benchmark lets us examine whether \csa\ responds to difficulty without relying on differences between benchmark sources. We apply the Qwen3-4B \csa\ policy trained on Math. Figure~\ref{fig:omnimath-difficulty} shows that its \selfsolve\ rate decreases from 90.7\% in the easiest interval to 5.8\% in the hardest, with a strong linear association between self-solve rate and difficulty ($R^2=0.938$). The policy also achieves query-level \MF\ of 0.67. Together, these results show that learned \csa\ transfers to unseen Math queries. It also adjusts its willingness to solve with difficulty while retaining query-level discrimination.

\newlength{\pairedtableheight}
\setlength{\pairedtableheight}{0.35\textwidth} %
\begin{table}[t]
\centering

\begin{minipage}[t][\pairedtableheight][t]{0.49\textwidth}
\vspace{0pt}
\centering
\small
\captionof{table}{\textbf{\rlvr\ shows better cross-domain transfer between science and math.}
Each entry reports \MF. Models are trained on one domain and evaluated on another (e.g., Math $\rightarrow$ Science: trained on Math, evaluated on Science).
}
\label{tab:cross-domain-compact}
\vfill
\renewcommand{\arraystretch}{1.15}
\begin{adjustbox}{width=\linewidth,valign=b}
\begin{tabular}{@{}lrrrr@{}}
\toprule
Model & \sftlabel & \sftself & \sftteacher & RLVR \\
\midrule
\multicolumn{5}{@{}l}{\textit{Science $\rightarrow$ Math}} \\
Qwen3-1.7B & 0.68 & 0.66 & \underline{0.73} & \textbf{0.79} \\
Qwen3-4B   & 0.75 & 0.67 & \underline{0.75} & \textbf{0.78} \\
\midrule
\multicolumn{5}{@{}l}{\textit{Math $\rightarrow$ Science}} \\
Qwen3-1.7B & 0.40 & \underline{0.53} & 0.51 & \textbf{0.63} \\
Qwen3-4B   & 0.39 & \textbf{0.53} & \underline{0.52} & 0.51 \\
\bottomrule
\end{tabular}
\end{adjustbox}
\end{minipage}
\hfill
\begin{minipage}[t][\pairedtableheight][t]{0.49\textwidth}
\vspace{0pt}
\centering
\small
\captionof{table}{\textbf{\rlvr\ shows zero-shot transfer to coding.}
Policies trained on mathematics or science are evaluated without
further training on  \(722\)
Python problems.
Pass@1 measures coding ability, SSR the self-solve rate, and \MF{}
assessment quality. Results are averaged over three decoding seeds.
}
\vspace{-1pt}
\label{tab:code-transfer-main}
\vfill
\qquad
\begin{adjustbox}{max width=\linewidth,valign=b}
\begin{tabular}{@{}llccc@{}}
\toprule
Model & \csa\ training & Pass@1 & SSR & \MF \\
\midrule
\multirow{3}{*}{Qwen3-1.7B}
& None    & 0.56 & 1.00 & 0.36 \\
& Math    & 0.58 & 0.83 & 0.64 \\
& Science & 0.57 & 0.87 & 0.58 \\
\midrule
\multirow{3}{*}{Qwen3-4B}
& None    & 0.68 & 1.00 & 0.41 \\
& Math    & 0.68 & 0.92 & 0.57 \\
& Science & 0.68 & 0.89 & 0.63 \\
\bottomrule
\end{tabular}
\end{adjustbox}
\end{minipage}
\vspace{5pt}
\end{table}

\paragraph{Transfer to Coding}

Table~\ref{tab:code-transfer-main} tests a larger distribution shift:
policies trained only on mathematics or science are evaluated on
\(722\) Python programming problems. For
Qwen3-1.7B, the best transferred policy raises coding-assessment
\MF{} from \(0.36\) to \(0.64\). For Qwen3-4B, it rises from
\(0.41\) to \(0.63\). Coding ability is preserved, with pass@1
changing from \(0.562\) to \(0.566\)--\(0.578\) for the 1.7B model
and from \(0.675\) to \(0.676\) for the 4B model. Together, these
results show that \csa\ transfers to a previously unseen domain and task format. Full details are provided in Appendix~\ref{app:code-transfer}.

\subsection{\csa\ Tracks Model-Specific Capability}
\label{sec:model-specific}
The alignment between \csa\ decisions and problem difficulty does not by itself establish model-specific self-assessment. A policy could learn which problems are generally difficult for language models. We test this alternative explanation on \(722\) coding problems using OLMoE-1B-7B~\citep{muennighoff2025olmoe} as an independent
comparator. For each Qwen policy's own model and OLMoE, we estimate per-problem competence in the same way by sampling \(K=32\) solutions.

Table~\ref{tab:model-specific-main} shows that \csa\ decisions track the policy model's own competence more strongly than OLMoE's competence. After controlling for OLMoE's pass rate, the partial correlation between the \csa\ decision and the policy model's own pass rate ranges from \(0.224\) to \(0.322\). In the reverse direction, after controlling for the policy model's own pass rate, the partial correlation with OLMoE ranges from only \(0.042\) to \(0.093\).

\begin{wraptable}{r}{0.58\textwidth}
\vspace{-11pt}
\centering\small
\caption{
\textbf{\csa\ tracks model-specific competence.}
Partial correlations $r_{x\mid y}=r(\mathrm{dec},x\mid y)$ and \MF\ versus an OLMoE-based rule at matched delegation rates. Pass rates use $K=32$ attempts per query.
}
\label{tab:model-specific-main}
\setlength{\tabcolsep}{4pt}
\renewcommand{\arraystretch}{1.05}
\begin{adjustbox}{max width=\linewidth}
\begin{tabular}{@{}llcccc@{}}
\toprule
Model
& Training
& \(r_{\mathrm{own}\mid\mathrm{OLMoE}}\)
& \(r_{\mathrm{OLMoE}\mid\mathrm{own}}\)
& \shortstack{M-F1\\\csa}
& \shortstack{M-F1\\OLMoE rule} \\
\midrule
\multirow{2}{*}{Qwen3-1.7B}
& Math & 0.256 & 0.093 & \textbf{0.639} & 0.576 \\
& Science & 0.224 & 0.067 & \textbf{0.580} & 0.521 \\
\midrule
\multirow{2}{*}{Qwen3-4B}
& Math & 0.280 & 0.042 & \textbf{0.571} & 0.522 \\
& Science & 0.322 & 0.063 & \textbf{0.629} & 0.576 \\
\bottomrule
\end{tabular}
\end{adjustbox}
\vspace{-5pt}
\end{wraptable}

We also construct an OLMoE-based difficulty rule that delegates exactly as many problems as each \csa\ policy, selecting those with the lowest OLMoE pass rates. \csa\ achieves \(0.05\)--\(0.06\) higher Macro-F1 than this rule
across all four settings, even though \csa\ observes only the problem text while the OLMoE rule uses \(32\) executed solutions per problem. Thus, the learned policy captures information about which problems its own model can solve beyond a generic estimate of problem difficulty.

Cross-scale policy swapping provides a complementary test. A policy trained for Qwen3-1.7B nearly matches that solver's aggregate capability, with a coverage--accuracy gap of \(+0.003\), whereas attaching a policy trained for
the stronger Qwen3-8B model to the 1.7B solver produces an \(+0.110\) gap.
Across all pairings, policies matched to their solver have a smaller mean absolute coverage--accuracy gap than mismatched policies (\(0.049\) versus \(0.072\)). While these cross-scale results describe aggregate alignment, the query-level comparisons provide more direct evidence of model-specific assessment. Full results are reported in Appendix~\ref{app:model-specific}.

\setlength{\parskip}{3pt}
\begin{table}[!t]
\centering
\begin{minipage}[t]{0.48\linewidth}
\vspace{0pt}\centering
  \begin{minipage}[t][43pt][t]{\linewidth}
  \caption{Math routing to Qwen3-235B-A22B.}
  \label{tab:routing-app}
  \end{minipage}\par\vspace{-31pt}
  \footnotesize
  \setlength{\tabcolsep}{1pt}
  \renewcommand{\arraystretch}{0.95}
  \begin{tabular*}{\linewidth}{@{}l@{\extracolsep{\fill}}cc@{}}
    \toprule
    Method & \textbf{Qwen3-1.7B} & \textbf{Qwen3-4B} \\
    \midrule
    \multicolumn{3}{c}{\textbf{Routing Accuracy}} \\
    Local only & 0.60 & 0.69 \\
    Cloud only & 0.86 & 0.86 \\
    Vanilla & 0.64 & 0.70 \\
    Random & 0.65 & 0.72 \\
    \textbf{CSA} & 0.73 & 0.80 \\
    \midrule
    \multicolumn{3}{c}{\textbf{Routing Overhead}} \\
    CSA cloud calls & 96 & 73 \\
    Time (s) / Tokens & 0.53 / 304.9 & 0.39 / 150.7 \\
    \bottomrule
  \end{tabular*}
\end{minipage}\hfill
\begin{minipage}[t]{0.50\linewidth}
  \vspace{0pt}\centering
  \begin{minipage}[t][43pt][t]{\linewidth}
  \caption{Data selection results on \textsc{MedMCQA}.}
  \label{tab:science-data-selection-results}
  \end{minipage}\par
  \vspace{-31pt}
  \footnotesize
  \setlength{\tabcolsep}{2pt}
  \renewcommand{\arraystretch}{0.9}
    \begin{tabular*}{\linewidth}{@{}l@{\extracolsep{\fill}}cccccc@{}}
      \toprule
      & \multicolumn{3}{c}{\textbf{Qwen3-1.7B}}
      & \multicolumn{3}{c}{\textbf{Qwen3-4B}} \\
      \cmidrule(lr){2-4} \cmidrule(lr){5-7}
      Method & 5\% & 10\% & 20\% & 5\% & 10\% & 20\% \\
      \midrule
      \textit{Eval only}
        & \multicolumn{3}{c}{\rule[0.5ex]{0.8em}{0.4pt}\ \textit{0.40}\ \rule[0.5ex]{0.8em}{0.4pt}}
        & \multicolumn{3}{c}{\rule[0.5ex]{0.8em}{0.4pt}\ \textit{0.42}\ \rule[0.5ex]{0.8em}{0.4pt}} \\
      \textit{Train on all}
        & \multicolumn{3}{c}{\rule[0.5ex]{0.8em}{0.4pt}\ \textit{0.44}\ \rule[0.5ex]{0.8em}{0.4pt}}
        & \multicolumn{3}{c}{\rule[0.5ex]{0.8em}{0.4pt}\ \textit{0.48}\ \rule[0.5ex]{0.8em}{0.4pt}} \\
      \midrule
      Random         & 0.40 & 0.45 & 0.44 & 0.43 & 0.43 & 0.44 \\
      Random-band & 0.41 & 0.45 & 0.44 & 0.44 & 0.44 & 0.41 \\
      Dev Similarity & 0.40 & 0.39 & 0.40 & 0.43 & 0.45 & 0.34 \\
      High Loss      & 0.27 & 0.26 & 0.27 & 0.24 & 0.24 & 0.25 \\
      Low Loss       & 0.41 & 0.43 & 0.42 & 0.49 & 0.51 & 0.49 \\
      \textbf{CSA-Based} & \textbf{0.43} & \textbf{0.45} & \textbf{0.45} & \textbf{0.57} & \textbf{0.56} & \textbf{0.56} \\
      \bottomrule
    \end{tabular*}
\end{minipage}
\end{table}

\subsection{Applications}
\label{sec:applications}

\subsubsection{Self-Routing}\label{sec:app-routing}
\looseness=-1
A model with accurate CSA is a natural router in a local--cloud inference system. When the local model chooses to self-solve, the query is answered on-device. When it chooses to delegate, the query is routed to a powerful cloud model. The local model serves as its own router, offering a practical cost--accuracy trade-off between local-only and cloud-only inference. Table~\ref{tab:routing-app} presents the Math routing results using Qwen3-1.7B and Qwen3-4B as local models and Qwen3-235B-A22B as the cloud model. \textbf{Accuracy.} CSA routing reaches 0.73 and 0.80 accuracy for the 1.7B and 4B models, outperforming local-only and vanilla routing and improving over random routing by about 8.2 percentage points at matched cloud-call budgets. \textbf{Efficiency.} CSA routing is fast, taking an average of 0.53 seconds per query for Qwen3-1.7B and 0.39 seconds for Qwen3-4B. This enables prompt selection between local answering and cloud delegation with little delay from the routing decision. Appendix~\ref{app:self-routing} provides a detailed illustration and evaluation setup. Appendix~\ref{sec:ca-routing} further studies single-attempt routing with an expected-outcome reward that accounts for both answer accuracy and delegation cost.

\subsubsection{Self-Guided Data Selection}
\label{sec:app-data-selection}

\looseness=-1
We further test whether CSA can serve as a data-selection signal. Starting from a base model $M_0$, we train an RL-based CSA model $M_{\mathrm{RLVR}}$ on MMLU-Pro~\citep{wang2024mmluprorobustchallengingmultitask} and apply it to a fresh candidate pool from MedMCQA~\citep{pal2022medmcqa}. We select examples based on a metric which prioritizes examples that the CSA model is likely to \textsc{Delegate} while filtering for cases that are uncertain but not extreme loss outliers, and fine-tune the model on this subset. We then compare against standard selection baselines, including random, loss-based, and dev-similarity selection, to assess whether CSA identifies examples that more effectively improve the model's downstream performance.

Table~\ref{tab:science-data-selection-results} shows that CSA-based selection generally outperforms the baselines across model sizes and selection budgets. For Qwen3-1.7B, CSA selection improves over the eval-only model at every budget and slightly exceeds training on the full 50K candidate pool when using only 10--20\% of the data. The gains are especially pronounced for Qwen3-4B. CSA-selected subsets achieve 55.8--57.4\% accuracy, substantially outperforming both the eval-only baseline (42.2\%) and training on all examples (47.6\%). In contrast, high-loss selection performs poorly across all settings, indicating that simply selecting the hardest examples is not sufficient and can over-emphasize noisy or unlearnable samples. Low-loss selection is the strongest baseline for Qwen3-4B, but remains well below CSA-based selection. These results suggest that the learned CSA signal identifies examples that are not merely difficult or similar to the development set, but are especially useful for improving the model under a limited fine-tuning budget. See Appendix~\ref{app:data-selection} for the full details.

\enlargethispage{5pt}
\section{Conclusion}
\looseness=-1
We studied how LLMs can acquire \csa\ while retaining the solving ability being assessed. Through a systematic study, we showed \rlvr\ training provides the strongest overall assessment–retention trade-off among the evaluated training strategies. We also demonstrated out-of-distribution generalization of \csa. More broadly, \csa\ shifts self-knowledge from a safeguard against errors to a signal for adaptation, guiding both when a model acts and what it learns next.
We believe \csa\ is a promising building block for more adaptive, trustworthy, and responsible AI systems. Ultimately, intelligence is not only about solving problems, but also about recognizing the limits of one's own competence, and adapting when needed.

\par
\endgroup

\clearpage
 
\bibliography{iclr2027_conference}
\bibliographystyle{iclr2027_conference}

\newpage
\appendix

\newpage

\noindent{\Large\bfseries Appendix Contents}\par
\vspace{2em}

\begingroup
\hypersetup{linkcolor=black}
\setlength{\parindent}{0pt}
\setlength{\parskip}{0.25em}

\newcommand{\tocSec}[2]{%
  \noindent\textbf{\hyperref[#1]{#2}}\dotfill\pageref{#1}\par%
}
\newcommand{\tocSub}[2]{%
  \noindent\hspace{1.8em}\hyperref[#1]{#2}\dotfill\pageref{#1}\par%
}
\newcommand{\tocSubSub}[2]{%
  \noindent\hspace{3.6em}\hyperref[#1]{#2}\dotfill\pageref{#1}\par%
}

\vspace{0.3em}
\tocSec{sec:rel-work-full}{\ref{sec:rel-work-full}\quad Related Work}
\tocSub{app:rel-self-eval}{\ref{app:rel-self-eval}\quad Self-Evaluation and Calibration of Correctness in LLMs}
\tocSub{app:rel-selective}{\ref{app:rel-selective}\quad Selective Prediction, Abstention, and Risk-Coverage Evaluation}
\tocSub{app:rel-hallucination}{\ref{app:rel-hallucination}\quad Hallucination Detection}
\tocSub{app:rel-conformal}{\ref{app:rel-conformal}\quad Conformal Prediction and Statistical Risk Control for LLMs}
\tocSub{app:rel-rl}{\ref{app:rel-rl}\quad Learning Decision Policies with RL}
\tocSub{app:rel-data}{\ref{app:rel-data}\quad Data Selection and Curriculum Learning}
\tocSub{app:rel-adjacent}{\ref{app:rel-adjacent}\quad Relation to Adjacent Formulations}

\vspace{0.3em}
\tocSec{app:algorithm}{\ref{app:algorithm}\quad RLVR Training Algorithm}

\vspace{0.3em}
\tocSec{app:setup}{\ref{app:setup}\quad Experimental Setup}
\tocSub{app:main-setup}{\ref{app:main-setup}\quad Main Experimental Setup}
\tocSubSub{app:models}{\ref{app:models}\quad Models}
\tocSubSub{app:datasets}{\ref{app:datasets}\quad Datasets}
\tocSubSub{app:csa-inference_setup}{\ref{app:csa-inference_setup}\quad \csa\ Inference Setup}
\tocSubSub{app:label}{\ref{app:label}\quad Label Construction}
\tocSubSub{app:metrics}{\ref{app:metrics}\quad Evaluation Metrics}
\tocSub{app:training}{\ref{app:training}\quad Training Settings}
\tocSub{app:training-baseline-setup}{\ref{app:training-baseline-setup}\quad Capability-Preserving Training Baseline Setup}

\vspace{0.3em}
\tocSec{app:more-results}{\ref{app:more-results}\quad Additional Results}
\tocSub{app:prompting-study}{\ref{app:prompting-study}\quad Prompt-Only \csa\ Study}
\tocSub{app:ablation-studies}{\ref{app:ablation-studies}\quad Ablation Studies}
\tocSubSub{app:label-ablations}{\ref{app:label-ablations}\quad Capability-Label Construction}
\tocSubSub{app:rollout-warmup}{\ref{app:rollout-warmup}\quad Diversity-Filtered Warm-up}
\tocSubSub{app:csa-output-order-ablation}{\ref{app:csa-output-order-ablation}\quad \csa\ Output Order}
\tocSub{app:training-dynamics}{\ref{app:training-dynamics}\quad Training Dynamics of SFT and RLVR}
\tocSub{app:training-baseline-results}{\ref{app:training-baseline-results}\quad Capability-Preserving Training Baselines}
\tocSub{app:training-free-baselines}{\ref{app:training-free-baselines}\quad Training-Free Baselines}
\tocSub{app:csa-efficiency}{\ref{app:csa-efficiency}\quad \csa\ Inference Efficiency}
\tocSub{app:capability-stability}{\ref{app:capability-stability}\quad Query-Level Capability Stability}
\tocSub{app:within-source-math}{\ref{app:within-source-math}\quad Within-Source Discrimination on Math}
\tocSub{app:code-transfer}{\ref{app:code-transfer}\quad Zero-Shot Transfer to Coding}
\tocSub{app:model-specific}{\ref{app:model-specific}\quad \csa\ Is Model Specific}
\tocSub{app:case-study-full}{\ref{app:case-study-full}\quad Case Studies}
\tocSub{app:applications}{\ref{app:applications}\quad Applications of \csa}
\tocSubSub{app:self-routing}{\ref{app:self-routing}\quad Self-Routing}
\tocSubSub{sec:ca-routing}{\ref{sec:ca-routing}\quad Cost-Aware Routing}
\tocSubSub{app:data-selection}{\ref{app:data-selection}\quad Self-Guided Data Selection}

\vspace{0.3em}
\tocSec{app:prompts}{\ref{app:prompts}\quad Prompt Templates}
\tocSub{app:csa_inference_template}{\ref{app:csa_inference_template}\quad \csa\ Prompt Template}
\tocSub{app:self-rationale-prompt}{\ref{app:self-rationale-prompt}\quad Self-Rationale Prompt (\sftself)}
\tocSub{app:teacher-rationale-prompt}{\ref{app:teacher-rationale-prompt}\quad Teacher-Rationale Prompt (\sftteacher)}

\endgroup

\newpage

\section{Related Work}\label{sec:rel-work-full}

Our work draws on research on self-evaluation and calibration, selective prediction and abstention, uncertainty estimation, learning to defer, and decision-making with external tools. These areas have shown that LLMs can expose useful signals about correctness and answerability and can use such signals to abstain or seek assistance. We focus on a particular coupling within this broader literature: the same generative model is trained to assess its own query-level success before answering, even though this training may alter the capability being assessed. We therefore study self-assessment quality together with capability retention and evaluate whether the resulting behavior transfers beyond its training distribution.

\subsection{Self-Evaluation and Calibration of Correctness in LLMs}
\label{app:rel-self-eval}
\citet{kadavath2022llms-mostly-know} study whether models can estimate the correctness of their own outputs through quantities such as P(True) and P(IK). That work shows that larger models can exhibit useful self-evaluation signals when queried in the right format, and it helped establish a widely used framing of LLM self-knowledge as a calibration~\citep{gneiting2007calibration1,guo2017calibration2} problem. Follow-up work has stressed both the promise and limits of this perspective. \citet{yin2023do-llms-know-what-they-dont-know} ask whether LLMs know what they do not know, focusing on unanswerable questions and explicit uncertainty recognition, while \citet{kapoor2024llms-must-be-taugh} argue that prompting alone is not enough and show that uncertainty estimation benefits substantially from explicit fine-tuning. \citet{zhang2024r} show models can be instruction-tuned to say I don't know. \citet{barkan2026do} study whether LLMs can predict their own task success and find all tested LLMs are overconfident. \citet{damani2026beyond} jointly train for answer correctness and calibrated confidence.
Together, these studies show that uncertainty awareness is partly learnable and can be expressed through scores, verbal judgments, or abstention behavior.

A large practical literature studies how to extract usable confidence from aligned systems~\citep{chuang2025learning}. \citet{tian2023just-ask-for-calibration} show that for RLHF-tuned models, verbalized confidences (numbers or linguistic markers produced as tokens) can be more calibrated than raw conditional probabilities on several QA benchmarks, and they catalog prompting/temperature-scaling strategies for eliciting confidence.  \citet{mielke2022reducing-overconfience} propose linguistic calibration for dialogue agents to better align the style of uncertainty expression with correctness, highlighting that calibration is not only numeric but also communicative.

A parallel line scrutinizes verbalized uncertainty and its failure modes. \citet{sun2025large} find five LLMs overestimate the probability that their answer is correct by 20--60\% on reasoning problems. \citet{chhikara2025mind} analyze overconfidence as mismatch between predicted confidence and true correctness across QA datasets. \citet{xiong2023can-llm-express-uncertainty} show that LLMs can be systematically overconfident when asked to verbalize confidence, though calibration and error prediction improve with scale and with specific elicitation strategies (e.g., consistency-based approaches).  \citet{groot2024overconfidence-is-key} further evaluate verbalized uncertainty across LLMs/VLMs and emphasize overconfidence as a recurring concern.  \citet{yang2024verbalized} synthesize why empirical findings about verbalized confidence disagree across papers (prompting styles, evaluation setups), reinforcing the need for decision-centric evaluation rather than relying on any single elicitation recipe.

These methods provide natural signals and baselines for \csa, since confidence estimates can be converted into answer-or-abstain decisions through thresholding. Our focus is therefore not a categorical distinction between scores and policies, but the shared-parameter setting in which the same model learns a pre-answer, model-specific decision while retaining the problem-solving ability being assessed. Cross-domain evaluation tests whether this behavior extends beyond its training distribution.

\subsection{Selective Prediction, Abstention, and Risk-Coverage Evaluation}
\label{app:rel-selective}
The classical foundation for abstention~\citep{liu2019deep-gamblers} is the reject option, often traced to \citet{chow1970optimum}, who derives optimal decision rules trading misclassification error against rejection cost.  Modern selective prediction~\citep{el2010foundations} formalizes this as optimizing a risk--coverage trade-off---achieving low error on the subset of instances the model elects to answer.

In deep learning, \citet{geifman2019selectivenet} treat selective classification as choosing a subset to minimize selective risk subject to a coverage constraint. The broader “learning with rejection” literature develops surrogate losses and theoretical analyses for reject option learning~\citep{bartlett2008classification}.

Within NLP specifically, \citet{xin2021art-of-abstetion} revisit selective prediction for language tasks, comparing confidence estimators and introducing regularization tricks to improve confidence without heavy compute.  \citet{gu2023evaluation} provide a methodological audit of selective prediction evaluation in NLP and propose additional metrics (e.g., refinement) to separate the quality of confidence scores from downstream accuracy.

For LLMs, a current synthesis is \citet{wen2025abstention-survey}, which organizes abstention methods by lifecycle stage (pretraining, alignment~\citep{alignment-for-honesty}, inference), and emphasizes that abstention must be evaluated not only technically but also relative to user values and deployment context.

\subsection{Hallucination Detection}
\label{app:rel-hallucination}
Hallucination detection addresses related failures in which a model produces unsupported or incorrect content, sometimes with unwarranted confidence~\citep{ji2023survey-hallu-1,huang2025survey-hallu-2,wang2026mitigatingfactualhallucinationlarge,wang2025damo,ji-etal-2024-llm,gekhman-etal-2024-fine}. A prominent black-box detection approach is SelfCheckGPT~\citep{manakul2023selfcheckgpt} which detects hallucinated sentences by sampling multiple outputs and measuring disagreement/contradiction—motivated by the intuition that knowledgeable generations are more self-consistent.  A complementary statistically grounded approach is semantic entropy: \citet{farquhar2024detecting-hallucinations-entropy} propose uncertainty estimators at the level of meaning, showing that entropy over semantic equivalence classes can detect a subset of hallucinations (confabulations) and generalize to unseen questions.  \citet{kossen2024semantic-entropy} then introduce semantic entropy probes designed to make such uncertainty signals cheaper while retaining detection quality.

Internal representations can carry correctness/answerability signals even when outputs are overconfident~\citep{binder2025looking}.
\citet{azaria2023internal} show that linear classifiers on hidden states can predict statement truthfulness better than sequence probability alone, suggesting that internal representations contain information relevant to output reliability.
\citet{slobodkin2023curious} focus on unanswerability: they find indications that models encode answerability, even when they still generate hallucinated answers for unanswerable questions.  Newer probe-based hallucination detectors~\citep{han2025simple-factuality-probes} push this toward practical deployment (e.g., linear probes for long-form generation) while reducing the cost relative to multi-sampling.  A related direction~\citep{cencerrado2025no-answer-needed} predicts answer correctness from question-only activations, indicating that some correctness signals may be present before decoding the full answer—useful for early deferral/abstention.

\subsection{Conformal Prediction and Statistical Risk Control for LLMs}
\label{app:rel-conformal}
Conformal prediction (CP)  methods usually provide post-hoc risk control (thresholding, filtering, prediction sets) rather than directly modifying the LLM's parameters.
CP provides distribution-free coverage guarantees under exchangeability assumptions~\citep{angelopoulos2023conformal,shafer2008tutorial-conformal,vovk2005algorithmic}.  Because exact conditional coverage is generally impossible in finite samples, modern work develops controlled relaxations and conditional targets; \citet{gibbs2025conformal} formalize a spectrum between marginal and conditional validity that is useful when risk differs by topic or subgroup.

Several recent works adapt CP to LMs and text. \citet{quach2023conformal} propose Conformal Language Modeling, producing sets/subsets of outputs (or filtered segments of text) with validity guarantees for language model generations.  \citet{kumar2023conformal} apply conformal prediction to multiple-choice QA with LLMs, arguing CP produces useful uncertainty sets correlated with accuracy.  Because many deployed LLMs are API-only, \citet{su2024api} propose CP methods that do not require access to logits, improving applicability in closed models.

For open-ended generation~\citep{campos2024conformal}, correctness-labeling and scoring functions become the bottleneck: CP guarantees depend on a nonconformity score that must detect errors in generated text. \citet{lee2024selective} address this by using textual entailment to define correctness of generated sequences and propose selective generation algorithms controlling false discovery rates (FDR) over entailment relations.  \citet{cherian2024large} propose enhanced conformal methods for validity guarantees on LLM outputs, explicitly noting issues of topic-dependent reliability and the utility loss from overly aggressive filtering; they propose conditional-conformal-inspired adaptations and methods to improve the scoring function.

\subsection{Learning Decision Policies with RL}
\label{app:rel-rl}
RL with verifiable rewards can optimize terminal decisions while allowing intermediate reasoning to emerge from interaction with the reward signal~\citep{guo2025deepseek-r1,shao2024deepseekmath}. A broader literature studies tool use and help-seeking through prompting, self-supervised training, supervised fine-tuning, and reinforcement learning~\citep{yao2022react,schick2023toolformer,nakano2021webgpt,huang2023metatool,xu2023toolbench,feng2025retool,du2024anytool,gul2025mash}. These methods typically learn whether an external resource improves downstream task utility. Our setting instead isolates prediction of the assessed model's own success and evaluates how learning this decision affects the model's underlying problem-solving ability. We use RL as one approach to optimizing this decision and compare it directly with supervised alternatives.

\subsection{Data Selection and Curriculum Learning}
\label{app:rel-data}
\citet{albalak2024survey} survey data selection for language models across goals such as quality, diversity, difficulty, and efficiency, and highlight that naively training on all available data may be suboptimal or infeasible~\citep{mindermann2022prioritizedtrainingpointslearnable}. A prominent practical method is DoReMi~\citep{xie2023doremi}, which optimizes domain mixture weights via distributionally robust optimization on a proxy model and then scales training on a reweighted mixture, improving efficiency and average downstream performance. More recently, data selection has been formulated as an optimal control problem over training dynamics~\citep{gu2024data-selection}, providing language to treat data scheduling as a sequential decision problem.
Classical curriculum learning shows that organizing examples by difficulty can improve optimization and generalization~\citep{bengio2009curriculum}.

Our application uses \csa{} scores as a model-specific signal for data selection. Examples receiving stronger delegation scores are likely to be difficult for the current model, while loss-based filtering removes examples that are already too easy or appear excessively difficult or noisy. This connects \csa{} to difficulty-aware data selection.

\subsection{Relation to Adjacent Formulations}
\label{app:rel-adjacent}
 Our distinction from related work concerns the joint setting in which the target is the assessed model's own success and self-assessment training may alter the capability being assessed. Moreover, unlike much prior work that treats self-knowledge as a safeguard against hallucination, using confidence or abstention to avoid unreliable answers, we explore the model's own capability boundary as a signal for adaptation. The same learned decision determines when the model acts or seeks help at inference time and which examples it learns from during training.

\section{RLVR Training Algorithm}
\label{app:algorithm}

We provide details and pseudocode for the two-stage GRPO procedure summarized in Section~\ref{sec:training}. The binary reward is $+1$ when a rollout predicts the ground-truth \csa\ label and $-1$ otherwise. Prompted base models often predict \selfsolve\ on nearly every query (Appendix~\ref{app:prompting-study}). Applying GRPO directly can therefore yield rollout groups with a single label, vanishing reward variance, and no useful policy-gradient signal.

\paragraph{Stage 1: Diversity-Filtered Warm-up (DFW).}
We construct a fixed diversified subset $\mathcal{D}_{\text{div}}\subseteq\mathcal{D}$ by retaining only queries whose $G$ rollouts from the initial policy $\pi_{\theta_0}$ contain both \selfsolve\ and \delegate. Under our decision reward, these screening groups contain both reward values. We then run $T_{\text{warm}}$ GRPO steps on fresh rollouts from $\mathcal{D}_{\text{div}}$, concentrating early training on queries that exhibited decision diversity and producing checkpoint $\theta_{\text{warm}}$.

DFW is related to DAPO's dynamic sampling~\citep{yu2025dapoopensourcellmreinforcement}, which filters uniformly rewarded rollout groups while constructing training batches. Both use within-query reward variation to identify useful training opportunities. DFW instead selects a fixed query subset using the initial policy, trains on fresh rollouts from that subset for a limited warm-up period, and then returns to the full dataset.

\paragraph{Stage 2: Full GRPO Training.}
Starting from $\theta_{\text{warm}}$, we train for another $T_{\text{full}}$ steps on the full dataset $\mathcal{D}$, where the now-informative advantages drive learning across both solvable and unsolvable queries. Algorithm~\ref{alg:grpo-crow} gives the complete procedure.

\begin{algorithm}
\caption{Two-Stage GRPO with Diversity-Filtered Warm-up.}
\label{alg:grpo-crow}
\small
\begin{algorithmic}[1]
\Require Dataset $\mathcal{D}=\{(x_i, y_i)\}_{i=1}^N$, initial policy $\pi_{\theta_0}$, reference policy $\pi_{\text{ref}}$, group size $G$, warm-up steps $T_{\text{warm}}$, full-phase steps $T_{\text{full}}$
\Ensure Trained policy $\pi_{\theta_{\text{final}}}$
\Statex
\State \textbf{// Stage 1: Diversity-Filtered Warm-up (DFW)}
\State $\mathcal{D}_{\text{div}} \gets \{ x_i \in \mathcal{D} : G\text{ rollouts from }\pi_{\theta_0}\text{ yield both }\textsc{Self-Solve}\text{ and }\textsc{Delegate}\}$ \Comment{filter on initial policy}
\State $\theta_{\text{warm}} \gets \textsc{Train-GRPO}(\theta_0, \mathcal{D}_{\text{div}}, T_{\text{warm}})$
\Statex
\State \textbf{// Stage 2: Full GRPO Training}
\State $\theta_{\text{final}} \gets \textsc{Train-GRPO}(\theta_{\text{warm}}, \mathcal{D}, T_{\text{full}})$ \Comment{continue on full data}
\State \Return $\pi_{\theta_{\text{final}}}$
\Statex
\Procedure{Train-GRPO}{$\theta_{\text{init}}, \mathcal{B}_{\text{src}}, T$}
    \State $\theta \gets \theta_{\text{init}}$
    \For{$t = 1, \ldots, T$}
        \State $\theta_{\text{old}} \gets \theta$
        \State Sample mini-batch $\mathcal{B} \subset \mathcal{B}_{\text{src}}$
        \For{each $x_i \in \mathcal{B}$}
            \State Draw $G$ candidates $\{o_i^{(g)}\}_{g=1}^{G} \sim \pi_{\theta_{\text{old}}}(\cdot \mid x_i)$, each with predicted label $\hat{y}_i^{(g)}$
            \State Compute reward $R_i^{(g)} = +1$ if $\hat{y}_i^{(g)} = y_i$, else $-1$
            \State Compute advantage $A_i^{(g)} = \big(R_i^{(g)} - \mathrm{mean}_{g'} R_i^{(g')}\big) / \mathrm{std}_{g'}\, R_i^{(g')}$
        \EndFor
        \State Update $\theta$ by minimizing $\mathcal{L}_{\text{GRPO}}(\theta; \mathcal{B})$ \Comment{Eq.~\eqref{eq:loss-grpo}}
    \EndFor
    \State \Return $\theta$
\EndProcedure
\end{algorithmic}
\end{algorithm}

\section{Experimental Setup}
\label{app:setup}

\subsection{Main Experimental Setup}
\label{app:main-setup}

\subsubsection{Models}
\label{app:models}
Our prompt-only \csa\ evaluation (Appendix~\ref{app:prompting-study}) spans the Qwen3~\citep{yang2025qwen3technicalreport}, Gemma3~\citep{gemmateam2025gemma3technicalreport}, Llama3~\citep{grattafiori2024llama3herdmodels}, OLMo2~\citep{olmo20242}, GPT-5~\citep{openai2025gpt5}, and Gemini 3.1~\citep{google2026gemini31} families, covering both open-weight and frontier closed-weight models.

For \csa\ through training experiments, we experiment with two model families. From the Qwen3 family~\citep{yang2025qwen3technicalreport}, we use four scales: 1.7B, 4B, 8B, and 14B parameters, to study how \csa\  varies with model capacity within a single family. We additionally include OLMo2~\citep{olmo20242} at 7B and 13B parameters to assess generalization across model families. For the SFT Teacher strategy, we use Qwen3-235B-A22B as the teacher model.

\subsubsection{Datasets}
\label{app:datasets}
We evaluate on two domains: \textbf{Math} and \textbf{Science}. The prompt-only study in Appendix~\ref{app:prompting-study} uses the Math test set. Our main results in Section~\ref{sec:experiments} cover both domains.

\paragraph{Math.} We aggregate problems from three sources spanning a wide difficulty spectrum: GSM8K~\citep{cobbe2021trainingverifierssolvemath} (grade-school arithmetic word problems), MATH500~\citep{lightman2023lets} (competition-style problems across algebra, geometry, number theory, etc.), and AIME~\citep{aime_1983_2024} (American Invitational Mathematics Examination problems from 1983 to 2024). All problems have open-ended numerical or symbolic answers. Table~\ref{tab:dataset_math} summarizes the size of each source and the train/test split used in our experiments.

\paragraph{Science.} We use multiple-choice questions from MMLU-Pro~\citep{wang2024mmluprorobustchallengingmultitask}, restricted to four science-relevant categories: \texttt{biology}, \texttt{chemistry}, \texttt{health}, and \texttt{physics}. Each question has about 10 answer options, and we feed the question together with all options as the query to the model. Table~\ref{tab:dataset_science} reports the per-category counts and the overall train/test split.

\begin{table}[ht]
\centering
\caption{Dataset composition.}
\label{tab:datasets}
\small
\begin{subtable}{0.48\textwidth}
\centering
\caption{Math}
\label{tab:dataset_math}
\begin{tabular}{lccc}
\toprule
Source & \#Train & \#Test & \#Total \\
\midrule
GSM8K   & 1000 & 300 & 1300 \\
MATH500 & 400  & 100 & 500  \\
AIME    & 860  & 90  & 950  \\
\midrule
\textbf{Total} & \textbf{2260} & \textbf{490} & \textbf{2750} \\
\bottomrule
\end{tabular}
\end{subtable}
\hfill
\begin{subtable}{0.48\textwidth}
\centering
\caption{Science}
\label{tab:dataset_science}
\begin{tabular}{lccc}
\toprule
Category  & \#Train & \#Test & \#Total \\
\midrule
Biology   & 590  & 127 & 717  \\
Chemistry & 932  & 200 & 1132 \\
Health    & 674  & 144 & 818  \\
Physics   & 1070 & 229 & 1299 \\
\midrule
\textbf{Total} & \textbf{3266} & \textbf{700} & \textbf{3966} \\
\bottomrule
\end{tabular}
\end{subtable}
\end{table}

\subsubsection{\csa\ Inference Setup}
\label{app:csa-inference_setup}

We use vLLM~\citep{kwon2023efficientmemorymanagementlarge} for inference across all evaluated models. Sampling parameters are kept consistent across models:

\begin{itemize}
    \item Temperature: 1.0
    \item Top-$p$: 1.0
    \item Top-$k$: -1
    \item Number of samples per query: 1
\end{itemize}

For closed-weight models (GPT-5, Gemini 3.1), we query the official APIs provided by \href{https://platform.openai.com/docs/api-reference}{OpenAI}
and \href{https://ai.google.dev/api}{Google}.

\subsubsection{Label Construction}
\label{app:label}
A single sampled answer can succeed or fail because of decoding variability. We therefore sample $K$ independent task solutions and aggregate their correctness into a binary capability label. Repeated attempts provide evidence about the model's ability under a specified sampling protocol, rather than treating one failed generation as proof that the query is unsolvable. We use $K=5$ to obtain this evidence at a moderate evaluation cost. Increasing $K$ also changes the meaning of an any-correct label, so $K$ is part of the target definition rather than only a variance-reduction parameter. Appendix~\ref{app:label-ablations} examines other values of $K$ and aggregation thresholds.

Let $c_i^{(k)}\in\{0,1\}$ indicate whether attempt $k$ answers query $i$ correctly. We assign \selfsolve\ when the relevant condition below holds, and \delegate\ otherwise.

\paragraph{Math: any-correct.} For open-ended Math, the condition is $\sum_{k=1}^{K}c_i^{(k)}\geq1$: at least one of the five attempts must be correct. Guessing an exact open-ended answer is generally less likely than guessing a multiple-choice option. A correct attempt therefore provides useful evidence that the model can solve the problem, even if its sampled solutions are not consistently successful. This label measures success within $K$ attempts, not reliable success on a single attempt. If the single-attempt success probability is $p$, independent sampling gives an any-correct probability of $1-(1-p)^K$.

\paragraph{Science: majority-correct.} For multiple-choice Science, we require $\sum_{k=1}^{K}c_i^{(k)}>K/2$, or at least three correct answers out of five. A single correct option can arise by guessing, so a majority provides stronger evidence of consistent success. We independently shuffle the option order for each attempt to reduce reliance on option position. Neither majority voting nor option shuffling eliminates all chance successes; the rule makes the target more stringent than any-correct for this answer format.

Training labels are constructed from the initial model and held fixed. For evaluation, we repeat the same procedure with each trained model to assess decisions against its current capability. The single-attempt routing objective is discussed separately in Appendix~\ref{sec:ca-routing}.

\subsubsection{Evaluation Metrics}
\label{app:metrics}

\paragraph{\MFfull\ (\MF).} Our primary assessment metric, \MF, is the unweighted mean of the per-class F1 scores for the \selfsolve\ and \delegate\ classes,
\begin{equation}
\MF = \tfrac{1}{2}\left(\mathrm{F1}_{\selfsolve} + \mathrm{F1}_{\delegate}\right),
\end{equation}
where each per-class F1 is the harmonic mean of precision and recall on that class:
\begin{equation}
\mathrm{F1}_{c} = \frac{2\,\mathrm{Prec}_{c} \cdot \mathrm{Rec}_{c}}{\mathrm{Prec}_{c} + \mathrm{Rec}_{c}}, \qquad c \in \{\selfsolve, \delegate\}.
\end{equation}
The two classes receive equal weight, so performance on both contributes to the score. Class prevalence and errors on a rare class can still affect \MF.

\emph{Example:} on a test set of 100 queries with 60 true \selfsolve\ and 40 true \delegate, suppose the model predicts \selfsolve\ on 70 queries (50 correct, 20 false) and \delegate\ on 30 queries (20 correct, 10 false). Then $\mathrm{Prec}_{\selfsolve} = 50/70 \approx 0.71$, $\mathrm{Rec}_{\selfsolve} = 50/60 \approx 0.83$, giving $\mathrm{F1}_{\selfsolve} \approx 0.77$. Similarly, $\mathrm{Prec}_{\delegate} = 20/30 \approx 0.67$, $\mathrm{Rec}_{\delegate} = 20/40 = 0.50$, giving $\mathrm{F1}_{\delegate} \approx 0.57$. Thus $\MF = \tfrac{1}{2}(0.77 + 0.57) \approx 0.67$.

\paragraph{\CDSfull\ (\CDS).} We report \CDS\ as an auxiliary measure of the separation in capability-label proportions between queries assigned \selfsolve\ and those assigned \delegate. Concretely, we partition the test set by the model's predicted label and let $p_S$ and $p_D$ denote the fractions of queries labeled solvable within the \selfsolve\ and \delegate\ groups, respectively, with group sizes $n_S$ and $n_D$. These fractions use the same $K=5$ capability labels as \MF. \CDS\ is the (unpooled) two-proportion $z$-statistic
\begin{equation}
\label{eq:CDS}
\CDS = \frac{p_S - p_D}{\sqrt{\dfrac{p_S(1-p_S)}{n_S} + \dfrac{p_D(1-p_D)}{n_D}}},
\end{equation}
which normalizes the difference in solvable-label proportions by its estimated standard error. Its magnitude depends on sample size and group balance, and it can be unstable when a group is small or its accuracy is near zero or one. We therefore use \MF\ for our main assessment comparisons and do not interpret \CDS\ as a comparable effect size across datasets.

\emph{Example:} suppose the policy assigns $n_S = 200$ queries to \selfsolve\ and $n_D = 100$ to \delegate. Under the $K=5$ capability-label protocol, 170 queries in the first group and 20 in the second are labeled solvable. Thus $p_S = 170/200 = 0.85$ and $p_D = 20/100 = 0.20$, giving
\begin{equation*}
\CDS = \frac{0.85 - 0.20}{\sqrt{\dfrac{0.85 \cdot 0.15}{200} + \dfrac{0.20 \cdot 0.80}{100}}} \approx 13.7.
\end{equation*}

\paragraph{\CRfull\ (\CR).} To verify that \csa\  training preserves the model's original capabilities, we report \CR, defined as the percentage of the original model's problem-solving accuracy retained after \csa\  training:
\begin{equation}
    \CR = \frac{\text{Acc}_{\text{post}}}{\text{Acc}_{\text{pre}}} \times 100\%,
\end{equation}
where $\text{Acc}_{\text{pre}}$ and $\text{Acc}_{\text{post}}$ denote the model's problem-solving accuracy before and after \csa\  training, respectively. To reduce variance from stochastic decoding, both $\text{Acc}_{\text{pre}}$ and $\text{Acc}_{\text{post}}$ are averaged over 5 independent runs. We compute ratios and aggregate statistics from unrounded values and round only for presentation.

\emph{Example:} if the model solves 60\% of the test queries before \csa\  training ($\text{Acc}_{\text{pre}} = 0.60$) and 57\% after ($\text{Acc}_{\text{post}} = 0.57$), then $\CR = 0.57 / 0.60 \times 100\% = 95\%$.

\subsection{Training Settings}
\label{app:training}

Table~\ref{tab:hparams} reports the hyperparameters for both SFT and GRPO. SFT and GRPO both use full-parameter updates, implemented with the TRL library\footnote{\url{https://github.com/huggingface/trl}} on top of DeepSpeed ZeRO-3 \citep{rasley2020deepspeed}.

\begin{table}[ht]
\centering
\caption{Training hyperparameters for SFT and GRPO. Both use full-parameter updates.}
\small
\begin{tabular}{lcc}
\toprule
\textbf{Hyperparameter} & \textbf{SFT} & \textbf{GRPO} \\
\midrule
\multicolumn{3}{l}{\textbf{\emph{Optimization}}} \\
\quad Optimizer            & AdamW & AdamW \\
\quad Learning rate        & $1\times 10^{-5}$ & $1\times 10^{-6}$ \\
\quad LR schedule          & cosine & cosine \\
\quad Warmup ratio         & 0.1 & 0.1 \\
\quad Weight decay         & 0.01 & 0.01 \\
\quad Epochs               & 1 / 3\footnotemark[2] & 1 / 3\footnotemark[2] \\
\quad Per-device batch size & 8 & 16 \\
\quad Gradient accumulation & 1 & 1 \\
\quad Precision            & bf16 & bf16 \\
\multicolumn{3}{l}{\textbf{\emph{GRPO-specific}}} \\
\quad Rollouts per prompt ($G$) & --- & 16 \\
\quad KL coefficient ($\beta$)  & --- & 0.01 \\
\quad Clipping threshold ($\epsilon$) & --- & 0.2 \\
\quad Sampling temperature      & --- & 1.0 \\
\quad Top-$p$                   & --- & 1.0 \\
\quad Rollout backend           & --- & vLLM \\
\multicolumn{3}{l}{\textbf{\emph{Training framework}}} \\
\quad Framework             & TRL & TRL \\
\quad Distributed strategy  & DeepSpeed ZeRO-3 & DeepSpeed ZeRO-3 \\
\quad CPU offload           & False & False \\
\bottomrule
\end{tabular}
\label{tab:hparams}
\end{table}

\footnotetext[2]{We train for 1 epoch on Math and 3 epochs on Science.}

\paragraph{DFW Hyperparameters.} For DFW training, we use the same hyperparameter setup as in Table~\ref{tab:hparams}, except for the number of training steps, which we report in Table~\ref{tab:dfw-steps} for each model. Because the size of the diversified subset varies across models, the training schedule differs accordingly: smaller models yield larger subsets and are typically early-stopped before a full epoch, while larger models, with smaller subsets, are trained for 1--2 epochs. This keeps the total number of training steps roughly comparable across model sizes.

\paragraph{Training Pipeline across Model Families.} Another point worth noting is the difference in training pipelines across model families. For Qwen3 models, we follow the full two-stage RLVR pipeline described in Section~\ref{sec:training}. For OLMo2 models, we omit the DFW stage, since they already produce sufficiently diverse responses during rollout as shown in Table \ref{tab:dfw-steps}. This is consistent with Figure~\ref{fig:overconfidence} in Appendix~\ref{app:prompting-study}: while Qwen3 models exhibit \selfsolve\ rates close to 100\%, OLMo2 models show noticeably lower \selfsolve\ rates, indicating greater rollout diversity without explicit intervention.

\begin{table}[t]
\centering
\caption{Subset sizes and training steps for DFW across different models on Math and Science, where \emph{Subset Size} denotes the number of samples selected by DFW from the full training set.}
\small
\begin{tabular}{lcccc}
\toprule
& \multicolumn{2}{c}{\textbf{Math}} & \multicolumn{2}{c}{\textbf{Science}} \\
\cmidrule(lr){2-3} \cmidrule(lr){4-5}
\textbf{Model} & Subset Size & Training Steps & Subset Size & Training Steps \\
\midrule
Qwen3-1.7B & 544 & 150 & 1781 & 100 \\
Qwen3-4B   & 183 & 46  & 111 & 150 \\
Qwen3-8B & 165 & 42  & 302 & 152  \\
Qwen3-14B & 148 & 57 & 171 & 129 \\
OLMo2-7B   & 1832 & --- & 3131 & --- \\
OLMo2-13B   & 2035 & --- & 2977 & --- \\
\bottomrule
\end{tabular}
\label{tab:dfw-steps}
\end{table}

\paragraph{Hardware.} All main experiments are run on NVIDIA B200 GPUs. SFT runs use a single B200, except for Qwen3-14B and OLMo2-13B, which use 2 $\times$ B200s. All GRPO runs use 4 $\times$ B200s.

\subsection{Capability-Preserving Training Baseline Setup}
\label{app:training-baseline-setup}
This experiment extends the main training comparison in Table~\ref{tab:main} by testing whether RLVR's advantage persists when supervised training better preserves capability or uses the same example-selection curriculum. Table~\ref{tab:training-alignment} summarizes the controls; results appear in Table~\ref{tab:rvs} and Appendix~\ref{app:training-baseline-results}. This is a separate controlled comparison on two model sizes, not a rerun of every cell in Table~\ref{tab:main}.

\begin{table}[t]
\centering\small
\caption{Training changes relative to the original \sftteacher\ setup. Supervised variants retain teacher-rationale supervision and the same capability labels unless stated otherwise.}
\label{tab:training-alignment}
\renewcommand{\arraystretch}{1.18}
\begin{tabular}{@{}p{0.22\linewidth}p{\dimexpr0.78\linewidth-2\tabcolsep\relax}@{}}
\toprule
\textbf{Method} & \textbf{Changes relative to \sftteacher} \\
\midrule
SFT with smaller lr & LR: $10^{-5}$ $\rightarrow$ best of $\{10^{-5},10^{-6}\}$, including RLVR's LR. \\
LoRA SFT & Full-parameter updates $\rightarrow$ rank-16 LoRA; LR: best of $\{10^{-4},10^{-6}\}$. \\
KL SFT & SFT loss $\rightarrow$ SFT loss + forward-KL penalty to the initial model (weight $0.05$). \\
Task-mixed SFT & CSA-only supervision $\rightarrow$ CSA + task-solving rehearsal, mixed 1:1. \\
DFW-matched SFT & Full-data training $\rightarrow$ RLVR's diversified-subset curriculum followed by full-data training. \\
Query Probe & Generative SFT $\rightarrow$ a classifier on frozen query representations; no solver updates. \\
\bottomrule
\end{tabular}
\end{table}

\subsubsection{Models and data}
We compare capability self-assessment methods using Qwen3-1.7B and Qwen3-4B, training separate models for math and science. The math dataset contains 2,260 training and 490 test questions from GSM8K, MATH500, and AIME. The science dataset contains 3,266 training and 700 test questions from MMLU-Pro biology, chemistry, health, and physics. For each model and domain, we reserve 300 training questions for validation using a split stratified by the initial model's correctness label.

We obtain supervision by sampling five task answers per question from the initial model. A math question is labeled solvable if any sample is correct, while a science question is labeled solvable if a majority of five option-shuffled attempts are correct. Solvable and unsolvable questions receive \texttt{SELF\_SOLVE} and \texttt{DELEGATE} labels, respectively. For SFT, the teacher model generates one rationale conditioned on each label. All generative methods use an analysis-first response format consisting of an \texttt{<analysis>} block followed by a \texttt{<decision>} block.

\subsubsection{Baselines}
We include the prompted, untrained model; full-parameter SFT with smaller lr; LoRA~\citep{hu2021lora} SFT with rank 16, scaling factor 32, and dropout 0.05 on attention and MLP projections; KL-regularized full SFT with a forward $D_{\mathrm{KL}}(\pi_{\mathrm{initial}}\Vert\pi)$ penalty of weight 0.05. We also train an additional SFT baseline with a 1:1 mixture of capability self-assessment examples and task-solving rehearsal examples. Each rehearsal target is the initial model's shortest correct generation for a training question, paired with the exact prompt that produced it. For science, this preserves the sampled option ordering. The two example types receive equal loss weight. We also compare with SFT with the same example-selection curriculum as DFW RLVR, and a supervised classifier on frozen, question-only model representations.

\subsubsection{Optimization and selection}
For the generative methods, we use an effective batch size of 64 questions, yielding 31 updates for math and 47 for science. SFT uses completion-only loss and a 4,096-token sequence limit. Training uses four GPUs, bfloat16, AdamW~\citep{loshchilov2017adamw}, weight decay 0.01, a cosine learning-rate schedule, and 10\% warm-up. We select the best config between learning rates of $10^{-6}$ and $10^{-5}$ for full-parameter SFT, $10^{-6}$ and $10^{-4}$ for LoRA SFT, and $10^{-6}$ for RLVR.

\section{Additional Results}
\label{app:more-results}

\subsection{Prompt-Only \csa\ Study}
\label{app:prompting-study}

We examine whether prompting alone elicits decisions that reflect a model's measured ability. Prior work reports overconfidence in verbalized confidence~\citep{xiong2023can-llm-express-uncertainty} and predictions of task success~\citep{barkan2026do}. Here we examine this issue through the \selfsolve\ / \delegate\ decision used in our \csa\ formulation.

\paragraph{Experimental setup.}
We evaluate the Qwen3, Gemma3, Llama3, OLMo2, GPT-5, and Gemini 3.1 families, covering open-weight and closed-weight models at multiple scales (Appendix~\ref{app:models}). All models use the same combined Math test set of 490 queries: 300 from GSM8K, 100 from MATH500, and 90 from AIME. Dataset details appear in Appendix~\ref{app:datasets}. No \csa\ training is applied in this study.

For each query, we prompt the model to assess its own capability and choose \selfsolve\ or \delegate\ before generating a task solution. The prompt requests an analysis followed by a structured decision (Appendix~\ref{app:csa_inference_template}). We obtain one assessment per query. Separately, we measure task accuracy using one solution attempt per query and compare the answer with ground truth. Task accuracy is measured on the full test set, including queries the model chooses to delegate. Open-weight inference uses vLLM, while closed-weight models are accessed through their official APIs. The assessment sampling settings are reported in Appendix~\ref{app:csa-inference_setup}.

\paragraph{Measurements.}
The \emph{self-solve rate} is the fraction of queries assigned \selfsolve. The \emph{task accuracy} is the fraction answered correctly in the separate solving evaluation. We compare these two rates for each model. This preliminary study uses one task attempt per query. The main training experiments instead construct capability labels from five attempts, as described in Appendix~\ref{app:label}.

\begin{figure}[t]
\centering
\includegraphics[width=1.0\textwidth]{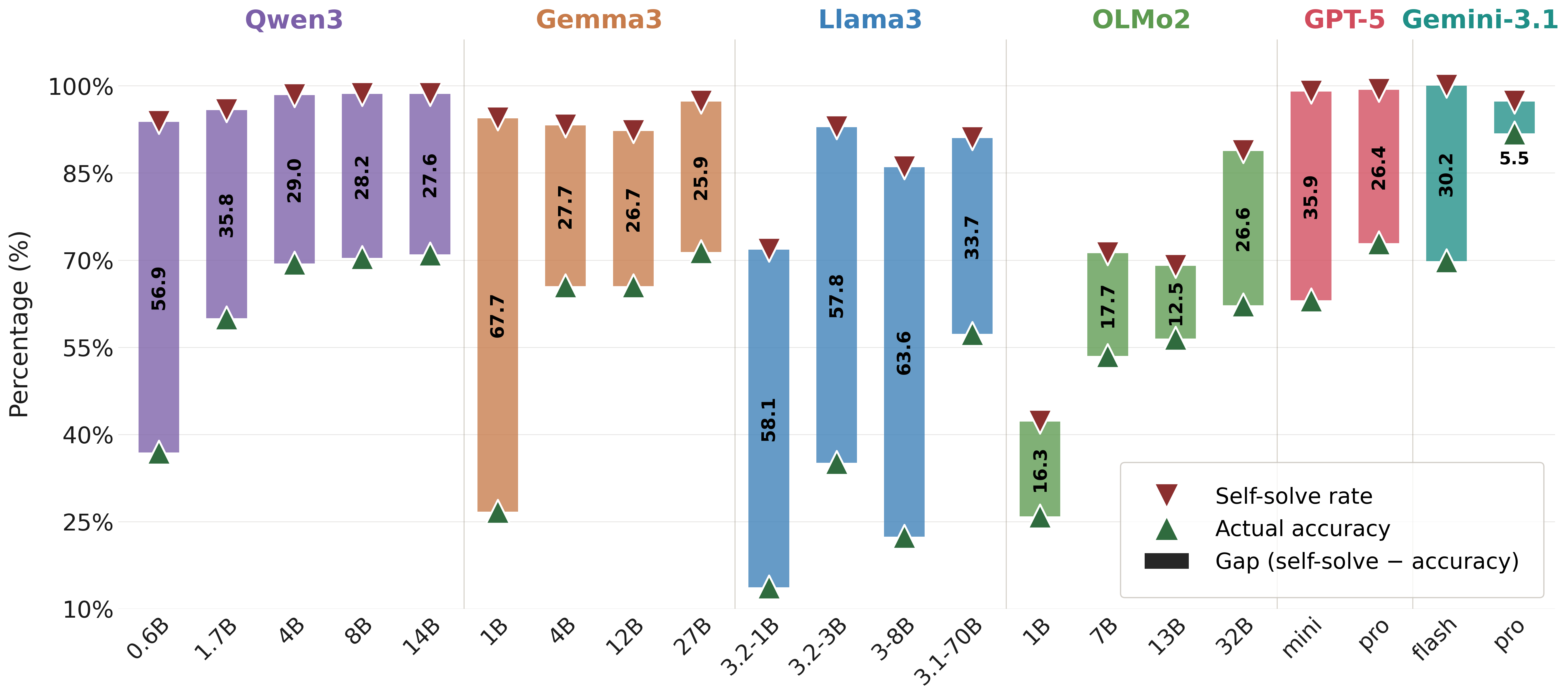}
\caption{Prompted self-solve rates and measured task accuracy across model families and sizes on a combined math test set. The
    {\color{selfsolvecolor}$\blacktriangledown$} marker denotes the \emph{self-solve rate}, and the
    {\color{accuracycolor}$\blacktriangle$} marker denotes task accuracy under the solving protocol. The shaded gap captures the model's aggregate tendency to attempt more queries than it solves successfully.}
\label{fig:overconfidence}
\end{figure}

\paragraph{Results and interpretation.}
Figure~\ref{fig:overconfidence} shows that the self-solve rate exceeds task accuracy for every evaluated model. Under this protocol, models choose to answer more queries than they successfully solve. This pattern is consistent with capability overestimation and extends the prior observations to our explicit decision setting. Some models choose \selfsolve\ on nearly every query, leaving little scope for selective help-seeking.

The gap measures aggregate mismatch, rather than whether the model identifies the right queries to solve or delegate. Matching the two rates alone would not establish accurate self-assessment: a model could select the correct number of queries but select the wrong ones. A single solution attempt also provides a noisy estimate of query-level capability. We therefore use repeated attempts and query-level assessment metrics in the main experiments, alongside capability retention. This study motivates learning \csa\ beyond the behavior elicited by the evaluated prompt. It does not establish that all prompting strategies fail.

\FloatBarrier

\subsection{Ablation Studies}
\label{app:ablation-studies}

\subsubsection{Capability-Label Construction}
\label{app:label-ablations}

We examine how the number of samples $K$ and the aggregation threshold affect Math \csa\ performance. Results are broadly similar across the tested settings.

\begin{figure}[t]
  \centering
  \begin{subfigure}[t]{0.33\textwidth}
    \centering
    \captionsetup{width=\linewidth,justification=centering,singlelinecheck=false}
    \includegraphics[width=\linewidth]{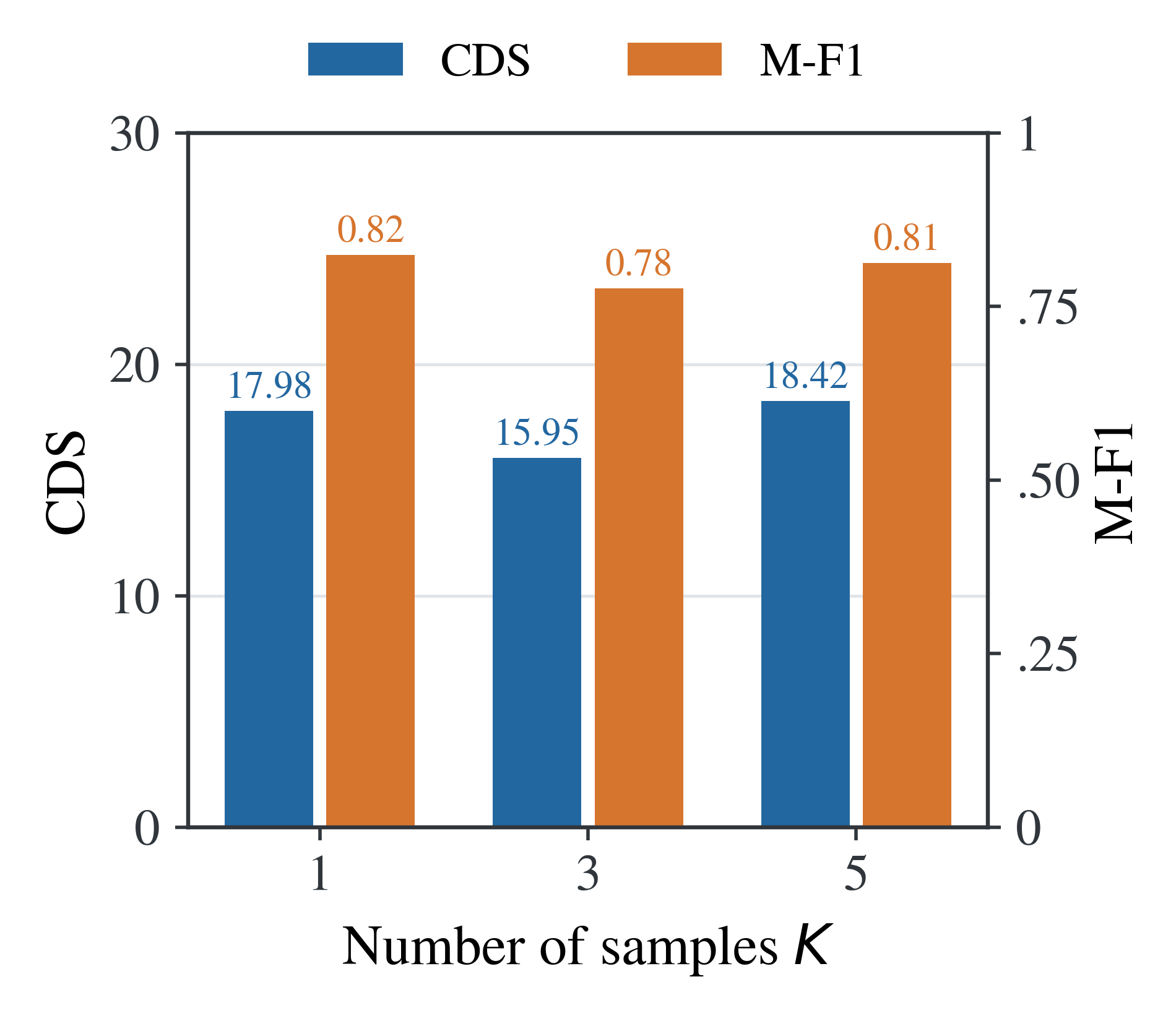}
    \caption{Number of samples $K$\\under any-correct aggregation.}
    \label{fig:math-sample-count}
  \end{subfigure}\hfill
  \begin{subfigure}[t]{0.33\textwidth}
    \centering
    \captionsetup{width=\linewidth,justification=centering,singlelinecheck=false}
    \includegraphics[width=\linewidth]{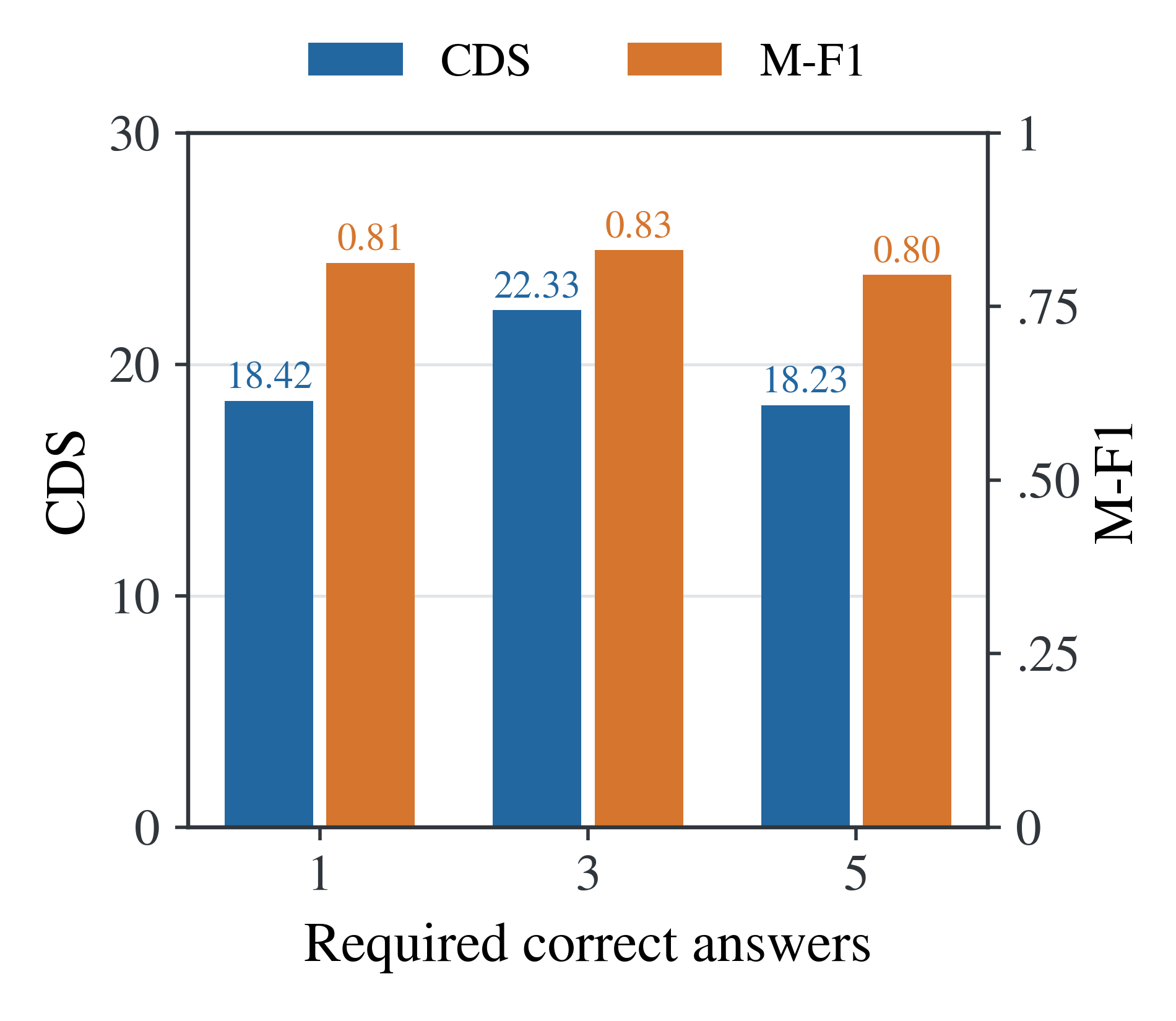}
    \caption{Required correct answers\\with $K=5$.}
    \label{fig:math-correct-threshold}
  \end{subfigure}\hfill
  \begin{subfigure}[t]{0.33\textwidth}
    \centering
    \captionsetup{width=\linewidth,justification=centering,singlelinecheck=false}
    \includegraphics[width=\linewidth]{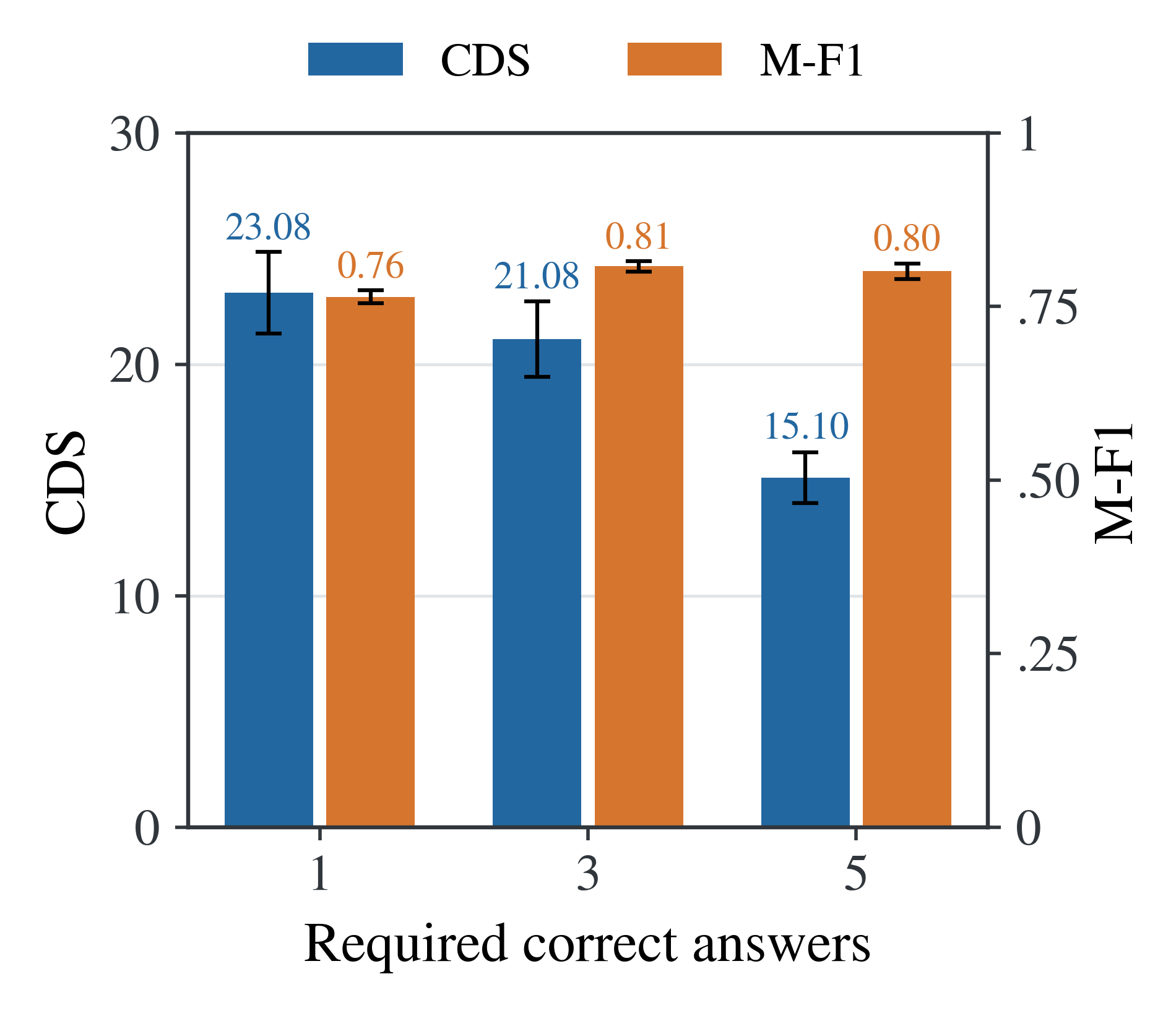}
    \caption{Threshold comparison under\\common single-draw evaluation.}
    \label{fig:math-threshold-single-draw}
  \end{subfigure}
  \caption{Sensitivity of Math RLVR to capability-label construction. Panels (a) and (b) use original-model capability labels. Panel (c) evaluates routers trained with $K=5$ and one-, three-, or five-correct thresholds under a common single-draw capability-label rule. Bars show means across five trials and error bars show sample standard deviations. Blue bars report \CDS{} and orange bars report \MF.}
  \label{fig:math-label-ablations}
\end{figure}

Figure~\ref{fig:math-sample-count} varies $K$ under any-correct aggregation, with \MF{} ranging from 0.78 to 0.82. Figure~\ref{fig:math-correct-threshold} fixes $K=5$ and varies the required number of correct answers. All tested thresholds remain effective, with the three-correct rule reaching \MF{} of 0.83. Overall, performance does not depend narrowly on a particular sample count or aggregation threshold.

Because these thresholds define different capability targets, we also compare the policies under a common single-attempt evaluation protocol. Each query receives a capability label based on one task answer from the corresponding trained model, and results are averaged over five trials. Figure~\ref{fig:math-threshold-single-draw} shows that the policies remain effective under this protocol. Overall, performance is broadly similar across the tested thresholds, supporting robustness to the aggregation threshold.

\subsubsection{Diversity-Filtered Warm-up}
\label{app:rollout-warmup}

Table~\ref{tab:ablation-warmup} confirms the importance of DFW: without it, the policy collapses to almost always predicting \selfsolve\ (441/490 at 4B and 484/490 at 8B), and both \MF\ and \CDS\ drop substantially. The reward standard deviation also falls sharply, showing that the rollout reward is too uniform to drive meaningful policy updates without DFW.

\begin{table}[t]
  \centering
  \caption{Ablation of DFW on Math. $n_S/n_D$: \selfsolve/\delegate\ prediction counts.}
  \label{tab:ablation-warmup}
  \small
  \setlength{\tabcolsep}{7pt}
  \renewcommand{\arraystretch}{0.94}
  \begin{tabular}{lcccc}
    \toprule
    & \multicolumn{2}{c}{\textbf{Qwen3-4B}} & \multicolumn{2}{c}{\textbf{Qwen3-8B}} \\
    \cmidrule(lr){2-3} \cmidrule(lr){4-5}
    \textbf{Metric} & w/o & w/ & w/o & w/ \\
    \midrule
    \MF           & 0.70  & \textbf{0.80} & 0.49  & \textbf{0.77} \\

    \CDS          & 12.10  & \textbf{17.17}  & 4.08   & \textbf{14.13}  \\
    n\textsubscript{S} / n\textsubscript{D} & 441/49 & \textbf{417/73} & 484/6 & \textbf{426/64} \\
    Reward std    & 0.133  & \textbf{0.214} & 0.106  & \textbf{0.232} \\
    \bottomrule
  \end{tabular}
\end{table}

Beyond this ablation, we find that stronger models yield smaller diversified subsets and that rollout diversity emerges primarily on queries the model should delegate. We examine both patterns next.

\paragraph{Two findings on the diversified subset.}
\label{app:diversified}

\begin{figure}[ht]
  \centering
  \begin{minipage}[t]{0.48\textwidth}
    \centering
    \includegraphics[width=\linewidth]{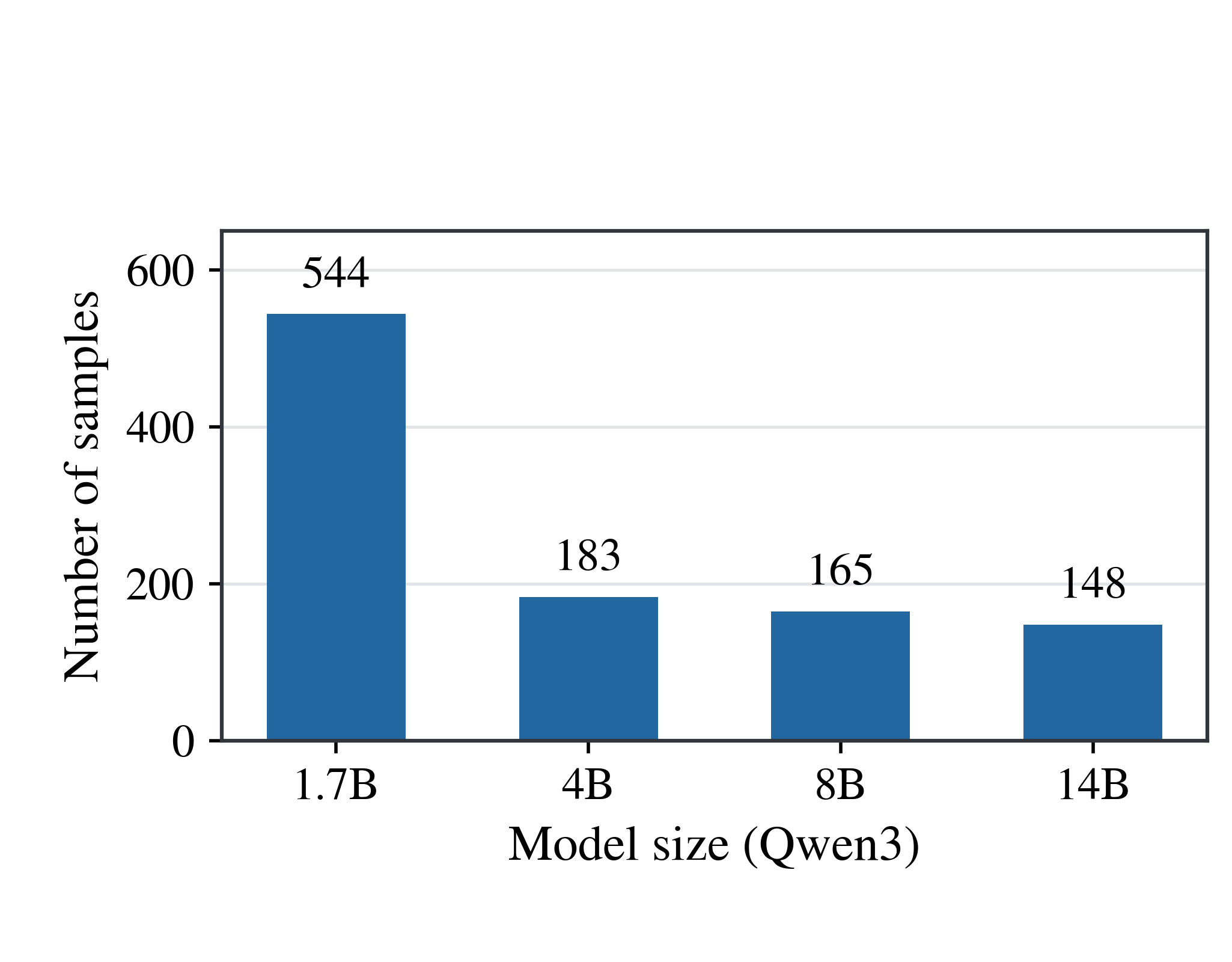}
    \caption{Number of samples per model size on Math whose 16 rollouts during DFW yield both \selfsolve\ and \delegate\ responses.}
    \label{fig:diversified-counts}
  \end{minipage}%
  \hfill
  \begin{minipage}[t]{0.48\textwidth}
    \centering
    \includegraphics[width=\linewidth]{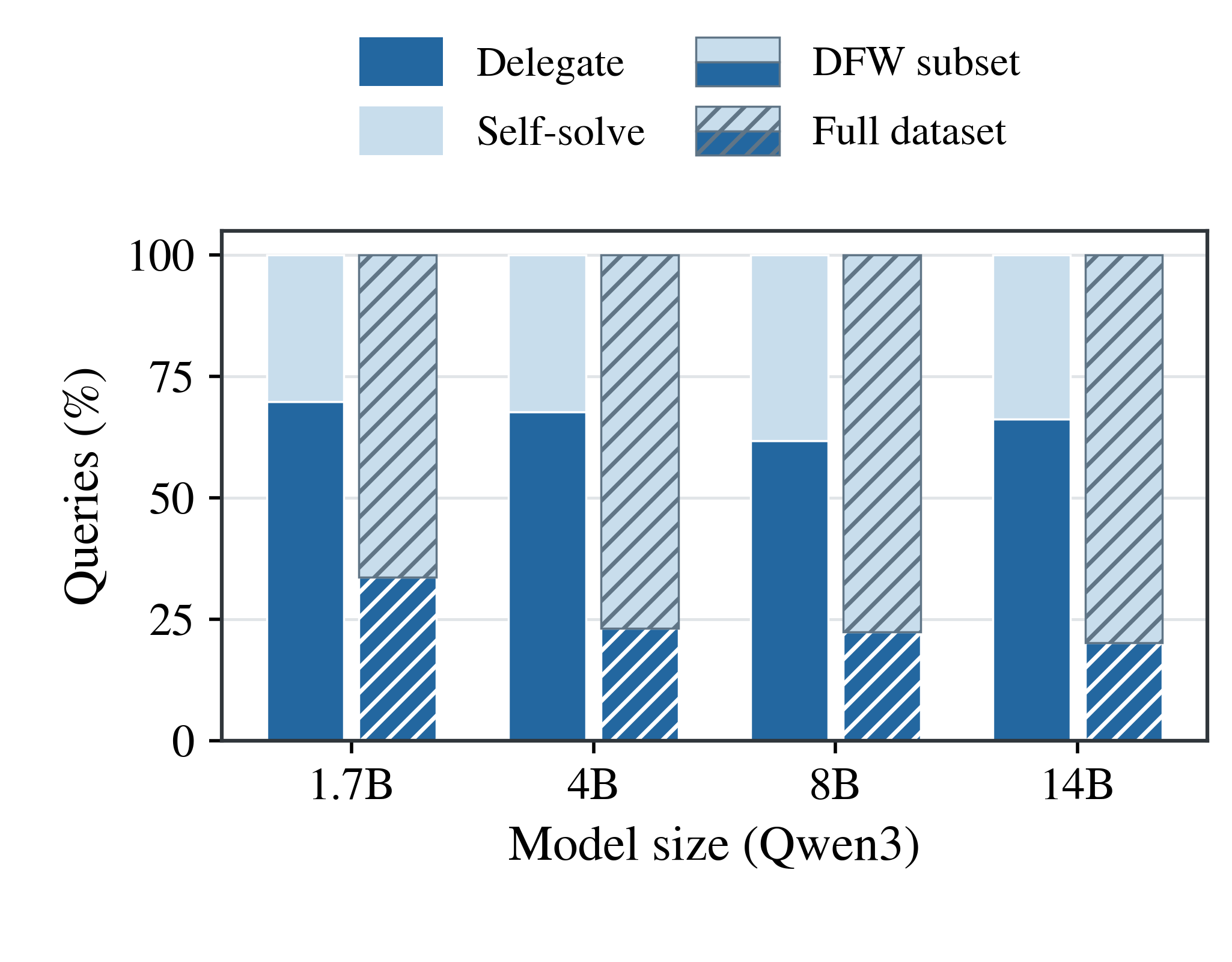}
    \caption{Ground-truth label distribution (\selfsolve\ vs.\ \delegate) in the diversified subset during DFW (solid) compared to the full training set (hatched).}
    \label{fig:diversified-breakdown}
  \end{minipage}
  \vspace{-10pt}
\end{figure}

\textbf{Finding 1: Stronger models yield smaller diversified subsets.} Figure~\ref{fig:diversified-counts} shows the number of Math queries that pass the diversification filter as a function of model size. The subset shrinks sharply with scale: the 1.7B policy contributes roughly 550 diversified samples, and the 4B/8B/14B policies all fall below 200. This is consistent with the diversification criterion acting as an \emph{uncertainty filter}: as the policy becomes stronger, its 16 rollouts on a given query are more likely to agree (all \textsc{Self-Solve} or all \textsc{Delegate}), and fewer queries survive the ``both labels present'' condition. Larger models are simply more internally consistent about what they can and cannot do, and the per-step cost of DFW decreases naturally with scale.

\textbf{Finding 2: Rollout diversity emerges primarily on queries the model should delegate.} Figure~\ref{fig:diversified-breakdown} contrasts the label distribution of the diversified subset (solid) against the full training set (hatched). The comparison against the full set is what makes this finding meaningful: since the diversified subset is by construction a subset of the full training data, if the model had no preference about \emph{which} queries trigger rollout disagreement, the \textsc{Self-Solve} / \textsc{Delegate} proportions in the two distributions should roughly match. Instead, we observe a clear and systematic shift---across all four model sizes, the diversified subset is skewed toward \textsc{Delegate} (62\%--70\%), compared to a 20\%--34\% \textsc{Delegate} rate in the full data. This shift indicates that rollout diversity does not arise uniformly across the data, but emerges preferentially on queries that lie \emph{beyond} the model's reliable solving capability. On clearly-within- and clearly-beyond-capability queries the model's 16 rollouts tend to agree. The boundary cases---where the model \emph{should} delegate but is occasionally tempted to attempt---are exactly what surface in the diversified subset, and exactly where DFW's training signal is most needed.

\subsubsection{\csa\ Output Order}\label{app:csa-output-order-ablation}

Because the models are autoregressive, the position of the \texttt{<decision>} tag determines how soon a routing decision is available. Under analysis-first the decision is emitted only after the rationale, a median of $81$--$384$ tokens into the response depending on model and task. Under decision-first, inference can stop once the decision field is decoded, so routing requires only the prompt and decision tokens. Table~\ref{tab:csa-output-order} tests this change on three settings: decision-first CDS differs from analysis-first by 0.05, 1.41, and 0.05, respectively, while M-F1 is lower by 0.3--4.7 percentage points, computed before rounding the displayed means. These checkpoints were trained with analysis preceding the decision, so decision-first prompting also changes their expected output format. Across three random seeds, the standard deviation never exceeds 0.8 percentage points in M-F1 or accuracy and $0.51$ in CDS.

\begin{table}[t]
\centering
\small
\caption{\csa\ output-order ablation. We evaluate analysis-first \csa\ checkpoints
prompted in both output orders on the held-out test split. Values are mean
$\pm$ standard deviation over three random seeds.}
\label{tab:csa-output-order}
\begin{tabular}{llcc}
\toprule
Model / Task & Output order & M-F1 & CDS \\
\midrule
\multirow{2}{*}{Qwen3-1.7B / Math} & Analysis-first & $0.80 \pm 0.00$ & $16.13 \pm 0.22$ \\
 & Decision-first & $0.80 \pm 0.01$ & $16.08 \pm 0.20$ \\
\cmidrule(lr){1-4}
\multirow{2}{*}{Qwen3-1.7B / Science} & Analysis-first & $0.67 \pm 0.01$ & $10.72 \pm 0.51$ \\
 & Decision-first & $0.62 \pm 0.00$ & $9.31 \pm 0.36$ \\
\cmidrule(lr){1-4}
\multirow{2}{*}{Qwen3-4B / Science} & Analysis-first & $0.65 \pm 0.01$ & $8.30 \pm 0.50$ \\
 & Decision-first & $0.61 \pm 0.01$ & $8.25 \pm 0.09$ \\
\bottomrule
\end{tabular}
\end{table}

\subsection{Training Dynamics of SFT and RLVR}
\label{app:training-dynamics}

\begin{figure}[t]
  \centering
  \includegraphics[width=\textwidth]{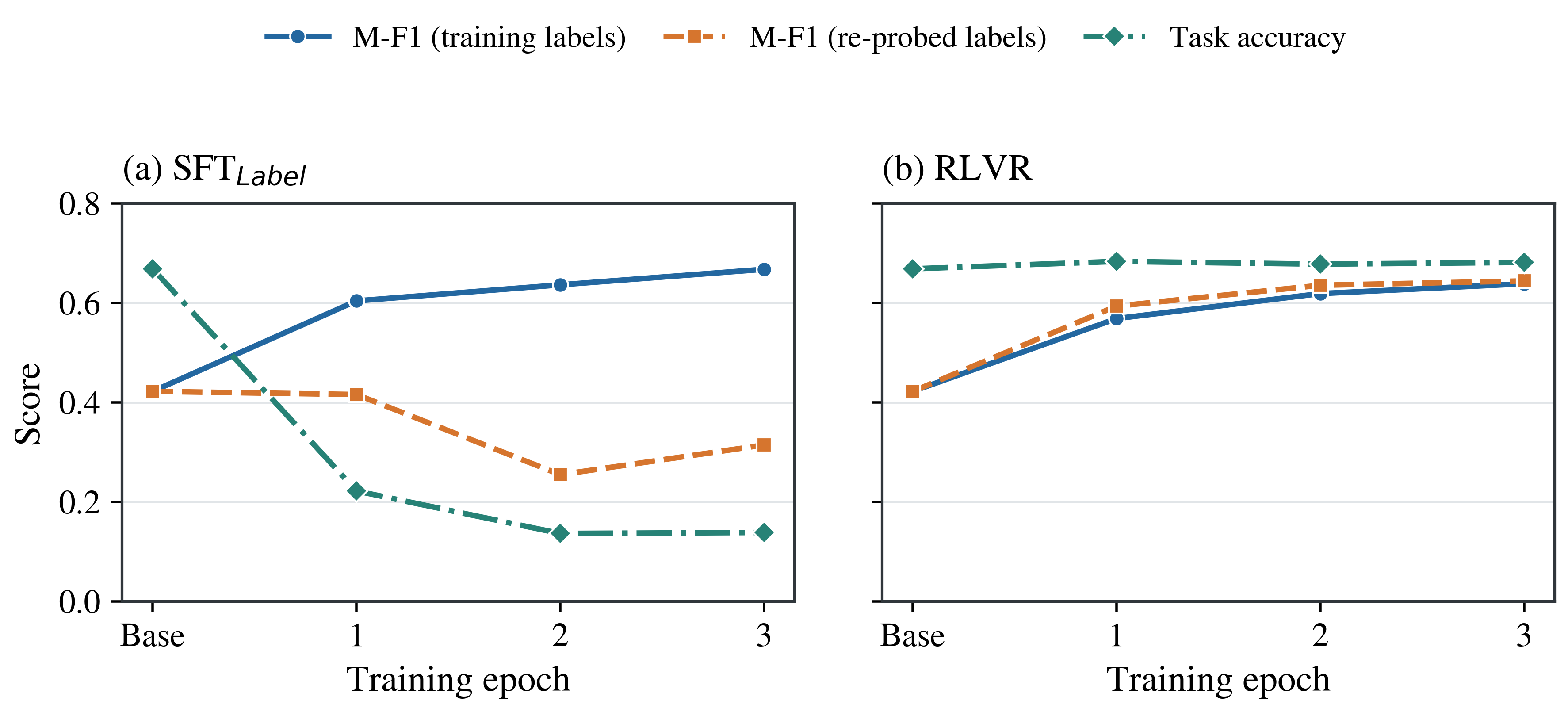}
  \caption{Training dynamics of Qwen3-4B on Science. \MF\ is evaluated against the original and freshly re-probed labels. Accuracy measures original task performance. \textbf{(a)} \sftlabel. \textbf{(b)} RLVR.}
  \label{fig:sft_vs_rlvr}
\end{figure}

As shown in Figure~\ref{fig:sft_vs_rlvr}, we track Qwen3-4B over three training epochs on Science, and the two methods exhibit sharply different trends. For \sftlabel, \MF\ against the original labels keeps rising while \MF\ against the re-probed labels drops, and problem-solving accuracy collapses sharply after the first epoch. Thus, the model learns a fixed delegation rule frozen at the base-model snapshot while its underlying capability changes. For RLVR, the two \MF\ curves rise together and problem-solving accuracy stays at the base level, indicating that self-assessment remains aligned with the model's current capability. This contrast supports the assumption in Section~\ref{sec:csa-formulation}: acquiring \csa\ should not come at the cost of the underlying problem-solving ability.

\subsection{Capability-Preserving Training Baselines}
\label{app:training-baseline-results}
Experimental settings are given in Appendix~\ref{app:training-baseline-setup}.
Table~\ref{tab:rvs-full} provides the full mean $\pm$ standard deviation results for the supervised baselines summarized in Table~\ref{tab:rvs}. The results separate two effects that are confounded in standard SFT: learning the CSA decision and preserving the solver whose capability is being assessed. These results are averaged across three runs; Table~\ref{tab:main} reports the main training comparison.

SFT with smaller lr improves Macro-F1 over the Base model, but can substantially alter the underlying solver. This effect is most pronounced for Qwen3-1.7B, where \CR\ falls to $72.4$ on Math and $84.5$ on Science. Qwen3-4B is more robust on Math ($97.3$ \CR), but still loses substantial capability on Science ($88.6$). Thus, improved CSA accuracy under SFT with smaller lr does not necessarily correspond to faithful self-assessment of the original capability boundary.

LoRA SFT largely removes this failure mode: \CR\ remains approximately $99$--$100\%$ in all four settings. However, preserving capability alone is not sufficient to obtain consistently strong CSA. LoRA reaches a relatively high \MF\ of $0.73$ for Qwen3-1.7B on Math, but drops to $0.51$ on Science. For Qwen3-4B, it obtains $0.62$ and $0.54$ on Math and Science, respectively. The relatively large variation for some LoRA settings, such as $0.73 \pm 0.07$ on Qwen3-1.7B Math, also indicates greater run-to-run sensitivity than \rlvr\ in these experiments.

KL-regularized SFT provides an intermediate trade-off. Relative to SFT with smaller lr, it generally improves capability retention---for example, from $72.4$ to $90.6$ \CR\ on Qwen3-1.7B Math and from $84.5$ to $92.0$ on Science---but its CSA performance remains below \rlvr. Task-mixed SFT similarly reduces capability degradation in several settings and performs strongly on Qwen3-4B Math, reaching $0.75$ \MF\ with $100.2$ \CR. However, its behavior is not uniform across domains: on Qwen3-1.7B Science, capability falls to $82.0 \pm 3.0$ despite mixing task-solving supervision into training.

In comparison, \rlvr\ achieves the highest \MF\ in every model--domain setting: $0.84$ and $0.66$ for Qwen3-1.7B on Math and Science, and $0.85$ and $0.67$ for Qwen3-4B. At the same time, its \CR\ remains close to or above the Base model in every case ($100.3$--$102.8\%$). The standard deviations are also small for \MF\ ($0.01$--$0.02$ across the four settings). Taken together, these results indicate that \rlvr's advantage over supervised training is not merely due to the severe capability loss of naive SFT: the same pattern persists against LoRA, KL regularization, and task-mixed training, which explicitly or implicitly constrain changes to the base solver.  RLVR preserves solving ability and beats DFW-matched SFT, suggesting that neither retention nor DFW’s training curriculum alone explains its gains. See \citet{chu2025sft} and \citet{shenfeld2026rls} for related discussions.

\begin{table}[t]
\centering
\small
\caption{
Full comparison of capability self-assessment training methods on Math and Science for Qwen3-1.7B and Qwen3-4B. We report Macro-F1 (\MF) for self-assessment quality and Capability Ratio (\CR) for preservation of the model's original task capability. Higher is better for both metrics. Values are mean $\pm$ standard deviation across three runs. Base denotes the untrained model.
}
\label{tab:rvs-full}
\setlength{\tabcolsep}{4pt}
\renewcommand{\arraystretch}{1.08}
\begin{tabular*}{\textwidth}{@{\extracolsep{\fill}}lcccc@{}}
\toprule
& \multicolumn{2}{c}{\textbf{Qwen3-1.7B}} & \multicolumn{2}{c}{\textbf{Qwen3-4B}} \\
\cmidrule(lr){2-3}\cmidrule(lr){4-5}
\textbf{Method} & \MF & \CR & \MF & \CR \\
\midrule
\multicolumn{5}{@{}l}{\textbf{Math}} \\
\addlinespace[2pt]
Base & 0.53 & 100.0 & 0.51 & 100.0 \\
SFT with smaller lr & 0.64 \(\pm\) 0.01 & 72.4 \(\pm\) 0.1 & 0.67 \(\pm\) 0.01 & 97.3 \(\pm\) 1.9 \\
LoRA SFT & 0.73 \(\pm\) 0.07 & 99.0 \(\pm\) 1.0 & 0.62 \(\pm\) 0.05 & 100.1 \(\pm\) 0.2 \\
KL SFT & 0.63 \(\pm\) 0.02 & 90.6 \(\pm\) 0.0 & 0.66 \(\pm\) 0.01 & 97.5 \(\pm\) 0.2 \\
Task-mixed SFT & 0.63 \(\pm\) 0.01 & 94.6 \(\pm\) 0.2 & 0.75 \(\pm\) 0.07 & 100.2 \(\pm\) 0.2 \\
DFW-matched SFT & 0.69 \(\pm\) 0.02 & 72.2 \(\pm\) 0.2 & 0.71 \(\pm\) 0.01 & 97.4 \(\pm\) 1.6 \\
Query Probe & 0.70 \(\pm\) 0.01 & 100 & 0.71 \(\pm\) 0.01 & 100 \\
\textbf{\rlvr} & \textbf{0.84 \(\pm\) 0.01} & \textbf{100.3 \(\pm\) 0.0} & \textbf{0.85 \(\pm\) 0.01} & \textbf{102.8 \(\pm\) 0.6} \\
\midrule
\multicolumn{5}{@{}l}{\textbf{Science}} \\
\addlinespace[2pt]
Base & 0.46 & 100.0 & 0.42 & 100.0 \\
SFT with smaller lr & 0.58 \(\pm\) 0.02 & 84.5 \(\pm\) 2.6 & 0.60 \(\pm\) 0.06 & 88.6 \(\pm\) 1.3 \\
LoRA SFT & 0.51 \(\pm\) 0.04 & 100.0 \(\pm\) 1.2 & 0.54 \(\pm\) 0.03 & 99.5 \(\pm\) 0.1 \\
KL SFT & 0.60 \(\pm\) 0.00 & 92.0 \(\pm\) 2.2 & 0.59 \(\pm\) 0.05 & 90.4 \(\pm\) 0.1 \\
Task-mixed SFT & 0.55 \(\pm\) 0.01 & 82.0 \(\pm\) 3.0 & 0.61 \(\pm\) 0.06 & 96.4 \(\pm\) 0.2 \\
DFW-matched SFT & 0.59 \(\pm\) 0.02 & 82.1 \(\pm\) 2.9 & 0.65 \(\pm\) 0.04 & 89.0 \(\pm\) 0.9 \\
Query Probe & 0.55 \(\pm\) 0.01 & 100 & 0.56 \(\pm\) 0.01 & 100 \\
\textbf{\rlvr} & \textbf{0.66 \(\pm\) 0.02} & \textbf{101.3 \(\pm\) 1.8} & \textbf{0.67 \(\pm\) 0.02} & \textbf{101.5 \(\pm\) 0.2} \\
\bottomrule
\end{tabular*}
\end{table}

\subsection{Training-Free Baselines}
\label{app:training-free-baselines}

We describe the baselines in the order used in Table~\ref{tab:training-free-full}, grouped by the task outputs available when making the assessment. Baseline references are included in the table.

\paragraph{Query-based baselines.}
These methods assess a query before generating a task answer. RLVR is the trained reference: it generates a CSA rationale and a binary decision. The remaining methods do not update model parameters. We describe the implemented scoring rules below, retaining the method names used in Table~\ref{tab:training-free-full}.

\emph{Internal confidence.} We prompt the model to indicate whether it can answer correctly using Yes or No, without solving the problem. At each of the last six prompt positions, we apply the model's output head to each returned hidden state, including the embedding state and transformer-layer states. We normalize the Yes and No logits to obtain $q_{t,l}=\exp(z^{\mathrm{Yes}}_{t,l})/[\exp(z^{\mathrm{Yes}}_{t,l})+\exp(z^{\mathrm{No}}_{t,l})]$. Internal confidence averages these values with normalized Gaussian weights that favor the final token and final layer. With locality parameter 1, the unnormalized weight is $\exp[-(T-t)^2-(L-l)^2]$. No Yes/No response needs to be generated.

\emph{P(Yes).} This uses only $q_{T,L}$, the normalized Yes probability at the final prompt position and final layer. It measures the model's explicit pre-answer judgment of its own ability.

\emph{P(Yes)-Mean.} This takes the uniform mean of $q_{t,l}$ over the last six prompt positions and all returned hidden states. Unlike internal confidence, it does not emphasize later positions or layers.

The next four scores use the likelihood of the input question itself, with answer options appended for Science. Let $\ell_t=\log p(x_t\mid x_{<t})$ for input tokens after the first. The implementation floors probabilities at $10^{-8}$ and orients all scores so that larger values indicate greater confidence.

\emph{Max token surprisal.} We measure uncertainty using the largest input-token surprisal, $\max_t(-\ell_t)$. The confidence score is its negative, $-\max_t(-\ell_t)=\min_t\ell_t$, so larger scores indicate greater confidence. This tests whether any input token is particularly unexpected to the model.

\emph{Predictive entropy.} The implemented score is $\sum_t\ell_t$, the negative total surprisal of the observed question. This is an input-likelihood score, not a full-vocabulary entropy calculation, and it is sensitive to question length.

\emph{Min-K.} We select the five input tokens with the lowest log probabilities, or all available scored tokens if fewer than five exist. The score is the negative perplexity of this subset, $-\exp[-\operatorname{mean}_{t\in I_5}\ell_t]$. Here the five-token subset is unrelated to the number of sampled task answers used to construct capability labels.

\emph{Perplexity.} We use negative question perplexity, $-\exp[-\operatorname{mean}_t\ell_t]$. Unlike Min-K, this averages over all scored input tokens; unlike the predictive-entropy score above, it normalizes by input length.

\paragraph{Single-answer-based baselines.}
These methods generate one task answer before assessing it. Let $a_1,\ldots,a_m$ denote the generated solution tokens.

\emph{Sequence log-probability.} The confidence score is the geometric mean of the generated tokens' conditional probabilities, $\exp[m^{-1}\sum_j\log p(a_j\mid x,a_{<j})]$. This measures average likelihood of the sampled solution rather than its verified correctness.

\emph{Minimum token probability.} The score is $\min_j p(a_j\mid x,a_{<j})$. It treats the least likely generated token as a signal of uncertainty, in contrast to averaging likelihood over the entire solution.

\emph{P(True).} After producing a solution, the model receives the question and candidate answer and judges whether the answer is correct. The probability assigned to the affirmative judgment is used as confidence. This assesses a particular generated answer, whereas P(Yes) above asks about capability before an answer is available.

\emph{Verbalized confidence.} The model is prompted to provide an answer together with a numerical probability that it is correct. We parse the stated probability and use it as the score, rather than reading confidence directly from token likelihoods.

\emph{Answer probability.} We suppress reasoning and score the answer to which the model directly commits. For Science, the score is the largest option-letter probability after normalization over valid choices. For Math, it is the length-normalized likelihood of the directly generated answer span. It therefore excludes the likelihood of a full reasoning trace.

\paragraph{Multi-answer-based baselines.}
Both methods use $K=3,5,10$ independently sampled task solutions. Here $K$ controls the evidence available to the assessment method.

\emph{Self-consistency.} We extract a final answer from each solution and use the number of votes for the most frequent valid answer divided by $K$. Greater agreement yields higher confidence. Unparseable generations remain in the denominator but do not count as valid votes.

\emph{Semantic entropy.} We group the same samples by final-answer equivalence, using exact-match answer strings for Math and option letters for Science. With empirical cluster proportions $f_j$, we compute $H=-\sum_j f_j\log f_j$ and use $1-H/\log K$ as confidence. Concentrated answers give high confidence; dispersed answers give low confidence. Unparseable generations form an additional cluster. This implementation uses answer equivalence rather than an entailment model, and agreement among unparseable outputs can inflate its score.

\paragraph{Simple reference strategies.}
\emph{Always SELF-SOLVE} assigns every query to the local model, while \emph{Always DELEGATE} assigns every query to delegation. They show how the label distribution affects constant-decision Macro-F1. \emph{Random @ matched SSR} assigns decisions randomly at the target self-solve rate, providing a reference that matches decision frequency without using query information.

\paragraph{Evaluation setting.}
Figure~\ref{fig:training-free-efficiency} reports Qwen3-4B Math. Table~\ref{tab:training-free-full} reports all four model-domain settings, including $K=3,5,10$ for self-consistency and semantic entropy. Figure~\ref{fig:training-free-efficiency} uses the $K=5$ results and reports \MF{} as a percentage.
Solutions use temperature 1.0, top-$p$ 1.0, no top-$k$ truncation, and a maximum length of 4096 tokens. The auxiliary read-outs disable thinking. Direct-answer and verbalized-confidence generations use greedy decoding with maximum lengths of 64 and 128 tokens, respectively.

\paragraph{Labels and thresholds.}
Training-free baselines are evaluated against the base model's capability labels, and RLVR is evaluated against its own post-training labels. Query-based and single-answer-based baselines and simple strategies use the $K=5$ evaluation labels. For each model, domain, method, and $K$, we select the threshold that maximizes Macro-F1 on 500 calibration questions sampled from the training split with seed 42. Calibration questions are deduplicated and exclude all test questions. The selected threshold is then applied unchanged to the test set. The random reference's self-solve rate is also set from calibration labels. Unresolved read-outs receive \delegate\ decisions. Unavailable \CDS\ results are marked with a dash.

\begin{table}[!t]
\centering\small
\setlength{\tabcolsep}{0pt}
\renewcommand{\arraystretch}{1.03}
\caption{Training-free baselines across model-domain settings. Entries are \MF{} / \CDS{} on the $[0,1]$ M-F1 scale. Query-based and single-answer-based baselines and simple strategies use the $K=5$ evaluation labels. For self-consistency and semantic entropy, $K$ specifies the number of sampled solutions.}
\label{tab:training-free-full}

\begin{tabular*}{\textwidth}{@{\extracolsep{\fill}}lcccc@{}}
\toprule
& \multicolumn{2}{c}{Qwen3-4B} & \multicolumn{2}{c}{OLMo2-7B} \\
\cmidrule(lr){2-3}\cmidrule(l){4-5}
Method & Math & Science & Math & Science \\
\midrule
\multicolumn{5}{l}{\textbf{Query-based baselines}} \\
\addlinespace[2pt]
RLVR (ours) & 0.80 / 17.17 & 0.64 / 8.07 & 0.89 / 26.60 & 0.69 / 10.94 \\
Internal confidence~\citep{chen2026querylevel} & 0.57 / 2.89 & 0.51 / 1.15 & 0.56 / 3.31 & 0.63 / 6.85 \\
P(Yes)~\citep{chen2026querylevel} & 0.52 / 1.26 & 0.53 / 1.72 & 0.55 / 2.89 & 0.63 / 6.46 \\
P(Yes)-Mean~\citep{chen2026querylevel} & 0.43 / -3.62 & 0.50 / 0.02 & 0.66 / 7.05 & 0.63 / 7.52 \\
Max token surprisal~\citep{chen2026querylevel} & 0.44 / -2.75 & 0.50 / 0.83 & 0.52 / 2.35 & 0.60 / 5.30 \\
Predictive entropy~\citep{chen2026querylevel} & 0.52 / 0.74 & 0.54 / 1.92 & 0.57 / 3.66 & 0.53 / 1.52 \\
Min-K~\citep{chen2026querylevel} & 0.45 / -2.49 & 0.55 / 2.57 & 0.50 / 1.85 & 0.53 / 1.78 \\
Perplexity~\citep{chen2026querylevel} & 0.32 / -8.61 & 0.52 / 1.20 & 0.36 / -3.93 & 0.42 / -1.28 \\
\midrule
\multicolumn{5}{l}{\textbf{Single-answer-based baselines}} \\
\addlinespace[2pt]
Sequence log-probability~\citep{malinin2021uncertainty} & 0.76 / 12.30 & 0.61 / 6.86 & 0.72 / 10.46 & 0.59 / 5.66 \\
Min. token probability~\citep{manakul2023selfcheckgpt} & 0.76 / 11.57 & 0.58 / 4.17 & 0.72 / 10.48 & 0.68 / 9.23 \\
P(True)~\citep{kadavath2022llms-mostly-know} & 0.59 / 3.95 & 0.63 / 6.98 & 0.68 / 8.55 & 0.61 / 6.01 \\
Verbalized confidence~\citep{xiong2023can-llm-express-uncertainty} & 0.59 / 6.48 & 0.57 / 3.56 & 0.47 / -0.91 & 0.58 / 6.39 \\
Answer probability~\citep{kadavath2022llms-mostly-know} & 0.59 / 3.56 & 0.54 / 5.01 & 0.68 / 9.13 & 0.67 / 9.22 \\
\midrule
\multicolumn{5}{l}{\textbf{Multi-answer-based baselines}} \\
\addlinespace[2pt]
\multicolumn{5}{l}{Self-consistency~\citep{wang2023selfconsistency}} \\
\quad $K=3$ & 0.78 / 14.21 & 0.76 / 14.58 & 0.81 / 17.21 & 0.71 / 11.03 \\
\quad $K=5$ & 0.78 / 13.75 & 0.75 / 14.84 & 0.81 / 15.95 & 0.76 / 13.83 \\
\quad $K=10$ & 0.74 / 10.31 & 0.77 / 18.45 & 0.84 / 18.20 & 0.79 / 20.44 \\
\multicolumn{5}{l}{Semantic entropy~\citep{farquhar2024detecting-hallucinations-entropy}} \\
\quad $K=3$ & 0.76 / 11.48 & 0.76 / 14.58 & 0.81 / 17.21 & 0.71 / 10.71 \\
\quad $K=5$ & 0.75 / 12.14 & 0.75 / 14.84 & 0.82 / 17.11 & 0.75 / 13.36 \\
\quad $K=10$ & 0.74 / 9.83 & 0.78 / 17.39 & 0.84 / 18.53 & 0.79 / 17.64 \\
\midrule
\multicolumn{5}{l}{\textbf{Simple reference strategies}} \\
\addlinespace[2pt]
Always SELF-SOLVE & 0.43 / -- & 0.41 / -- & 0.39 / -- & 0.22 / -- \\
Always DELEGATE & 0.20 / -- & 0.24 / -- & 0.26 / -- & 0.42 / -- \\
Random @ matched SSR & 0.47 / -- & 0.50 / -- & 0.46 / -- & 0.49 / -- \\
\bottomrule
\end{tabular*}
\par\smallskip
\begin{minipage}{\textwidth}\small
-- indicates unavailable results.
\end{minipage}
\end{table}

\subsection{\csa\ Inference Efficiency}
\label{app:csa-efficiency}

\begin{table}[t]
\centering
\begin{minipage}{0.62\linewidth}
  \caption{CSA generation efficiency across model scales and domains. Time is mean generation latency per example at batch size 1. Tokens are generated completion tokens per example.}
  \label{tab:csa-generation-efficiency}
  \vspace{4pt}
  \small
  \setlength{\tabcolsep}{3pt}
    \begin{tabular*}{\linewidth}{@{}l@{\extracolsep{\fill}}cccc@{}}
      \toprule
      & \multicolumn{2}{c}{\textbf{Math}} & \multicolumn{2}{c}{\textbf{Science}} \\
      \cmidrule(lr){2-3} \cmidrule(lr){4-5}
      \textbf{Qwen3} & \textbf{Time (s)} & \textbf{Tokens} & \textbf{Time (s)} & \textbf{Tokens} \\
      \midrule
      1.7B  & 0.53 & 304.9 & 0.93 & 540.2 \\
      4B    & 0.39 & 150.7 & 0.47 & 184.9 \\
      8B    & 0.74 & 198.4 & 1.32 & 359.9 \\
      14B   & 0.93 & 156.7 & 1.16 & 195.2 \\
      \bottomrule
    \end{tabular*}
\end{minipage}
\end{table}

Table~\ref{tab:csa-generation-efficiency} shows that \csa\ assessments are fast in the measured settings: mean latency ranges from 0.39 to 1.32 seconds per query across four Qwen3 scales and two domains, with all Math settings below one second. We measure inference using vLLM on a single NVIDIA B200 GPU with batch size 1. Each query produces one non-thinking CSA response, with prefix caching disabled. Time is the mean wall-clock duration of generation requests after warm-up and excludes model loading and subsequent task solving. Generated-token counts come directly from vLLM token IDs and exclude prompt tokens.

\subsection{Query-Level Capability Stability}
\label{app:capability-stability}

Capability Ratio measures aggregate problem-solving performance, but it can
conceal changes in which individual queries a model can solve. For example,
CSA training could improve performance on some queries while degrading it on
others, leaving average accuracy unchanged. We therefore evaluate whether CSA
training changes the model's query-level capability boundary.

\paragraph{Evaluation setting.}
The analysis covers eight model--evaluation--policy settings formed by two
evaluation domains (mathematics and code), two model sizes
(Qwen3-1.7B and Qwen3-4B), and two CSA policies for each model
(one trained on mathematics and one trained on science). The column
``CSA policy'' in Table~\ref{tab:query-level-stability} denotes the domain used
to train the CSA policy, while ``Evaluation'' denotes the domain on which
problem-solving capability is measured. For example,
``Code / Qwen3-1.7B / Math'' evaluates the coding capability of Qwen3-1.7B
after CSA training on mathematics.

The mathematics evaluation contains \(490\) queries and contributes
\(4\times490=1{,}960\) query--policy pairs. The coding evaluation contains
\(722\) queries and contributes \(4\times722=2{,}888\) pairs. The complete
analysis therefore contains \(4{,}848\) query--policy pairs.

For each setting, we compare the original base checkpoint with the
corresponding CSA-trained checkpoint. Both checkpoints use the same solving
prompt, sampling parameters, answer parser, and correctness criterion. A
mathematics response is correct when its extracted final answer matches the
reference answer. A coding response is correct when the generated program
passes the benchmark's unit tests.

\paragraph{Capability labels and high-sample probes.}
We construct the reported query-level capability labels from \(K=5\)
independent solution attempts per query and checkpoint. Under the any-correct
rule, a query receives the label \textsc{Self-Solve} if at least one of the
five attempts is correct and receives \textsc{Delegate} otherwise.

Because a binary label constructed from five stochastic attempts is itself
noisy, we additionally collect \(K=32\) attempts per query for both the base
and CSA-trained checkpoints. These high-sample probes are used to estimate the
label disagreement expected from finite sampling and to test for changes in
each query's underlying success probability. They are not used to redefine
the reported \(K=5\) capability labels.

\paragraph{Capability-label churn.}
Let \(y_q^{\mathrm{base}}\) and \(y_q^{\mathrm{CSA}}\) denote the \(K=5\)
capability labels assigned to query \(q\) by the base and CSA-trained
checkpoints. For an evaluation set containing \(N\) queries, we define
capability-label churn as
\begin{equation}
\mathrm{Churn}
=
\frac{1}{N}
\sum_{q=1}^{N}
\mathbf{1}
\left[
y_q^{\mathrm{base}}
\neq
y_q^{\mathrm{CSA}}
\right].
\label{eq:capability-churn}
\end{equation}
Churn is the fraction of queries that cross the binary capability boundary
after CSA training. For example, a churn value of \(0.05\) means that the
pre- and post-training labels differ on \(5\%\) of the evaluation queries.
Nonzero churn does not necessarily imply that the underlying capability
changed, because two finite stochastic probes of the same checkpoint can
produce different labels.

\paragraph{Cohen's \(\kappa\).}
We complement churn with Cohen's \(\kappa\), which measures agreement between
the base and CSA-trained labels after accounting for agreement expected from
their overall label frequencies. Let
\[
p_o = 1-\mathrm{Churn}
\]
be the observed fraction of queries on which the two checkpoints agree. Let
\(s_{\mathrm{base}}\) and \(s_{\mathrm{CSA}}\) denote the fractions of queries
labeled \textsc{Self-Solve} by the base and CSA-trained checkpoints,
respectively. The agreement expected from these marginal label frequencies is
\begin{equation}
p_e
=
s_{\mathrm{base}}s_{\mathrm{CSA}}
+
(1-s_{\mathrm{base}})(1-s_{\mathrm{CSA}}).
\label{eq:kappa-expected}
\end{equation}
Cohen's \(\kappa\) is
\begin{equation}
\kappa
=
\frac{p_o-p_e}{1-p_e}.
\label{eq:cohen-kappa}
\end{equation}

A value of \(\kappa=1\) denotes perfect agreement. A value of
\(\kappa=0\) means that the observed agreement equals the agreement expected
from the two checkpoints' marginal label frequencies. This correction is
useful when one label is common: if both checkpoints label almost every query
\textsc{Self-Solve}, their raw agreement can be high even when it contains
little query-specific information. In our experiments, mean raw agreement is
\(95.5\%\) and mean \(\kappa\) is \(0.895\), showing that the base and
CSA-trained checkpoints have highly similar query-level capability labels
even after accounting for their overall label distributions.

\paragraph{Capability Ratio.}
We also report Capability Ratio (CR) to measure aggregate capability
preservation:
\begin{equation}
\mathrm{CR}
=
\frac{\mathrm{Acc}_{\mathrm{CSA}}}
     {\mathrm{Acc}_{\mathrm{base}}}
\times 100\%,
\label{eq:stability-cr}
\end{equation}
where \(\mathrm{Acc}_{\mathrm{base}}\) and
\(\mathrm{Acc}_{\mathrm{CSA}}\) are the problem-solving accuracies of the two
checkpoints under the same evaluation protocol. Churn and \(\kappa\) measure
query-level stability, whereas CR measures aggregate performance retention.

\paragraph{Finite-sampling null.}
The \(K=5\) capability label can change across repeated probes even when the
underlying checkpoint is unchanged. We estimate this finite-sampling
variability using the \(K=32\) probe of the base checkpoint.

For query \(q\), let \(c_q\) be the number of correct responses among the
\(32\) base-checkpoint attempts. Under the any-correct rule, the probability
that a random subset of five attempts receives a \textsc{Self-Solve} label is
\begin{equation}
\pi_q
=
1-
\frac{\binom{32-c_q}{5}}
     {\binom{32}{5}}.
\label{eq:self-solve-probability}
\end{equation}
The probability that two independently sampled \(K=5\) labels disagree is
then
\begin{equation}
\nu_q
=
2\pi_q(1-\pi_q).
\label{eq:query-noise-null}
\end{equation}
We define the sampling-noise null for an evaluation cell as
\begin{equation}
\mathrm{Null}
=
\frac{1}{N}
\sum_{q=1}^{N}
\nu_q,
\label{eq:cell-noise-null}
\end{equation}
and define excess churn as
\begin{equation}
\mathrm{Excess}
=
\mathrm{Churn}-\mathrm{Null}.
\label{eq:excess-churn}
\end{equation}
Positive excess churn indicates more pre/post label disagreement than expected
from finite sampling alone. Excess churn near or below zero indicates that the
observed disagreement is comparable to the variability of repeated \(K=5\)
measurements.

\paragraph{Bootstrap uncertainty.}
We estimate uncertainty in excess churn using a paired bootstrap over queries.
Within each evaluation cell, we sample \(N\) queries with replacement and
recompute both observed churn and the sampling-noise null on the same resampled
queries. We record their difference for each of
\(2000\) bootstrap replicates and use the \(2.5\)th and
\(97.5\)th percentiles of the resulting distribution as the paired-bootstrap
\(95\%\) confidence interval.

\begin{table}[t]
\centering
\caption{
Query-level capability stability after CSA training.
\textbf{CR} is the Capability Ratio.
\textbf{Churn} is the percentage of queries whose \(K=5\) capability label
differs between the base and CSA-trained checkpoints.
\textbf{Null} is the expected disagreement between two \(K=5\) probes of an
unchanged base checkpoint, estimated from \(K=32\) attempts per query.
\textbf{Excess} is churn minus the sampling-noise null, reported in percentage
points with a paired-bootstrap \(95\%\) confidence interval.
\(\boldsymbol{\kappa}\) is Cohen's chance-corrected agreement.
\textbf{BH discoveries} is the number of per-query comparisons rejected after
Benjamini--Hochberg correction at FDR \(0.05\).
}
\label{tab:query-level-stability}
\small
\setlength{\tabcolsep}{3.2pt}
\resizebox{\linewidth}{!}{
\begin{tabular}{lllrrrrlc}
\toprule
Evaluation
& Model
& CSA policy
& CR
& Churn (\%)
& Null (\%)
& Excess in pp (95\% CI)
& \(\kappa\)
& BH discoveries \\
\midrule

Code
& Qwen3-1.7B
& Math
& 102.7\%
& 4.6
& 6.3
& \(-1.7\;[-3.4,-0.1]\)
& 0.907
& 3 \\

&
&
Science
& 100.7\%
& 6.0
& 6.3
& \(-0.3\;[-2.1,+1.6]\)
& 0.879
& 7 \\

&
Qwen3-4B
& Math
& 100.2\%
& 4.8
& 5.4
& \(-0.6\;[-2.3,+1.1]\)
& 0.889
& 11 \\

&
&
Science
& 100.2\%
& 4.3
& 5.4
& \(-1.1\;[-2.8,+0.4]\)
& 0.902
& 11 \\

\addlinespace

Math
& Qwen3-1.7B
& Math
& 100.6\%
& 3.7
& 5.4
& \(-1.7\;[-3.4,+0.1]\)
& 0.909
& 0 \\

&
&
Science
& 102.6\%
& 6.3
& 5.4
& \(+0.9\;[-0.9,+2.9]\)
& 0.841
& 0 \\

&
Qwen3-4B
& Math
& 101.3\%
& 3.5
& 4.3
& \(-0.8\;[-2.2,+0.8]\)
& 0.903
& 0 \\

&
&
Science
& 99.7\%
& 2.7
& 4.3
& \(-1.6\;[-3.1,-0.0]\)
& 0.927
& 0 \\

\midrule

\multicolumn{3}{l}{\textbf{Mean: CSA policies} (\(8\) cells)}
& \textbf{101.0\%}
& \textbf{4.5}
& \textbf{5.4}
& \(\mathbf{-0.9}\)
& \textbf{0.895}
& \(\mathbf{32/4{,}848}\) \\

\bottomrule
\end{tabular}
}
\end{table}

\paragraph{Aggregate stability results.}
Table~\ref{tab:query-level-stability} shows that query-level capability remains
stable within the resolution of our evaluation. Across the eight settings,
mean observed label churn is \(4.5\%\), compared with \(5.4\%\) disagreement
expected from finite sampling. Mean excess churn is therefore \(-0.9\)
percentage points. No evaluation cell has a statistically positive excess over
its sampling-noise null. Agreement between the pre- and post-training labels
also remains high, with mean \(\kappa=0.895\), while the mean Capability Ratio
is \(101.0\%\).

\paragraph{Per-query changes in success probability.}
Label churn measures changes in a thresholded \(K=5\) capability label. We
additionally test whether CSA training changes each query's underlying success
probability using the \(K=32\) high-sample probes.

For each query--policy pair, we construct the following \(2\times2\)
contingency table:
\begin{equation}
\begin{array}{c|cc}
 & \mathrm{Correct} & \mathrm{Incorrect} \\
\hline
\mathrm{Base}
& c_q^{\mathrm{base}}
& 32-c_q^{\mathrm{base}} \\
\mathrm{CSA}
& c_q^{\mathrm{CSA}}
& 32-c_q^{\mathrm{CSA}}.
\end{array}
\label{eq:fisher-table}
\end{equation}
We apply a two-sided Fisher exact test to each of the \(4{,}848\)
query--policy pairs and control the false discovery rate at \(0.05\) using the
Benjamini--Hochberg procedure across the complete family of tests.

Only \(32\) of the \(4{,}848\) comparisons (\(0.66\%\)) are identified as
changes after correction. All \(32\) occur in the coding evaluation: \(3\) and
\(7\) for the two Qwen3-1.7B policies, and \(11\) for each Qwen3-4B policy.
None of the \(1{,}960\) mathematics comparisons is identified as a change.

\paragraph{Measurement resolution and limitations.}
The churn and \(\kappa\) analyses concern the binary capability labels induced
by the specified \(K=5\) aggregation rule. They do not establish that the
checkpoints have identical continuous success probabilities on every query.
Likewise, a per-query comparison that is not rejected should not be interpreted
as proof that the two success probabilities are equal. With \(32\) attempts per
checkpoint, small changes may remain unresolved.

Each evaluation cell contains approximately \(500\) to \(700\) queries. At
this sample size, the paired-bootstrap confidence intervals rule out aggregate
excess label churn larger than approximately \(2\)--\(3\) percentage points.
Our conclusion is therefore that CSA training preserves the model's
query-level capability boundary within the resolution of this evaluation.

\subsection{Within-Source Discrimination on Math}
\label{app:within-source-math}
Figure~\ref{fig:difficulty-routing} tests whether Math CSA decisions identify solvable queries beyond benchmark source identity. We evaluate the six Math-trained RLVR models from Table~\ref{tab:main} on the same 490 questions: 300 GSM8K, 100 MATH500, and 90 AIME. Each decision is matched to the same query's post-training capability label, constructed by any-correct aggregation over five task attempts. We exclude invalid decisions: three for OLMo2-13B and none for the other models.

For each model, we hold the capability labels fixed and independently permute the binary decisions within each source. Every permutation preserves that model's exact numbers of \selfsolve\ and \delegate\ decisions in each source. Thus, a preference for delegating AIME is preserved, while any association between a decision and the particular query's solvability within AIME is removed. We compute pooled Macro-F1 across the three sources after each permutation, rather than averaging three source-specific F1 scores.

We generate $B=100{,}000$ null draws with random seed 20260924. The implementation samples the number of true positives within each source from its conditional hypergeometric distribution, which is equivalent to uniformly shuffling binary decisions while fixing their counts. The blue markers in Figure~\ref{fig:difficulty-routing} show the null mean; the bars show its 2.5th and 97.5th percentiles. Orange markers show the observed score, and annotations give its improvement over the null mean in percentage points. These bars describe the permutation null, not confidence intervals for the observed score. A one-sided Monte Carlo $p$-value is $(1+\#\{\MF_b\geq\MF_{\mathrm{obs}}\})/(B+1)$. The saved analysis applies Holm correction jointly to the six Math tests and an additional exploratory Omni-MATH test; all six Math results remain below $0.002$ under this seven-test correction.

Table~\ref{tab:within-source-math} reports the numerical values underlying Figure~\ref{fig:difficulty-routing}. Observed Macro-F1 exceeds the source-preserving null by 3.44--9.61 percentage points. Source-dependent routing frequencies alone therefore do not explain the observed performance. This pooled test demonstrates discrimination within sources overall; it does not imply equally strong discrimination in every individual source.

\begin{table}[!htb]
\centering\small
\caption{Numerical results for Figure~\ref{fig:difficulty-routing}. M-F1 and null intervals are on the 0--100 scale; gains are percentage points computed before rounding. $N$ counts valid decisions. The 95\% interval describes the within-source permutation null. Adjusted $p$-values use the Holm correction described in this section.}
\label{tab:within-source-math}
\setlength{\tabcolsep}{3pt}
\renewcommand{\arraystretch}{1.12}
\begin{tabular*}{\textwidth}{@{\extracolsep{\fill}}lrccccc@{}}
\toprule
Model & $N$ & RLVR \MF & Null mean & 95\% null interval & Gain & Adjusted $p$ \\
\midrule
Qwen3-1.7B & 490 & 80.35 & 73.91 & [71.37, 76.42] & +6.44 & $7.00 \times 10^{-5}$ \\
Qwen3-4B & 490 & 80.09 & 70.48 & [67.48, 73.46] & +9.61 & $7.00 \times 10^{-5}$ \\
Qwen3-8B & 490 & 77.23 & 68.11 & [64.42, 71.53] & +9.11 & $7.00 \times 10^{-5}$ \\
Qwen3-14B & 490 & 75.32 & 67.74 & [64.74, 71.09] & +7.58 & $1.00 \times 10^{-4}$ \\
OLMo2-7B & 490 & 88.55 & 83.32 & [80.92, 85.69] & +5.23 & $7.00 \times 10^{-5}$ \\
OLMo2-13B & 487 & 88.56 & 85.12 & [82.85, 87.13] & +3.44 & $1.78 \times 10^{-3}$ \\
\bottomrule
\end{tabular*}
\end{table}

\subsection{Zero-Shot Transfer to Coding}\label{app:code-transfer}

We test whether \csa\ learned from math or science transfers to Python programming without further training. Our evaluation contains 722 problems from HumanEval+~\citep{chen2021humaneval,liu2023humaneval-plus-mbpp-plus}, MBPP+~\citep{austin2021mbpp,liu2023humaneval-plus-mbpp-plus}, and a LiveCodeBench~\citep{jain2025livecodebench} subset with organizer-provided difficulty labels. To reduce contamination risk, we include only LiveCodeBench contests held after the models' reported pretraining cutoff. We assess solution correctness by running each generated program against the benchmark's unit tests in a sandbox. Both trained policies are compared with the base model using the same \csa\ prompt. We also measure each checkpoint's coding ability separately, so its assessment decisions are evaluated against its own post-training performance.

\paragraph{Transfer performance and capability retention.}
As shown in Table~\ref{tab:code-transfer}, \csa\ training improves coding assessment: Macro-F1 rises from 0.36 to 0.64 for the 1.7B model and from 0.41 to 0.63 for the 4B model, with similar gains from math and science training. The transferred policies choose \selfsolve{} on 95--99\% of easy problems, but only 11--33\% of hard problems. The prompted base models, by comparison, delegate at most two of the 722 problems, even though they solve fewer than 17\% of hard problems. These assessment gains accompany Capability Ratios of 100--103\%, indicating that overall coding ability is retained. Formatting failures are also rare, occurring in only 0.5--1.1\% of attempts.

\paragraph{Does the policy identify which problems to delegate?}
An increase in Macro-F1 could result simply from delegating more often. To test whether the policies identify the right problems to delegate, we randomly shuffle each policy's decisions across problems within sources. This preserves the number of \selfsolve{} and \delegate{} decisions but removes their association with individual problems. As shown in Table~\ref{tab:code-transfer}, the policies' observed Macro-F1 scores are 7.8--10.6 standard deviations above this shuffled baseline ($p<0.001$). Thus, their gains reflect an ability to distinguish solvable from unsolvable problems beyond what a change in delegation frequency alone would achieve. This provides evidence that \csa\ transfers to coding, a domain not included in \csa\ training.

\paragraph{Reporting CDS.}
We omit CDS for the base models because they delegate only one or two problems. Such small delegation groups make the statistic unstable and unsuitable for comparison.
\begin{table}
\centering
\small
\caption{%
\textbf{Zero-shot \csa\ transfer to code.} Policies trained on math or science are
evaluated, without any further training, on 722 Python problems
(HumanEval+ / MBPP+ / LiveCodeBench). Mean over 3 decoding seeds. Seed-level sd
is $\leq 0.02$ for M-F1 and $\leq 0.9$ for $z_{\text{M-F1}}$.
\emph{Solve} is each checkpoint's own pass@1, re-measured per policy.
\emph{SSR} is the self-solve rate and $n_{\text{del}}$ the number of delegated
problems. $p_S$ / $p_D$ are solve rates within the queries a policy self-solves
and delegates. $z_{\text{M-F1}}$ scores M-F1 against an SSR-matched permutation
null. All four \csa\ cells have $p < 0.001$.
The base policy delegates on at most 2 of 722 problems, so $p_D$, CDS and
$z_{\text{M-F1}}$ are undefined for it.}
\label{tab:code-transfer}
\setlength{\tabcolsep}{4.5pt}
\resizebox{0.8\textwidth}{!}{%
\begin{tabular}{llrrrrrrrr}
\toprule
Model & \csa\ policy & Solve & SSR & $n_{\text{del}}$ & $p_S$ & $p_D$ & M-F1 & CDS & $z_{\text{M-F1}}$ \\
\midrule
\multirow{3}{*}{Qwen3-1.7B}
 & none (base) & 0.562 & 1.000 & 0.3 & 0.563 & --- & 0.36 & --- & --- \\
 & \csa-Math & 0.578 & 0.828 & 124 & 0.667 & 0.150 & \textbf{0.64} & 13.83 & \textbf{10.59} \\
 & \csa-Science & 0.566 & 0.873 & 92 & 0.630 & 0.127 & 0.58 & 12.78 & 8.98 \\
\midrule
\multirow{3}{*}{Qwen3-4B}
 & none (base) & 0.675 & 0.998 & 1.3 & 0.676 & --- & 0.41 & --- & --- \\
 & \csa-Math & 0.676 & 0.920 & 58 & 0.717 & 0.203 & 0.57 & 9.39 & 7.83 \\
 & \csa-Science & 0.676 & 0.888 & 81 & 0.738 & 0.182 & \textbf{0.63} & 12.10 & \textbf{9.95} \\
\bottomrule
\end{tabular}
}
\end{table}

\begin{table}
\centering
\small
\caption{%
\textbf{Transferred policies track code difficulty they were never shown.}
Self-solve rate and true solve rate per difficulty tier. Every \csa\ policy's SSR falls monotonically with the tier
(Spearman $=1.00$ against solve rate in all four cases), while the base policy
stays at ceiling even on hard problems it solves 8--17\% of the time.}
\label{tab:code-difficulty}
\resizebox{0.8\textwidth}{!}{%
\begin{tabular}{llcccccc}
\toprule
& & \multicolumn{2}{c}{Easy ($n{=}60$)} & \multicolumn{2}{c}{Medium ($n{=}60$)} & \multicolumn{2}{c}{Hard ($n{=}60$)} \\
\cmidrule(lr){3-4}\cmidrule(lr){5-6}\cmidrule(lr){7-8}
Model & \csa\ policy & SSR & Solve & SSR & Solve & SSR & Solve \\
\midrule
\multirow{3}{*}{Qwen3-1.7B}
 & none (base) & 0.999 & 0.648 & 1.000 & 0.183 & 1.000 & 0.083 \\
 & \csa-Math    & 0.950 & 0.673 & 0.317 & 0.117 & 0.111 & 0.083 \\
 & \csa-Science & 0.962 & 0.653 & 0.594 & 0.167 & 0.256 & 0.100 \\
\midrule
\multirow{3}{*}{Qwen3-4B}
 & none (base) & 1.000 & 0.761 & 1.000 & 0.317 & 0.978 & 0.167 \\
 & \csa-Math    & 0.991 & 0.761 & 0.800 & 0.317 & 0.333 & 0.183 \\
 & \csa-Science & 0.973 & 0.764 & 0.644 & 0.333 & 0.283 & 0.133 \\
\bottomrule
\end{tabular}
}
\end{table}

\subsection{\csa\ Is Model Specific}\label{app:model-specific}

To test whether \csa\ teaches a signal specific to the model carrying it, we compared each Qwen policy against OLMoE-1B-7B~\citep{muennighoff2025olmoe}, from a different lab with a different training corpus. We measured per-problem competence for both the policy's own model and OLMoE by sampling 32 solutions each on the same 722 coding problems and running the unit tests, so both sides are estimated with identical precision. The question is then whether a policy's decisions follow its own model's measured success rate or \csa\ is just tracking problem difficulty.

All four policies follow their own model as shown in Table~\ref{tab:crossfamily-specificity}. The partial correlation with own competence is consistently stronger than the correlation with OLMoE's, and every bootstrap confidence interval excludes zero, with effects ranging from +0.157 to +0.258. Table~\ref{tab:crossfamily-baseline} gives the same conclusion in operational terms: delegating the problems OLMoE is least likely to solve (using 32 executed samples per problem) is a worse policy than what \csa\ produces from the problem text alone, in all four cells. \csa\ also discriminates among problems that OLMoE rates as equally hard (residual-null z of 3.79 to 5.44). Two checks rule out the obvious artifacts: the competence estimates are near-equally reliable (split-half 0.985 / 0.990 / 0.958), and OLMoE's pass rates correlate substantially with the Qwen models' (r = 0.666 and 0.528), so it is a relevant comparator rather than a noisy or irrelevant one.

Table~\ref{tab:within-family-calibration} examines aggregate alignment when each of three \csa\ policies is paired with each solver. Policy coverage increases with the size of the model it was trained for (0.785, 0.891, and 0.892). Pairing the 8B policy with the 1.7B solver yields a coverage--accuracy gap of approximately 11 percentage points.
Matched pairings have a smaller mean absolute coverage--accuracy gap than mismatched pairings (0.049 versus 0.072). The 1.7B matched pair has a gap of +0.003, while the 4B and 8B policies have nearly identical coverage. This aggregate comparison is suggestive and does not establish query-level model specificity on its own.

\begin{table}[t]
\centering
\small
\caption{%
\textbf{\csa\ decisions track their own model's competence more than an
independent model's.}
$\Delta = r(\text{decision}, \text{own pass rate} \mid \text{OLMoE})
- r(\text{decision}, \text{OLMoE} \mid \text{own})$, computed over all 722 code
problems with bootstrap CIs averaged across 3 decoding seeds. Both models'
per-problem competence is measured the same way, by sampling $K{=}32$
solutions and running the unit tests, so neither side is favoured by having
the better-measured signal. All four policies are positive with CIs excluding
zero. Split-half reliability of the competence estimates is 0.985 / 0.990 /
0.958 (Qwen3-1.7B / Qwen3-4B / OLMoE), so the effect is not an artefact of a
noisier comparator.}
\label{tab:crossfamily-specificity}
\setlength{\tabcolsep}{5pt}
\resizebox{\textwidth}{!}{%
\begin{tabular}{llccccc}
\toprule
Policy model & \csa\ policy & $r(\text{dec},\text{own}\mid\text{OLMoE})$ &
$r(\text{dec},\text{OLMoE}\mid\text{own})$ & $\Delta$ & 95\% CI &
$r(\text{own},\text{OLMoE})$ \\
\midrule
\multirow{2}{*}{Qwen3-1.7B}
 & \csa-Math    & 0.256 & 0.093 & $+0.163$ & $[+0.079, +0.240]$ & 0.666 \\
 & \csa-Science & 0.224 & 0.067 & $+0.157$ & $[+0.080, +0.225]$ & 0.666 \\
\midrule
\multirow{2}{*}{Qwen3-4B}
 & \csa-Math    & 0.280 & 0.042 & $+0.238$ & $[+0.171, +0.304]$ & 0.528 \\
 & \csa-Science & 0.322 & 0.063 & $+0.258$ & $[+0.181, +0.333]$ & 0.528 \\
\bottomrule
\end{tabular}
}
\end{table}

\begin{table}[t]
\centering
\small
\caption{%
\textbf{\csa\ beats a difficulty rule built from an independent model.}
The rule delegates exactly as many problems as the \csa\ policy, choosing the
ones OLMoE is least likely to solve (measured, $K{=}32$).
$z_{\text{residual}}$ scores \csa\ against a permutation null confined to bins of
OLMoE's pass rate, so it measures discrimination among problems an independent
model rates as equally hard. \csa\ wins in all four cells despite seeing only the
problem text, while the rule uses 32 executed solution samples per problem.}
\label{tab:crossfamily-baseline}
\setlength{\tabcolsep}{5pt}
\resizebox{\textwidth}{!}{%
\begin{tabular}{llccccc}
\toprule
Policy model & \csa\ policy & M-F1 \csa\ & M-F1 OLMoE rule & $\Delta$ &
AUC (OLMoE score) & $z_{\text{residual}}$ \\
\midrule
\multirow{2}{*}{Qwen3-1.7B}
 & \csa-Math    & \textbf{0.64} & 0.58 & $+0.06$ & 0.835 & 4.53 \\
 & \csa-Science & \textbf{0.58} & 0.52 & $+0.06$ & 0.836 & 3.79 \\
\midrule
\multirow{2}{*}{Qwen3-4B}
 & \csa-Math    & \textbf{0.57} & 0.52 & $+0.05$ & 0.795 & 3.97 \\
 & \csa-Science & \textbf{0.63} & 0.58 & $+0.05$ & 0.794 & 5.44 \\
\bottomrule
\end{tabular}
}
\end{table}

\begin{table}[t]
\centering
\small
\caption{\textbf{Cross-scale transfer of \csa\ models.} Rows are \csa\ policies
trained for one Qwen3 solver, applied to all three. The gap is coverage minus
the solver's accuracy (parenthesised). Matched pairs (bold) have a smaller mean absolute gap ($0.049$ vs.\ $0.072$ mean $|$gap$|$).}
\label{tab:within-family-calibration}
\resizebox{0.7\textwidth}{!}{%
\begin{tabular}{lcccc}
\toprule
& & \multicolumn{3}{c}{Coverage--accuracy gap vs.\ solver} \\ \cmidrule(lr){3-5}
\csa\ Policy (trained for) & Coverage & 1.7B ($0.782$) & 4B ($0.818$) & 8B ($0.820$) \\
\midrule
\csa\ for Qwen3-1.7B & 0.785 & $\mathbf{+0.003}$ & $-0.034$ & $-0.036$ \\
\csa\ for Qwen3-4B & 0.891 & $+0.110$ & $\mathbf{+0.073}$ & $+0.071$ \\
\csa\ for Qwen3-8B & 0.892 & $+0.110$ & $+0.074$ & $\mathbf{+0.072}$ \\
\midrule
\multicolumn{2}{l}{Mean $|$gap$|$} & \multicolumn{3}{c}{diagonal $\mathbf{0.049}$ \quad vs. \quad off-diagonal $0.072$} \\
\bottomrule
\end{tabular}
}
\end{table}

\begin{figure}[t]
  \centering
  \includegraphics[width=0.96\textwidth]{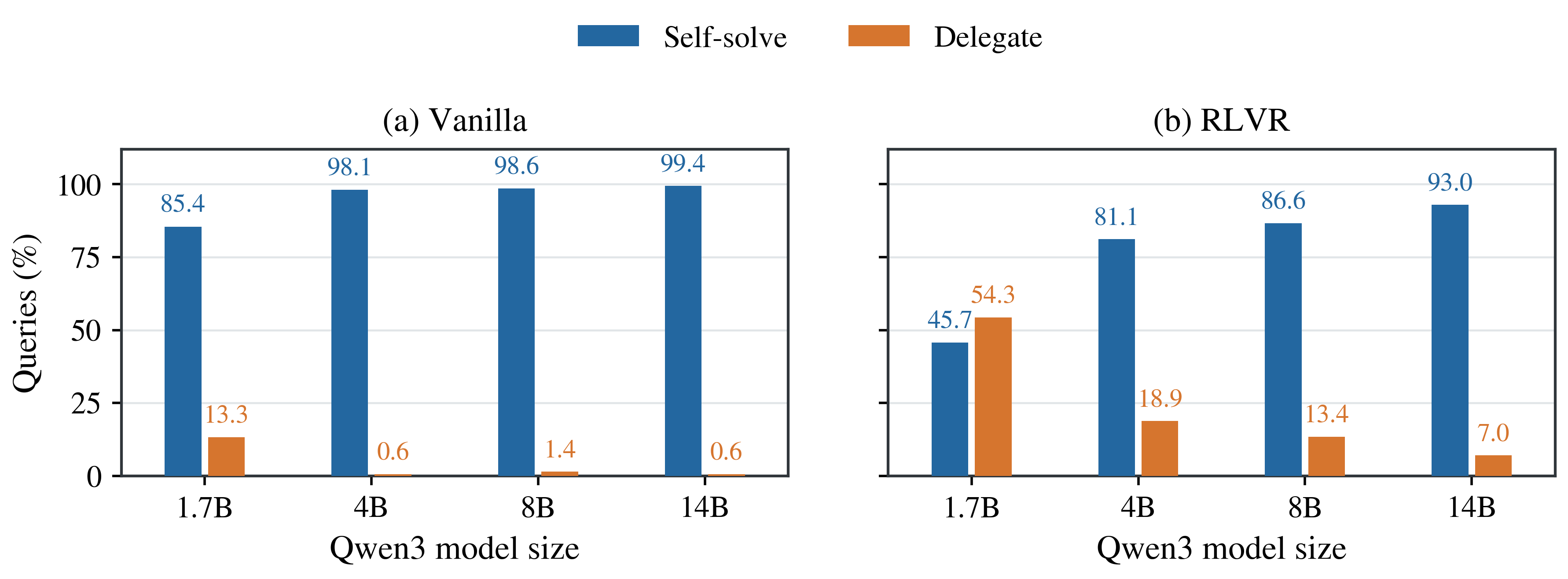}
  \caption{Qwen3 self-routing distributions on Science: (a) Vanilla (b) RLVR. Paired bars show self-solve and delegate rates among all 700 queries. Invalid Vanilla decisions (1.3\% for 1.7B and 4B) are excluded from both bars.}
  \label{fig:vanilla-vs-rlvr-dist}
\end{figure}

Figure~\ref{fig:vanilla-vs-rlvr-dist} further shows that RLVR decisions adapt to model capability: weaker models \delegate\ more, while stronger models \selfsolve\ more.
\FloatBarrier

\subsection{Case Studies}
\label{app:case-study-full}

To illustrate concretely how RLVR shapes \csa, we present case studies in Math and Science.

\paragraph{Math Problem}
\label{app:case-study-math}

\begin{figure}[ht]
  \centering
  \includegraphics[width=0.9\textwidth]{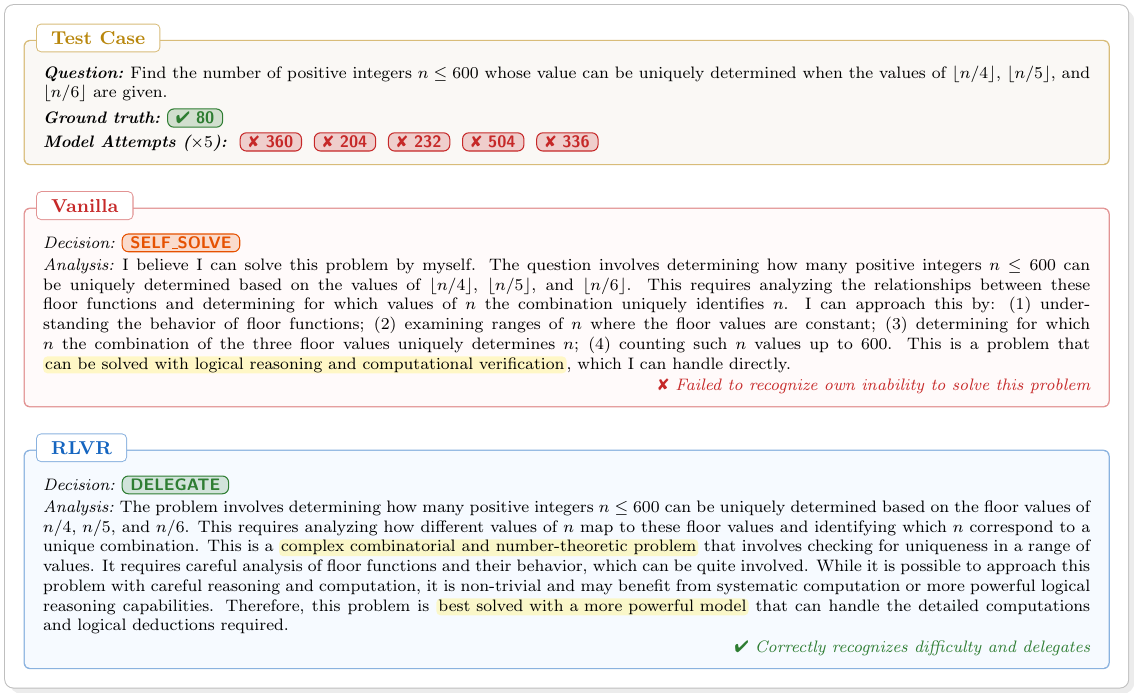}
  \caption{Full reasoning traces for the math case study: vanilla Qwen3-4B vs. RLVR-trained.}
  \label{fig:case-study-math-full}
\end{figure}

As shown in Figure \ref{fig:case-study-math-full}, we provide the complete reasoning traces for the math case study. The Vanilla Qwen3-4B model confidently chooses to \selfsolve, but produces an incorrect answer across all five sampled responses. In contrast, the RLVR-trained model recognizes that the problem is beyond its reliable capability and chooses to \delegate. This example illustrates how RLVR teaches the model to recognize when a query exceeds its competence.

\paragraph{Science Problem}
\label{app:case-study-science}

\begin{figure}[ht]
  \centering
  \includegraphics[width=0.9\textwidth]{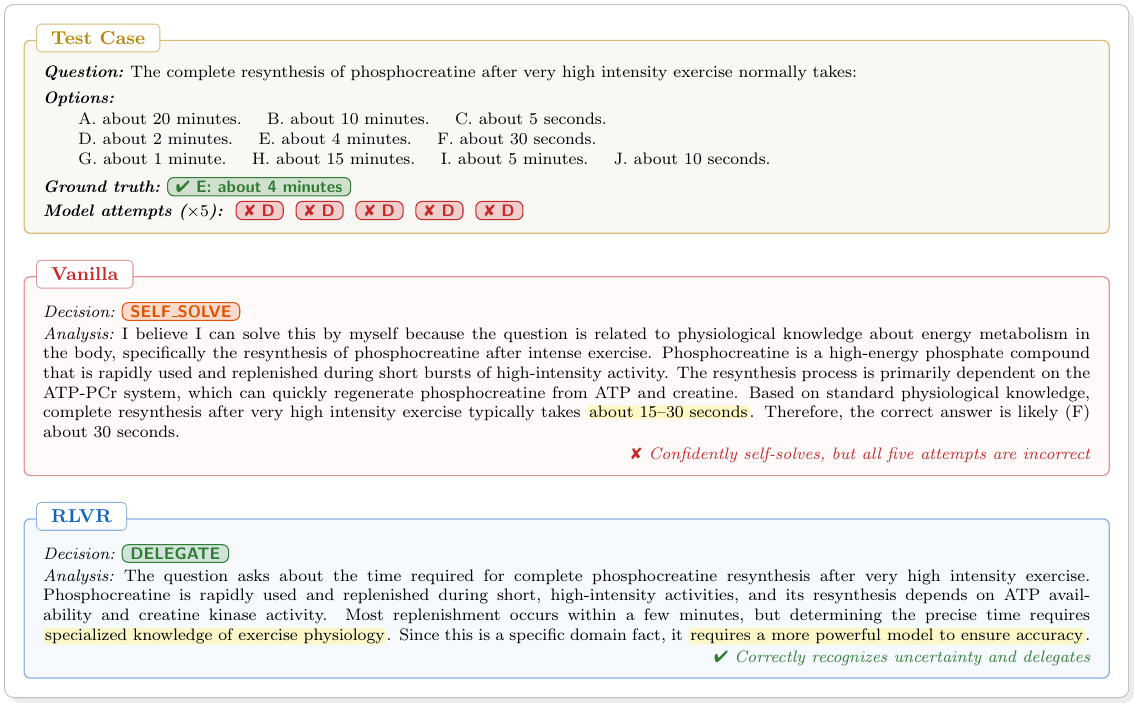}
  \caption{Case study on a science question: vanilla Qwen3-4B vs. RLVR-trained. }
  \label{fig:case-study-science}
\end{figure}

Figure \ref{fig:case-study-science} presents an additional case study on a science question, where the same pattern emerges: the vanilla model answers confidently but incorrectly, while the RLVR-trained model recognizes its uncertainty and delegates.

\subsection{Applications of \csa}\label{app:applications}

\subsubsection{Self-Routing}\label{app:self-routing}
\begin{figure}[ht]
    \centering
    \includegraphics[width=0.7\linewidth]{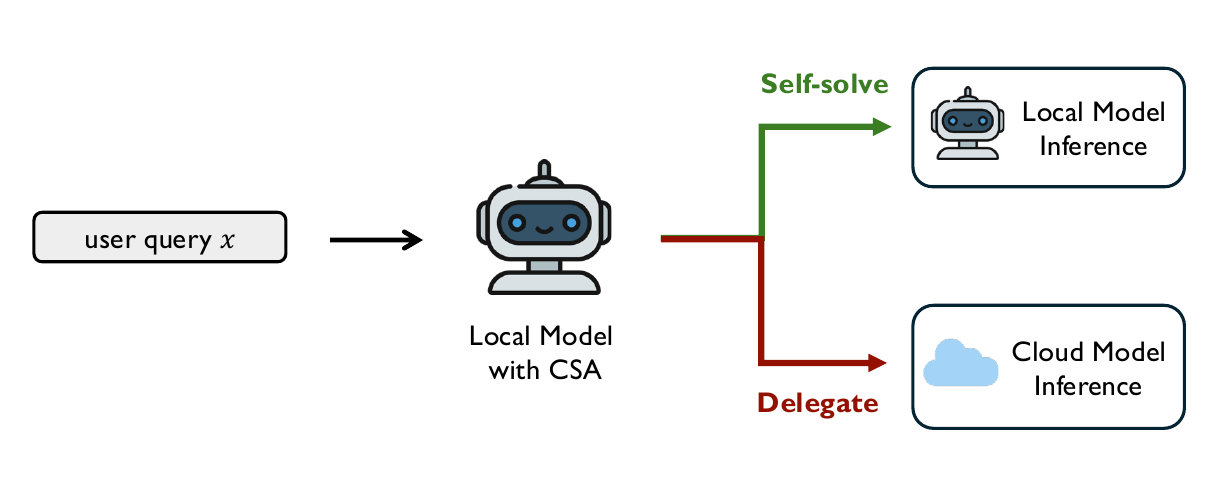}
    \caption{Illustration of \emph{self-routing}. Upon receiving a user query x, a local model equipped with \csa\  assesses whether the query falls within its reliable solving capability and decides between two actions: \selfsolve, where the local model answers directly, or \delegate, where the query is routed to a cloud model for inference.}
    \label{fig:self-routing}
\end{figure}

We provide illustrative figures for \emph{self-routing} in Figure \ref{fig:self-routing}. Formally, given a query $x$, the local model first produces a \csa\ decision $\hat{y} = \mathcal{M}_{\text{local}}^{\text{\csa}}(x) \in \{\selfsolve, \delegate\}$, and the routed answer is
\begin{equation}
\hat{a}(x) =
\begin{cases}
\mathcal{M}_{\text{local}}(x), & \hat{y} = \selfsolve, \\
\mathcal{M}_{\text{cloud}}(x), & \hat{y} = \delegate.
\end{cases}
\end{equation}
This setup exposes a direct cost--accuracy trade-off between the two extremes of local-only inference (cheap but limited by the local model's capability) and cloud-only inference (accurate but expensive on every query). An ideal \csa\ policy delegates exactly the queries the local model cannot solve, approaching cloud-only accuracy at a fraction of the cost.

\paragraph{Random routing at a matched cloud-call budget.}
For the Math routing comparison in Table~\ref{tab:routing-app}, random routing selects a uniformly random subset of exactly $m$ out of $N=490$ queries for cloud inference and answers the rest locally. We set $m$ to the corresponding CSA policy's cloud-call count: 96 for Qwen3-1.7B and 73 for Qwen3-4B. Let $L_i,C_i\in\{0,1\}$ indicate correctness of the stored single local and cloud answers to query $i$. The exact expected accuracy over random subsets is
\begin{equation}
\mathbb{E}[\operatorname{Acc}_{\mathrm{random}}]
=\frac{1}{N}\sum_{i=1}^{N}\left[(1-\rho)L_i+\rho C_i\right],
\qquad \rho=\frac{m}{N}.
\end{equation}
We use the same stored branch outputs to evaluate CSA and random routing. The expected random-routing accuracies are 0.6504 and 0.7182, compared with CSA accuracies of 0.7327 and 0.8000. These are exact expectations conditional on the recorded answers, not estimates from a single random subset. The comparison isolates query selection at the same cloud-call budget. The time and token measurements in Table~\ref{tab:routing-app} come from a separate generation run of each CSA model at batch size 1. They measure assessment overhead, excluding local and cloud answer generation. The routing accuracies and cloud-call counts come from the stored routing evaluation.

\subsubsection{Cost-Aware Routing as an Extension of RLVR-Trained \csa}
\label{sec:ca-routing}
\paragraph{Matching the single-attempt deployment setting.}
The Math any-correct label measures whether a model succeeds in at least one of five attempts. It does not represent single-attempt success: under independent attempts, a query with success probability $0.2$ receives a \selfsolve\ label with probability $1-(1-0.2)^5\approx0.67$. A deployed router that obtains only one answer from the selected solver instead benefits from predicting each action's expected outcome. The experiment below addresses this practical setting using an expected-outcome reward. Five sampled answers estimate each branch's single-call success probability. Evaluation averages these outcomes under the same routing protocol for every method, without selecting the best of five answers. We keep the local and cloud solvers fixed so that policy comparisons use the same branch outcomes.

\paragraph{From capability assessment to routing utility.}
The \csa\ policy decides whether a model should attempt a query or seek help. We extend this policy to a local--cloud setting in which the selected action is evaluated by the answer it produces and delegation incurs a cost~\citep{chen2024frugalgpt,ding2024hybrid,ong2025routellm,aggarwal2024automix}. The action space, assessment prompt, and RLVR training framework remain unchanged, and only the reward is replaced by downstream routing utility.
For a query $q$, let $p_{\mathrm{L}}(q)$ and $p_{\mathrm{C}}(q)$ denote the probabilities that the local and cloud models answer correctly. The utility of action $a$ is
\begin{equation}
\begin{cases}
    p_{\mathrm{L}}(q),
    & a=\selfsolve, \\
    p_{\mathrm{C}}(q)-\lambda C,
    & a=\delegate,
    \end{cases}
\label{eq:ca-routing}
\end{equation}
where $C$ is the cost of a cloud call and $\lambda$ controls the cost--accuracy trade-off. The corresponding oracle delegates exactly when
\begin{equation}
p_{\mathrm{C}}(q)-p_{\mathrm{L}}(q) > \lambda C.
\end{equation}
This objective generalizes the original \csa\ decision. If local capability is binary, delegation always succeeds, and delegation is free, Eq.~\eqref{eq:ca-routing} recovers the rule of delegating precisely when the local model cannot solve the query. In the more realistic setting, delegation is rewarded only when its expected improvement exceeds its cost.
\paragraph{Models and datasets.}
We use Qwen3-1.7B as the local model and Qwen3-32B as the cloud model. We evaluate on the same Science domain as in the main experiments, comprising the biology, chemistry, health, and physics categories of MMLU-Pro with 3,266 training and 700 test queries. We set $C=1$, so $\lambda$ directly represents the penalty assigned to one cloud call.
\paragraph{Estimating action outcomes.}
For each query, we sample $K=5$ answers independently from both the local and cloud models and compute empirical single-call success rates
\begin{equation}
\frac{1}{K}
\sum_{k=1}^{K}
\mathbf{1}
\left[
\hat{a}_{m}^{(k)}(q)=a^{*}(q)
\right],
\qquad
m\in{\mathrm{L},\mathrm{C}}.
\end{equation}
We use these pass rates in Eq.~\eqref{eq:ca-routing}.

\begin{figure}
    \centering
    \includegraphics[width=\linewidth]{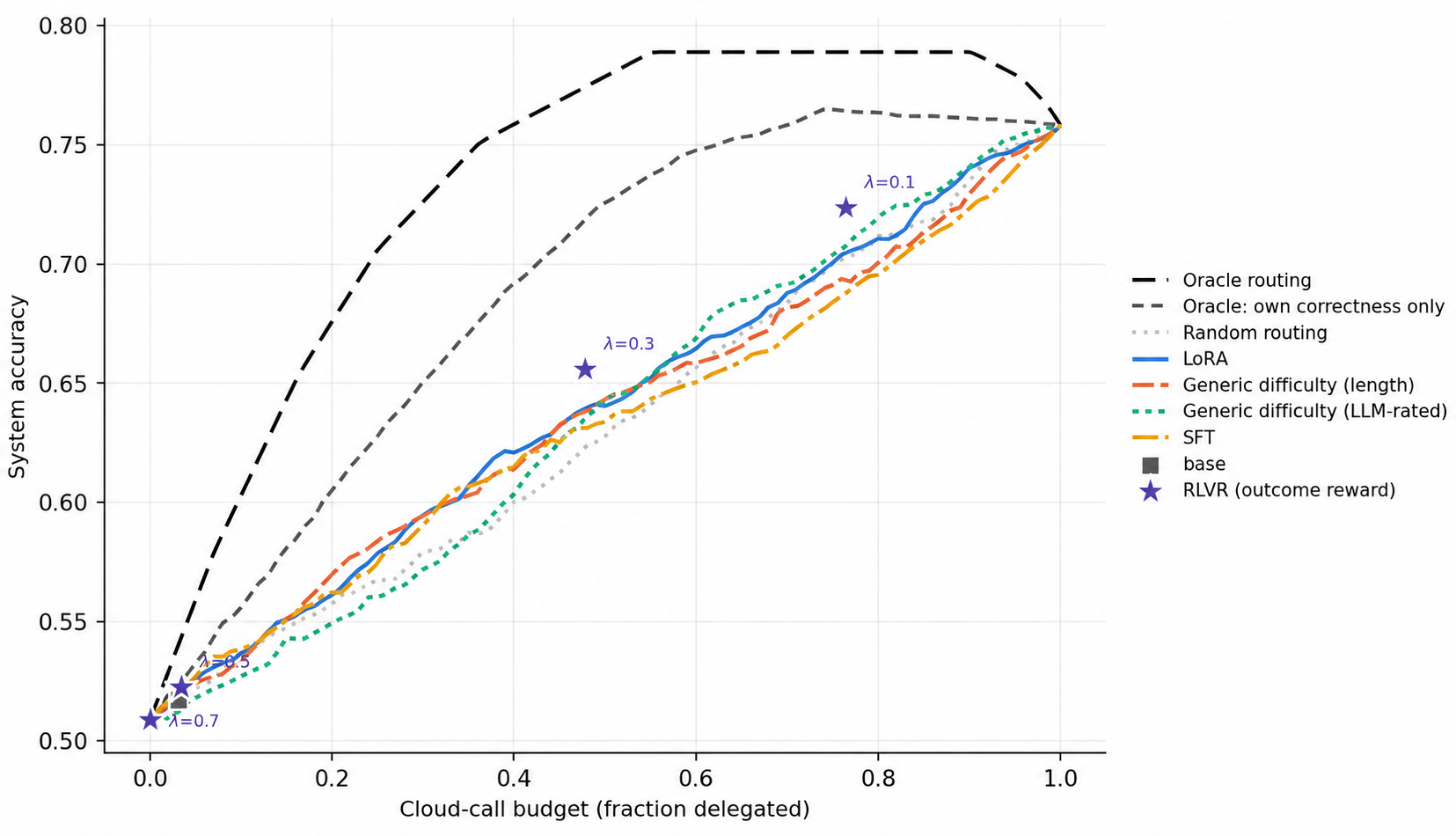}
    \caption{\textbf{Routing from Qwen3-1.7B to Qwen3-32B on science} ($n=700$). Accuracy as a function of the cloud-call budget for RLVR and the baselines. Stars denote RLVR policies trained with different values of $\lambda$. The dashed envelopes show true-gain oracle routing (upper) and oracle routing based only on local-model correctness (lower). Their separation captures the additional value of knowing whether the cloud model can improve on the local model. RLVR outperforms all deployable baselines.}
    \label{fig:pre-answer-cost-accuracy}
\end{figure}

For each value of $\lambda$, we apply the same two-stage RLVR procedure used for \csa. Diversity-Filtered Warm-up first retains training queries whose 16 policy rollouts contain both actions. Training then continues on the full distribution. GRPO uses 16 rollouts per prompt, a global batch size of 64, one epoch, a learning rate of $10^{-6}$, a KL coefficient of $0.01$, and full-parameter optimization with ZeRO-3 on four NVIDIA L40S GPUs.
During training, we use the empirical expected-outcome reward
\begin{equation}
\begin{cases}
\hat{p}_{\mathrm{L}}(q),
& a=\selfsolve, \\
\hat{p}_{\mathrm{C}}(q)-\lambda C,
& a=\delegate.
\end{cases}
\end{equation}
This is the expectation of a sampled correctness reward under the empirical branch outcomes. Using the expectation removes within-action sampling variance, so the GRPO advantage reflects the quality of the routing decision. Unparseable policy outputs receive a reward of $-1$, below the reward of either valid action.
\paragraph{Cost grid.}
Because $K=5$, each empirical success rate is quantized in increments of $0.2$. We use
\begin{equation}
\lambda \in \{0.1,0.3,0.5,0.7\},
\end{equation}
which places each cost between adjacent achievable outcome differences and avoids ties in Eq.~\eqref{eq:ca-routing}. We train a separate policy for each value and report its realized delegation rate rather than assuming that the grid produces evenly spaced operating points.
\paragraph{Evaluation protocol.}
We evaluate every routing policy using frozen copies of the local and cloud solvers. Freezing both branches keeps the outcome estimates stationary and ensures that differences between policies arise from which queries they delegate rather than changes in solving ability. Policy decisions are sampled at temperature $1.0$ over three random seeds.
Let $d_{\pi}(q)=1$ when policy $\pi$ delegates and $0$ otherwise. Its routed accuracy and delegation rate are
\begin{align}
\operatorname{Acc}(\pi)
&=
\frac{1}{N}
\sum_{q}
\left[
\bigl(1-d_{\pi}(q)\bigr)\hat{p}_{\mathrm{L}}(q)
+
d_{\pi}(q)\hat{p}_{\mathrm{C}}(q)
\right],
 \\
\rho(\pi)
&=
\frac{1}{N}
\sum_{q} d_{\pi}(q).
\label{eq:delegation-rate}
\end{align}
We report the resulting cost--accuracy frontier in Figure~\ref{fig:pre-answer-cost-accuracy}.

The figure shows that outcome-aware RLVR is the strongest deployable routing method for Qwen3-1.7B on science. The trained RLVR policies lie above the baseline cluster.

The two dashed curves clarify why outcome-aware training helps. The upper oracle knows the benefit of delegation, $p_{\mathrm{cloud}}-p_{\mathrm{local}}$, whereas the lower oracle knows only whether the local model will succeed. The local model fails on $49.1\%$ of queries, but the cloud model can rescue only $29.9$ percentage points. On the remaining $19.2$ points, both models fail. A capability-only signal cannot distinguish these cases and may therefore delegate queries for which a cloud call provides no benefit. By optimizing the outcomes of both actions, RLVR learns a substantially better routing policy.

The operating point still matters. With larger penalties, the policies converge to never delegating, partly because many training groups provide no decision-level learning signal. Nevertheless, the clear gains at matched delegation rates show that RLVR learns which queries benefit from cloud assistance substantially better than the available baselines.

\subsubsection{Self-Guided Data Selection}\label{app:data-selection}

\begin{figure}[ht]
    \centering
    \includegraphics[width=0.9\linewidth]{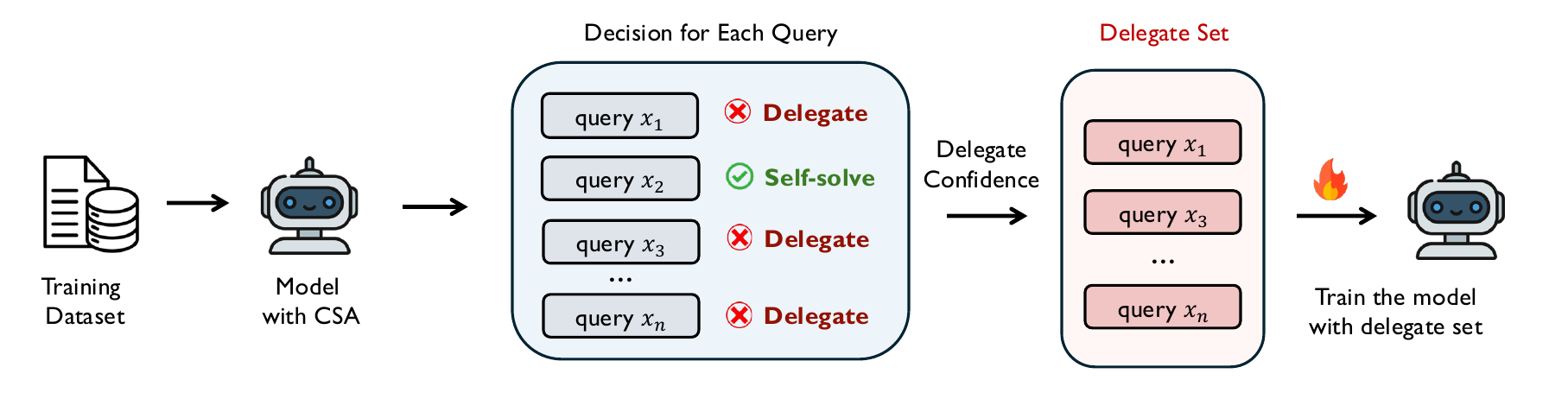}
    \caption{Overview of \csa-based data selection. A \csa-equipped model scores a fresh candidate pool, prioritizing examples it would delegate while filtering out extreme outliers. The selected subset is then used to fine-tune the model.}
    \label{fig:self-guided}
\end{figure}

Our main data-selection experiment tests whether examples preferred by an RLVR-trained \csa\ model are better fine-tuning data than examples selected by standard data-selection baselines. We begin with a base model \(M_0\) and train a \csa\ model \(M_{\mathrm{RLVR}}\) to decide whether a problem should be answered directly or delegated. After training, the model is used as a scorer: it is applied to a fresh candidate training pool
\begin{equation}
    \mathcal{D}_{\mathrm{pool}}=\{(x_i,y_i)\}_{i=1}^{N},
\end{equation}

and assigns each candidate a delegation-derived score. Let \(z_i^{\mathrm{self}}\) and \(z_i^{\mathrm{del}}\) denote the logits assigned by \(M_{\mathrm{RLVR}}\) to the \textsc{Self\_Solve} and \textsc{Delegate} decision tokens for example \(x_i\). We convert these logits into a normalized delegation probability
\begin{equation}
    p_i^{\mathrm{del}}
    =
    \frac{\exp(z_i^{\mathrm{del}})}
    {\exp(z_i^{\mathrm{del}})+\exp(z_i^{\mathrm{self}})}.
    \label{eq:delegate-prob}
\end{equation}

Intuitively, \(p_i^{\mathrm{del}}\) estimates how strongly the trained \csa\ policies believe that \(M_0\) should not solve \(x_i\) directly.

A naive delegate-only selector would choose examples with the largest \(p_i^{\mathrm{del}}\). However, this can over-select extreme examples that are far beyond the model's current capabilities or are noisy, ambiguous, or mismatched to the target distribution. To avoid this, we use a \emph{delegate-based loss-bandpass} selector. First, we compute the base model's supervised loss on each candidate:
\begin{equation}
    \ell_i
    =
    - \frac{1}{|y_i|}
    \sum_{t=1}^{|y_i|}
    \log p_{M_0}(y_{i,t}\mid x_i,y_{i,<t}).
    \label{eq:base-loss}
\end{equation}

We then retain only examples whose loss lies within a middle band:
\begin{equation}
    \mathcal{B}_{\alpha,\beta}
    =
    \left\{
    i :
    Q_{\alpha}(\{\ell_j\}_{j=1}^{N})
    \leq
    \ell_i
    \leq
    Q_{\beta}(\{\ell_j\}_{j=1}^{N})
    \right\},
    \label{eq:loss-band}
\end{equation}

where \(Q_{\alpha}\) and \(Q_{\beta}\) are empirical loss quantiles. This removes examples that are already too easy for \(M_0\) as well as extreme high-loss outliers. Among the remaining examples, we rank by delegation probability:
\begin{equation}
    s_i^{\mathrm{DLB}}
    =
    p_i^{\mathrm{del}}
    \cdot
    \mathbf{1}\{i\in\mathcal{B}_{\alpha,\beta}\}.
    \label{eq:delegate-loss-bandpass}
\end{equation}

For a target selection budget \(k\), the selected subset is
\begin{equation}
     \mathcal{S}_{\mathrm{DLB}}(k)
    =
    \operatorname{TopK}_{i\in\{1,\ldots,N\}}
    \left(s_i^{\mathrm{DLB}}, k\right).
    \label{eq:dlb-topk}
\end{equation}

We then fine-tune the model  on \(\mathcal{S}_{\mathrm{DLB}}(k)\). Figure~\ref{fig:self-guided} provides an overview of the method.

For the science-pool experiments, we train $M_0$ on MMLU-Pro~\citep{wang2024mmluprorobustchallengingmultitask} to get \(M_{\mathrm{RLVR}}\). The candidate pool is MedMCQA~\citep{pal2022medmcqa}. We randomly sample \(N=50{,}000\) examples from the MedMCQA training split and score them using our method and all baselines. We conduct experiments on Qwen3-1.7B and Qwen3-4B~\citep{yang2025qwen3technicalreport}. We use loss-bandpass quantiles $ Q_{\alpha} = 0.2, Q_{\beta} = 0.8$. We compare delegate-based loss-bandpass against random selection, high-loss selection, low-loss selection, and development-set similarity selection. We also include an all-samples condition, which fine-tunes on the full 50,000-example candidate pool. For development-set similarity, we reserve a held-out set
\begin{equation}
        \mathcal{D}_{\mathrm{dev}}=\{(x_j^{\mathrm{dev}},y_j^{\mathrm{dev}})\}_{j=1}^{m}
\end{equation}

of \(m=500\) randomly selected MedMCQA training examples that are excluded from the candidate pool.

The baseline selectors are defined as follows. Random selection samples a subset uniformly without replacement:
\begin{equation}
    \mathcal{S}_{\mathrm{rand}}(k)
    \sim
    \operatorname{Unif}
    \left(
    \{\mathcal{S}\subset \mathcal{D}_{\mathrm{pool}}:|\mathcal{S}|=k\}
    \right).
    \label{eq:random-selection}
\end{equation}

High-loss selection ranks examples by the base model loss in Eq.~\eqref{eq:base-loss} and selects the largest-loss examples:
\begin{equation}
    \mathcal{S}_{\mathrm{high}}(k)
    =
    \operatorname{TopK}_{i}
    \left(\ell_i,k\right).
    \label{eq:high-loss-selection}
\end{equation}

This baseline tests whether our method is simply performing hard-example mining. Low-loss selection selects the examples with the smallest base-model loss:
\begin{equation}
    \mathcal{S}_{\mathrm{low}}(k)
    =
    \operatorname{TopK}_{i}
    \left(-\ell_i,k\right).
    \label{eq:low-loss-selection}
\end{equation}

This baseline tests whether improvements can instead be obtained by emphasizing examples that are already easy and likely to be clean.

For development-set similarity, we embed each candidate \(x_i\) and each development example \(x_j^{\mathrm{dev}}\) using a fixed sentence encoder \(f(\cdot)\). We compute the maximum cosine similarity between each candidate and the held-out development set:
\begin{equation}
    s_i^{\mathrm{sim}}
    =
    \max_{1\leq j\leq m}
    \frac{
        f(x_i)^{\top} f(x_j^{\mathrm{dev}})
    }{
        \|f(x_i)\|_2 \, \|f(x_j^{\mathrm{dev}})\|_2
    }.
    \label{eq:dev-similarity-score}
\end{equation}

The selected subset is then
\begin{equation}
    \mathcal{S}_{\mathrm{sim}}(k)
    =
    \operatorname{TopK}_{i}
    \left(s_i^{\mathrm{sim}},k\right).
    \label{eq:dev-similarity-selection}
\end{equation}

This baseline controls for the possibility that delegate-based selection works only because it selects examples similar to the evaluation distribution.

Finally, the all-samples condition uses
\begin{equation}
    \mathcal{S}_{\mathrm{all}}=\mathcal{D}_{\mathrm{pool}},
    \label{eq:all-samples}
\end{equation}

and therefore measures the performance obtained when no data selection is performed. All selected subsets of size \(k\) are fine-tuned using the same optimization hyperparameters, number of epochs, and training budget, so differences in downstream performance can be attributed to the selection rule rather than to changes in compute or training procedure.

Given the scored candidate pool, we run fixed-budget SFT experiments. For each selection method, we fine-tune on one of three budget fractions of the 50,000-example pool: 5\%, 10\%, and 20\% corresponding to 2,500, 5,000, and 10,000 selected examples.  Fine-tuning is performed with full-parameter training on NVIDIA L40S nodes. The main hyperparameters are a learning rate of \(5\times 10^{-5}\), per-device batch size 4, maximum sequence length 4096, maximum training steps 2000, and bfloat16 training. Evaluation is performed on the MedMCQA validation split rather than the official test split. Each run evaluates on 500 randomly sampled validation examples with a maximum generation length of 256 tokens. The evaluation set is identical for all methods. The primary metric is validation accuracy after fine-tuning.

\section{Prompt Templates}
\label{app:prompts}

\subsection{\csa\ Prompt Template}
\label{app:csa_inference_template}

We use the following prompt template to elicit the model's \textsc{Self-Solve} / \textsc{Delegate} decision on each query. The template instructs the model to assess its own capability, and to emit its decision in a structured format consisting of an \texttt{<analysis>} block followed by a \texttt{<decision>} block. The same template is used for all subsequent \csa\  inference throughout the paper.

\begin{tcolorbox}[
    colback=gray!5,
    colframe=gray!100,
    title=\csa\  Prompt Template,
    fonttitle=\bfseries,
    boxrule=0.5pt,
    arc=2pt,
]
\small
\begin{verbatim}
Decide whether you can answer
the user's query.
- Choose SELF_SOLVE if you believe you can solve it by yourself.
- Choose DELEGATE if you believe it requires a more powerful model.

# Output Format
<analysis>
Explain why you chose SELF_SOLVE or DELEGATE.
</analysis>

<decision>
SELF_SOLVE or DELEGATE
</decision>

# Query:
{query}
\end{verbatim}
\end{tcolorbox}

We parse the model's response by extracting the content of the \texttt{<decision>} block and matching it against \textsc{Self-Solve} or \textsc{Delegate}.

\subsection{Self-Rationale Prompt (\sftself)}
\label{app:self-rationale-prompt}

The training model itself is prompted to produce a self-assessment justifying the ground-truth label. The placeholders \texttt{\{query\}} and \texttt{\{decision\}} are replaced with the query $q_i$ and the label $y_i \in \{\textsc{Self\_Solve}, \textsc{Delegate}\}$, respectively.

\begin{tcolorbox}[
    breakable,
    colback=gray!5,
    colframe=gray!100,
    title=Self-Rationale Prompt Template,
    fonttitle=\bfseries,
    boxrule=0.5pt,
    arc=2pt,
]
\small
\begin{verbatim}
You are given a user query, along with a **ground-truth label** that
correctly reflects your capability.

The label is:
- SELF_SOLVE: you can solve the query by yourself.
- DELEGATE: the query requires a more powerful model.

For the following query: {query}

The ground-truth label is: {decision}

Write a coherent self-assessment explaining
**why this query should be classified as {decision}**.

You do NOT need to actually solve the query.
\end{verbatim}
\end{tcolorbox}

\subsection{Teacher-Rationale Prompt (\sftteacher)}
\label{app:teacher-rationale-prompt}

A stronger teacher model $\mathcal{M}^{*}$ is prompted to produce a rationale on behalf of the target model. The teacher is instructed to write in the first-person perspective of the target model so that the resulting rationale stylistically resembles a self-assessment, even though it is generated externally.

\begin{tcolorbox}[
    breakable,
    colback=gray!5,
    colframe=gray!100,
    title=Teacher-Rationale Prompt Template,
    fonttitle=\bfseries,
    boxrule=0.5pt,
    arc=2pt,
]
\small
\begin{verbatim}
You are analyzing the behavior of a target language model.

You are given a user query, along with a **ground-truth label** that
correctly reflects the target model's capability:
- SELF_SOLVE: the target model can solve the query by itself.
- DELEGATE: the query requires a more powerful model.

For the following user query:
{query}

The correct label is: {decision}

Your goal is to explain **why this query should be classified as
{decision} for the target model**.

Write the explanation from the **first-person perspective of the
target model**, as if it is assessing its own capability.

You do NOT need to actually solve the query.
\end{verbatim}
\end{tcolorbox}

\end{document}